\documentclass[10pt,twocolumn,letterpaper]{article}

\usepackage[pagenumbers]{cvpr}              

\usepackage{array}
\newcolumntype{L}[1]{>{\raggedright\let\newline\\\arraybackslash\hspace{0pt}}m{#1}}
\newcolumntype{C}[1]{>{\centering\let\newline\\\arraybackslash\hspace{0pt}}m{#1}}
\newcolumntype{R}[1]{>{\raggedleft\let\newline\\\arraybackslash\hspace{0pt}}m{#1}}

\usepackage{multirow}
\usepackage[usestackEOL]{stackengine}
\usepackage[dvipsnames]{xcolor}

\definecolor{cvprblue}{rgb}{0.21,0.49,0.74}
\usepackage[pagebackref,breaklinks,colorlinks,citecolor=cvprblue]{hyperref}

\usepackage{array}
\usepackage[dvipsnames]{xcolor}
\usepackage{tcolorbox}
\newcommand{\smallTitle}[1]{{\noindent\textbf{{#1}}}}

\usepackage{makecell}
\usepackage{marvosym, ifsym} 

\title{VBench: Comprehensive Benchmark Suite for Video Generative Models}

\author{Ziqi Huang$^{1*}$
\quad
Yinan He$^{2*}$
\quad
Jiashuo Yu$^{2*}$
\quad
Fan Zhang$^{2*}$
\quad
Chenyang Si$^{1}$
\quad
Yuming Jiang$^{1}$\\
\quad
Yuanhan Zhang$^{1}$
\quad
Tianxing Wu$^{1}$
\quad
Qingyang Jin$^{1}$
\quad
Nattapol Chanpaisit$^{1}$\\
\quad
Yaohui Wang$^{2}$
\quad
Xinyuan Chen$^{2}$
\quad
Limin Wang$^{4,2}$
\quad
Dahua Lin\textsuperscript{2,3\Letter}
\quad
Yu Qiao\textsuperscript{2\Letter}
\quad
Ziwei Liu\textsuperscript{1\Letter}\\\\
$^{1}$S-Lab, Nanyang Technological University
\quad
$^{2}$Shanghai Artificial Intelligence Laboratory\\
\quad
$^{3}$The Chinese University of Hong Kong
\quad
$^{4}$Nanjing University
\\ \\
[-3pt]
\tt\normalsize\color{Magenta}\url{https://vchitect.github.io/VBench-project/} \\\\
}

\begin{document}

\twocolumn[{
            \renewcommand\twocolumn[1][]{#1}
            \vspace{-1em}
            \maketitle
            \vspace{-1em}
            \begin{center}
                \vspace{-20pt}
                \centering
                \includegraphics[width=1.0\textwidth]{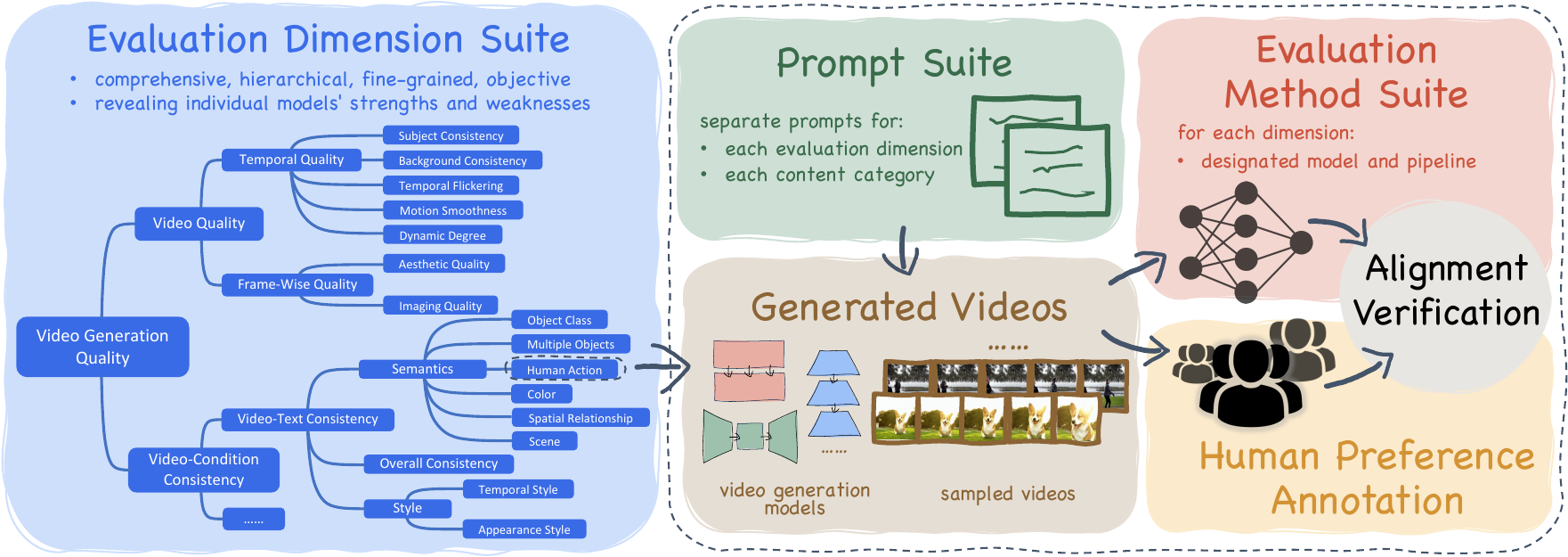}
                \vspace{-20pt}
                \captionof{figure} {
                \textbf{Overview of VBench.} We propose VBench, a comprehensive benchmark suite for video generative models. 
                We design a comprehensive and hierarchical \textbf{Evaluation Dimension Suite} to decompose ``video generation quality" into multiple well-defined dimensions to facilitate fine-grained and objective evaluation.  
                For each dimension and each content category, we carefully design a \textbf{Prompt Suite} as test cases, and sample \textbf{Generated Videos} from a set of video generation models. 
                For each evaluation dimension, we specifically design an \textbf{Evaluation Method Suite}, which uses a carefully crafted method or designated pipeline for automatic objective evaluation. We also conduct \textbf{Human Preference Annotation} for the generated videos for each dimension and show that VBench evaluation results are \textbf{well aligned with human perceptions}.
                VBench can provide valuable insights from multiple perspectives.
                }
                \label{fig:teaser}
            \end{center}
        }]


\begin{abstract}
Video generation has witnessed significant advancements, yet evaluating these models remains a challenge. A comprehensive evaluation benchmark for video generation is indispensable for two reasons: 1) Existing metrics do not fully align with human perceptions; 2) An ideal evaluation system should provide insights to inform future developments of video generation. 
%
\makeatletter{\renewcommand*{\@makefnmark}{}
\footnotetext{$^*$equal contributions. \textsuperscript{\Letter}corresponding authors. \hspace{5pt}\href{https://github.com/Vchitect/VBench}{Code} is available}\makeatother
}
To this end, we present \textbf{VBench}, a comprehensive benchmark suite that dissects ``video generation quality'' into specific, hierarchical, and disentangled dimensions, each with tailored prompts and evaluation methods. VBench has three appealing properties: \textbf{1) Comprehensive Dimensions:} VBench comprises 16 dimensions in video generation (\eg, subject identity inconsistency, motion smoothness, temporal flickering, and spatial relationship, \etc).  The evaluation metrics with fine-grained levels reveal individual models' strengths and weaknesses.  \textbf{2) Human Alignment:} We also provide a dataset of human preference annotations to validate our benchmarks' alignment with human perception, for each evaluation dimension respectively. \textbf{3) Valuable Insights:} We look into current models' ability across various evaluation dimensions, and various content types. We also investigate the gaps between video and image generation models. We will open-source VBench, including all prompts, evaluation methods, generated videos, and human preference annotations, and also include more video generation models in VBench to drive forward the field of video generation. 
\end{abstract}    
\vspace{-40pt}
\section{Introduction}

\label{sec:introduction}

\begin{figure}[t]
  \centering
    \vspace{-10pt}
   \includegraphics[width=0.9\linewidth]{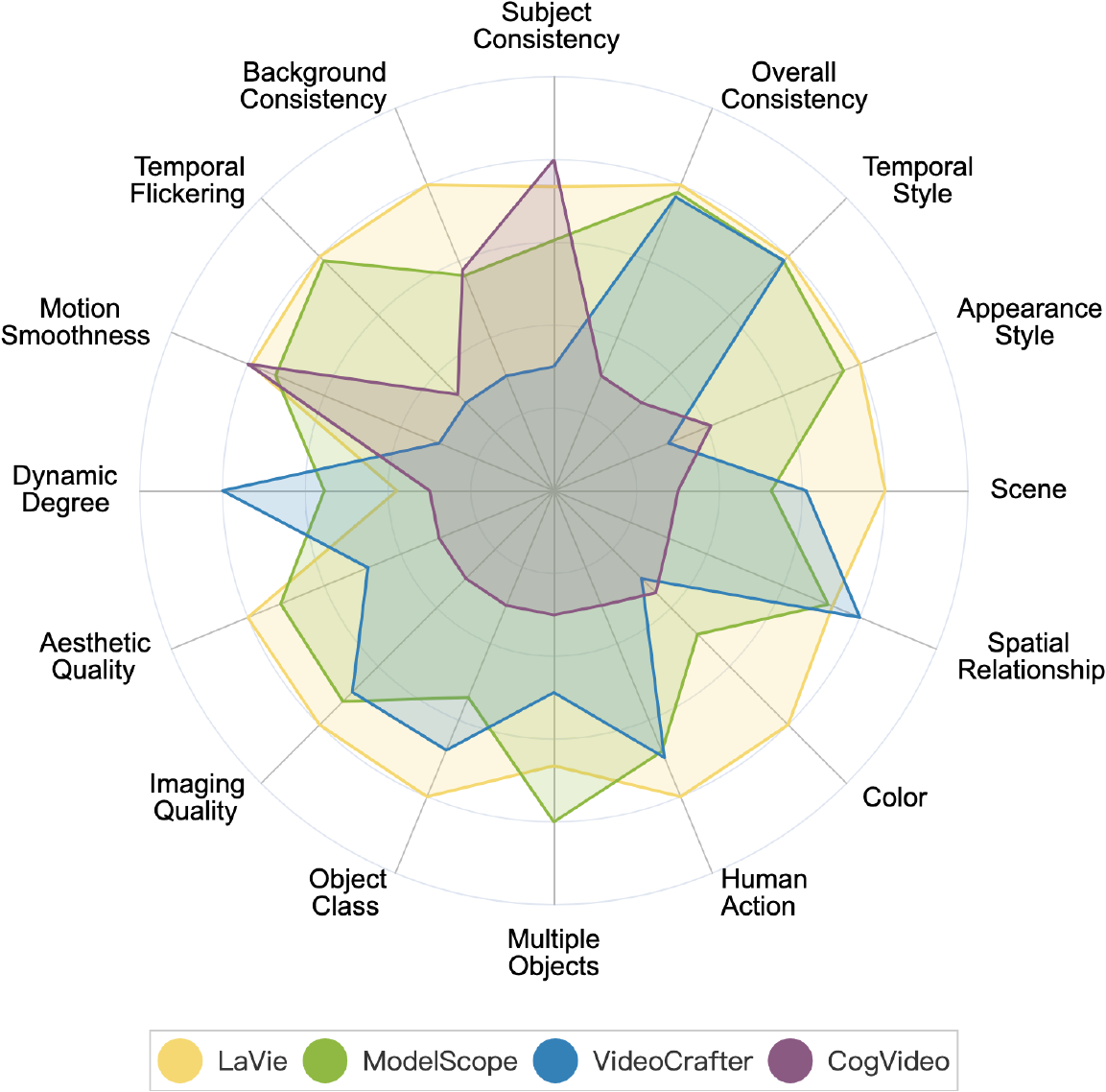}
   \vspace{-5pt}
   \caption{
   \textbf{VBench Evaluation Results of Video Generative Models.} We visualize the evaluation results of four video generation models in 16 VBench dimensions. We normalize the results per dimension for clearer comparisons.
   For comprehensive numerical results, please refer to Table~\ref{tab:raw_metrics}.
   }
    \label{fig:fig_paper_radar_big}
    \vspace{-10pt}
\end{figure}
%

Image generation models have made rapid progress in the past few years, such as Variational Autoencoders (VAEs)~\cite{kingma2013vae}, Generative Adversarial Networks (GANs)~\cite{goodfellow2014gan, mirza2014cgan, brock2018biggan, karras2018pggan, karras2019stylegan1, karras2020stylegan2, karras2021stylegan3, jiang2021talktoedit, fu2022stylegan, fu2023unitedhuman}, vector quantized (VQ) based approaches~\cite{van2017vqvae, esser2021vqgan, jiang2022text2human}, and diffusion models~\cite{ho2020ddpm, sohl2015deep, song2020score}. 
This fuels recent explorations in video generation~\cite{hong2022cogvideo, luo2023videofusion, singer2022makeavideo, wu2022tuneavideo, blattmann2023videoldm, he2022lvdm, zhang2023show1, wang2023lavie, wang2023modelscope}, which goes beyond static imagery and models the dynamics and kinematics of real-world scenes. 
With the growth of video generation models, there arises a critical need for effective evaluation methods. The evaluation should be able to accurately reflect human perception of generated videos, providing reliable measures of a model's performance. Additionally, it should reflect each model's specific strengths and weaknesses, offering insights that inform the data, training, and architectural choices of future video generation models.

However, existing metrics for video generation such as Inception Score (IS)~\cite{salimans2016inceptionscore}, Fréchet inception distance (FID)~\cite{heusel2017fid}, Fréchet Video Distance (FVD)~\cite{unterthiner2018fvd, unterthiner2019fvd}, and CLIPSIM~\cite{radford2021clip} are inconsistent with human judgement~\cite{ding2022cogview2, otani2023toward}. Meanwhile, the Video Quality Assessment (VQA) methods~\cite{wu2022fastervqa, wu2022fastvqa, wu2023dover, wu2023explainablevqa, wu2023bvqiplus, wu2023bvqi, wu2023discovqa, vsfa, videval} are primarily designed for real videos, thereby neglecting the unique challenges posed by generative models, such as artifacts in synthesized videos.
Hence, there is a pressing need for an evaluation framework that aligns closely with human perception, and specifically designed for the characteristics of video generation models. 

To this end, we introduce \textbf{VBench}, a comprehensive benchmark suite for evaluating video generation model performance. 
VBench has three appealing properties: 1) comprehensive evaluation dimensions, 2) human alignment, and 3) valuable insights.

First, our framework includes an \textit{evaluation dimension suite} that employs a hierarchical and disentangled approach to the decomposition of ``video generation quality''. 
This suite systematically breaks down the evaluation into two primary dimensions at a coarse level: \textit{Video Quality} and \textit{Video-Condition Consistency}. Each of these dimensions is further subdivided into more granular criteria. This hierarchical separation ensures that each dimension isolates and evaluates a single aspect of video quality, without interference from other variables, as illustrated in Figure~\ref{fig:teaser}.
Recognizing video generation's unique challenges, we have tailored evaluation dimensions to its specific characteristics. For example, in terms of \textit{Video Quality}, maintaining consistent subject identity (\eg, a teddy bear) in generated videos is crucial, and is a problem rarely encountered in real-world videos. Additionally, \textit{Video-Condition Consistency} is vital for conditional video generation tasks, requiring its dedicated evaluation criteria.
For each evaluation dimension, we carefully prepared around 100 text prompts as test cases for text-to-video (T2V) generation, and devised specialized  evaluation methods tailored to each dimension.
In addition to multi-dimensional evaluations, we also assess T2V models across \textit{diverse content categories}. 
We  organized prompt suites for eight distinct types, such as animal, architecture, human, and scenery, allowing for a separate evaluation within each category. 
This exploration reveals variable competencies in T2V generation across different content types, highlighting areas of proficiency and those requiring further enhancement.

Second, we systematically demonstrate that our evaluation method suite \textit{is closely aligned with human perception} in every fine-grained evaluation dimension.
We collected human preference annotations for each dimension. 
Specifically, we use various T2V models to sample videos from  our prompt suites. Then given two videos sampled from the same prompt, we ask human annotators to indicate preferences according to each VBench dimension respectively. We show that VBench evaluations highly correlate with human preferences. Additionally, the \textit{human preference annotations} can be utilized for multiple purposes, such as fine-tuning generation or evaluation models to enhance alignment with human perceptions. For instance, we utilize the annotations to implement Instruction Tuning within a Visual-Language Model (VLM), enhancing its T2V evaluation alignment with human preferences.

Third, VBench's multi-dimensional and multi-categorical approach can provide \textit{valuable insights} to the video generation community.
Our multi-dimensional system enables detailed feedback on the strengths and weaknesses of video generation models across various ability aspects. 
This approach not only ensures a comprehensive evaluation of existing models but also provides valuable insights into the training of advanced video generation models, guiding architectural and data choices for improved video generation outcomes. 
Additionally, VBench can be readily applied to evaluate image generation models, and thus we investigate the disparities between video and image generation models.
In Section~\ref{sec:insights}, we discuss in detail on various observations and insights drawn from VBench evaluations.

We are open-sourcing \textbf{VBench}, including its \textit{evaluation dimension suite}, \textit{evaluation method suite}, \textit{prompt suite}, \textit{generated videos}, and the dataset of \textit{human preference annotations}. We also encourage more video generation models to participate in the \textbf{VBench} challenge.

\section{Related Works}
 
\noindent \textbf{Video Generative Models.}
Recently, diffusion models~\cite{sohl2015deep, song2020score, ho2020ddpm, dhariwal2021beatgan} have achieved significant progress in image synthesis~\cite{nichol2021glide, gu2022vqdiffusion, saharia2022imagen, rombach2022ldm, podell2023sdxl, huang2023collaborative, huang2023reversion}, and enabled a line of works towards video generation~\cite{harvey2022fdm, ho2022videoDM, singer2022makeavideo, ho2022imagenvideo, wang2023lavie, he2022lvdm, zhou2023magicvideo, zhang2023show1, ge2023pyoco, blattmann2023videoldm, guo2023animatediff, wang2023modelscope, luo2023videofusion, zhou2022magicvideo, khachatryan2023text2videozero, chen2023videocrafter1, xing2023dynamicrafter, jiang2023text2performer}.
Many recent diffusion-based works~\cite{wang2023lavie, luo2023videofusion, wang2023modelscope, he2022lvdm} are text-to-video (T2V) models. Other guidance modalities are also available, including image-to-video~\cite{yin2023dragnuwa, esser2023structure, chen2023mcdiff, chen2023seine}, video-to-video~\cite{liew2023magicedit, qi2023fatezero, yang2023rerender, ouyang2023codef, chai2023stablevideo}, and a variety of control maps~\cite{zhang2023magicavatar, 2023videocomposer, khachatryan2023text2videozero, ma2023follow, zhang2023controlvideo, chen2023controlavideo, dreampose_2023} such as pose, depth, and sketch.
The boom of video generation models requires a comprehensive evaluation system to inform their current capabilities and guide future developments, and VBench takes the initiative in providing a comprehensive benchmark suite for fine-grained and human-aligned evaluation.

\noindent \textbf{Evaluation of Visual Generative Models.}
Existing video generation models typically use metrics like Inception Score (IS)~\cite{salimans2016inceptionscore}, Fréchet inception distance (FID)~\cite{heusel2017fid}, Fréchet Video Distance (FVD)~\cite{unterthiner2018fvd}, and CLIPSIM~\cite{radford2021clip} for evaluation.
The UCF-101~\cite{soomro2012ucf101} dataset's class labels often serve as text prompts for IS, FID, and FVD, whereas MSR-VTT~\cite{xu2016msr}'s human-labeled video captions are used for CLIPSIM.
Despite covering various real-world scenarios, these prompts lack diversity and specificity, limiting accurate and fine-grained evaluation of video generation.
For text-to-image (T2I) models, several benchmarks~\cite{huang2023t2icompbench, wang2022imagenedit, saharia2022imagen, lee2023holistic, basu2023editval, bakr2023hrsbench} are proposed to assess various capabilities like compositionality~\cite{huang2023t2icompbench} and editing ability~\cite{wang2022imagenedit, basu2023editval}.
However, video generative models still lack comprehensive evaluation benchmarks for detailed and human-aligned feedback. 
Our work differs from concurrent research~\cite{liu2023evalcrafter, liu2023fetv} in three key ways: 1) We have created 16 distinct evaluation dimensions, each with specialized prompts for precise assessment; 2) We have empirically validated that every dimension aligns closely with human perception; 3) Our multi-dimensional and multi-categorical evaluation offers valuable and comprehensive insights into video generation.
\section{VBench Suite}

In this section, we introduce the main components of VBench. In Section~\ref{subsec:evaluation_dimension_suite}, we present our rationale for designing the 16 evaluation dimensions, as well as each dimension's definition and evaluation method. We then elaborate on the prompt suites we use in Section~\ref{subsec:prompt_suite}. To validate VBench's alignment with human perception, we conduct human preference annotation for each dimension (see Section~\ref{subsec:human_preference_annotation}).
The experiments and the insights drawn from VBench will be detailed in Section~\ref{sec:experiments} and Section~\ref{sec:insights}.

\subsection{Evaluation Dimension Suite}
\label{subsec:evaluation_dimension_suite}

We first introduce our evaluation dimensions and their corresponding evaluation methods.

Existing evaluation metrics like FVD~\cite{unterthiner2018fvd} often conclude video generation model performance to a single number. This oversimplifies the evaluation and has several risks. 
First, a single number can obscure an individual model's strengths and weaknesses, and it fails to provide insights into specific areas where a model excels or underperforms. This makes it challenging to derive insights for future architectural and training designs based on single-valued metrics.
Second, the notion of ``high-quality video generation'' is complex and multifaceted, with individuals prioritizing different video attributes based on the intended application. 
For instance, some may prioritize the absence of temporal flickering, while others may consider fidelity to the text prompt as the most significant, with less emphasis on flickering.
Therefore, in contrast with performing single-valued evaluations of video generation quality, we propose a disaggregated approach by decomposing the brand notion of ``video generation performance'' into multiple discrete dimensions for fine-grained evaluation.

Specifically, we break ``video generation quality'' down into 16 disentangled dimensions in a top-down manner, with each evaluation dimension assessing one aspect of video generation quality. 
On the top level, we evaluate T2V performance from two broad perspectives:
\textbf{1) Video Quality} --- \textit{``Without considering alignment with the text prompt, does the video alone look good?''}, which focuses on the perceptual quality of the synthesized video, and does not consider the input condition (\eg, text prompt), and \textbf{2) Video-Condition Consistency} --- \textit{``Is the video consistent with what the user wants to generate?''}, which focuses on whether the synthesized video is consistent with the guiding condition that the user provides (\eg, the text prompt for T2V generation). Under both \textit{``Video Quality''} and \textit{``Video-Condition Consistency''}, we further break the coarse-grained dimensions into more fine-grained dimensions, as shown in Figure~\ref{fig:teaser}. 

\subsubsection{Video Quality}

We split \textit{``Video Quality''} into two disentangled aspects, \textit{``Temporal Quality''} and \textit{``Frame-Wise Quality''}, where the former only considers the cross-frame consistency and dynamics, and the latter only considers the quality of each individual frame without taking temporal quality into concern. For \textit{``Temporal Quality''}, we further devise five different evaluation dimensions, where each focusing on a different aspect of temporal quality. 
We briefly introduce each dimension here.
\textit{Please refer to the Supplementary File for the detailed definition and evaluation method of each dimension.}

\smallTitle{Temporal Quality - Subject Consistency.} 
For a subject (\eg, a person, a car, or a cat) in the video, we assess whether its appearance remains consistent throughout the whole video. To this end, we calculate the DINO~\cite{caron2021emerging} feature similarity across frames. 

\smallTitle{Temporal Quality - Background Consistency.}
We evaluate the temporal consistency of the background scenes by calculating  CLIP~\cite{radford2021clip} feature similarity across frames.

\smallTitle{Temporal Quality - Temporal Flickering.} Generated videos can exhibit imperfect temporal consistency at \textit{local and high-frequency details}. We take static frames and compute the mean absolute difference across frames.

\smallTitle{Temporal Quality - Motion Smoothness.} 
Both \textit{Subject Consistency} and \textit{Background Consistency} focus on temporal consistency of the ``look'' instead of the smoothness of ``movement and motion''. We believe it is important to evaluate whether the motion in the generated video is smooth, and follows the physical law of the real world. We utilize the motion priors in the video frame interpolation model~\cite{licvpr23amt} to evaluate the smoothness of generated motions (see the detailed method in Supplementary File).

\smallTitle{Temporal Quality - Dynamic Degree.} 
Since a completely static video can score well in the aforementioned temporal quality dimensions, it is important to also evaluate the degree of dynamics (\ie, whether it contains large motions) generated by each model. We use RAFT~\cite{teed2020raft} to estimate the degree of dynamics in synthesized videos.

\smallTitle{Frame-Wise Quality - Aesthetic Quality.} 
We evaluate the artistic and beauty value perceived by humans towards each video frame using the LAION aesthetic predictor~\cite{LAIONaes}. It can reflect aesthetic aspects such as the layout, the richness and harmony of colors, the photo-realism, naturalness, and artistic quality of the video frames.

\smallTitle{Frame-Wise Quality - Imaging Quality.} 
Imaging quality refers to the distortion \textit{(e.g., over-exposure, noise, blur)} presented in the generated frames, and we evaluate it using the MUSIQ~\cite{Ke2021MUSIQ} image quality predictor trained on the SPAQ~\cite{Fang2020spaq} dataset.

\begin{figure*}
    \centering
    \begin{minipage}{0.65\textwidth}
        \centering
        \vspace{-5pt}
        \includegraphics[width=0.99\linewidth]{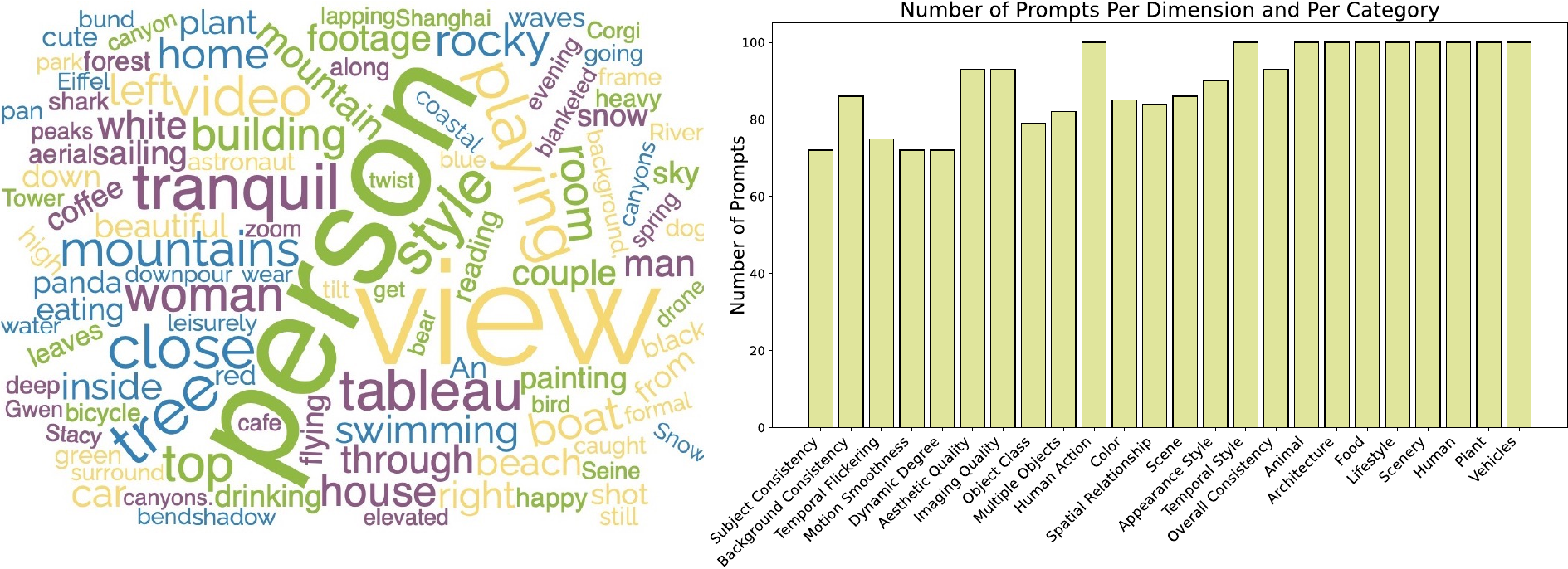}
        \vspace{-10pt}
        \caption{\textbf{Prompt Suite Statistics.} The two graphs provide an overview of our prompt suites. \textbf{\textit{Left:}} the word cloud to visualize word distribution of our prompt suites. \textbf{\textit{Right:}} the number of prompts across different evaluation dimensions and different content categories.}
        \label{fig:prompt_suite_statistics}
    \end{minipage}\hspace{0.01\textwidth}
    \begin{minipage}{0.33\textwidth}
        \centering
        \vspace{-5pt}
        \includegraphics[width=0.99\linewidth]{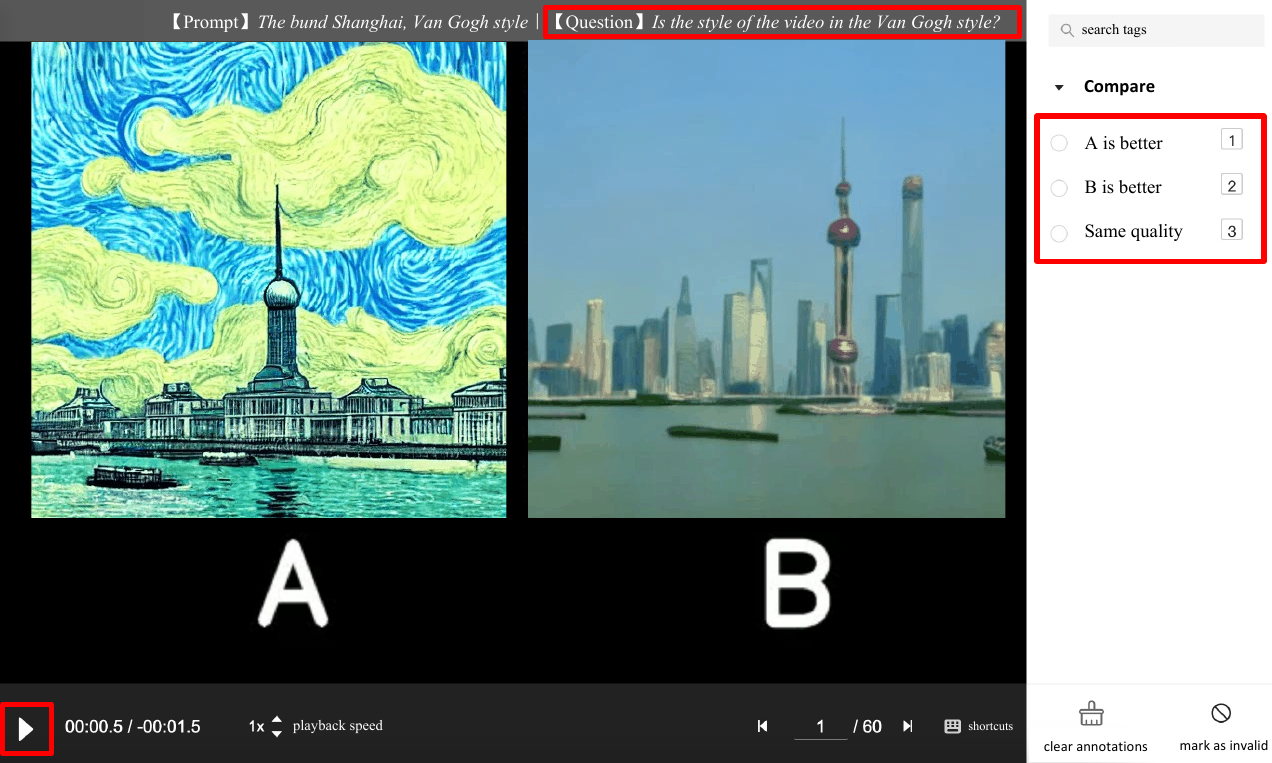}
        \vspace{-13pt}
        \caption{\textbf{Interface for Human Preference Annotation}. \textbf{\textit{Top:}} prompt and question. \textbf{\textit{Right:}} choices that annotators can make. \textbf{\textit{Bottom left:}} control for stop and playback.}
        \label{fig:fig_paper_interface}
    \end{minipage}
\vspace{-10pt}
\end{figure*}

\subsubsection{Video-Condition Consistency}

We mainly dissect \textit{``Video-Condition Consistency''} into \textit{``Semantics''} (\ie, the type of the entities and their attributes) and \textit{``Style''} (\ie, whether the generated video is consistent with user-requested style), with each decomposed into more fine-grained dimensions.

\smallTitle{Semantics - Object Class.} 
We use GRiT~\cite{wu2022grit} to detect the success rate of generating the specific class of objects depicted in the text prompt.

\smallTitle{Semantics - Multiple Objects.} 
Other than generating a single object of a particular class, the ability to compose multiple objects from different classes in the same frame is also an essential ability in video generation. We detect the success rate of generating all the objects specified in the text prompt within each video frame.

\smallTitle{Semantics - Human Action.} 
Human action is an important aspect in human-centric video generation. We apply UMT~\cite{li2023unmasked} to evaluate whether human subjects in generated videos can accurately execute the specific actions mentioned in the text prompts.

\smallTitle{Semantics - Color.}
To evaluate whether synthesized object colors align with the text prompt, we use GRiT~\cite{wu2022grit} to provide color captioning, and compare against the expected color.

\smallTitle{Semantics - Spatial Relationship.} 
Other than classes and attributes of synthesized objects, we also evaluate whether their spatial relationship follows what is specified by the text prompt.
We focus on four primary types of spatial relationships, and perform rule-based evaluation similar to~\cite{huang2023t2icompbench}.

\smallTitle{Semantics - Scene.} 
We need to evaluate whether the synthesized video is consistent with the intended scene described by the text prompt. For example, when prompted ``ocean'', the generated video should be ``ocean'' instead of ``river''.
We use Tag2Text~\cite{huang2023tag2text} to caption the generated scenes, and then check its correspondence with scene descriptions in the text prompt.

\smallTitle{Style - Appearance Style.} 
Apart from semantics consistency with the text prompt, another important pillar in video-condition consistency is \textit{style}. There are many styles that alter the look, color, and texture of synthesized video frames, such as ``oil painting style'', ``black and white style'', ``watercolor painting style'',       cyberpunk style'', ``black and white'' \etc. We calculate the CLIP~\cite{radford2021clip} feature similarity between synthesized frames and these style descriptions.

\smallTitle{Style - Temporal Style.} 
Apart from appearance styles, videos also have temporal styles like various camera motions. We use ViCLIP~\cite{wang2023internvid} to calculate the video feature and the temporal style description feature similarity to reflect temporal style consistency.

\smallTitle{Overall Consistency.} 
We further use overall video-text consistency computed by ViCLIP~\cite{wang2023internvid} on general text prompts as an aiding metric to reflect both semantics and style consistency.

\textit{For each dimension, please refer to the Supplementary File} for: 1) details of its definition, 2) positive and negative examples (\ie, synthesized videos) of each dimension, and 3) detailed evaluation method and pipeline implementations.

\subsection{Prompt Suite}
\label{subsec:prompt_suite}
The sampling procedure of current diffusion-based video generation models~\cite{wang2023lavie, wang2023modelscope, he2022lvdm} is time-consuming (\eg, 90 seconds per video for LaVie~\cite{wang2023lavie}, and more than 2 minutes per video for CogVideo~\cite{hong2022cogvideo}). Therefore, we need to control the amount of test cases for efficient evaluation. Meanwhile, we need to maintain the diversity and comprehensiveness of our prompt suite, so we design compact yet representative prompts in terms of both the evaluation dimensions and the content categories. We visualize our prompt suite distributions in Figure~\ref{fig:prompt_suite_statistics}.

\smallTitle{Prompt Suite per Dimension.}
For each VBench evaluation dimension, we carefully designed a set of around 100 prompts as the test cases. The prompt suite is carefully curated to test the specific ability corresponding to the dimension tested. For example, for the \textit{``Subject Consistency''} dimension which aims to evaluate the consistency of subjects' appearances throughout the video, we ensure every prompt has a movable subject (\eg, animals or vehicles) performing non-static actions, where their consistency might be compromised due to inconsistency introduced by their movements or changing locations.
For the dimension \textit{``Object Class''}, we ensure the existence of a specific class of object in every prompt. For \textit{``Human Action''}, each test prompt contains a human subject performing a well-defined action from the Kinetics-400 dataset~\cite{kay2017kinetics}, where 100 representative actions are selected with minimal semantic overlaps among themselves.
Please refer to the Supplementary File for the design rationale of the prompt suite for each of the 16 dimensions.

\smallTitle{Prompt Suite per Category.}
When designing prompts for each dimension, the focus was to showcase models' ability in that specific dimension. We further incorporate prompt suites for eight content categories to provide insights into the performance across varied content types.
To this end, we prepare a collection of human-curated prompts from the Internet and divide them into 8 distinctive categories following YouTube's categorization. Subsequently, we feed both the category labels and prompts into a Large Language Model (LLM)~\cite{zheng2023judging} (see more implementation details in Supplementary File), obtaining multi-label outputs for each caption.  We select 800 prompts and manually clean their labels to serve as per-category prompt suites. Finally, we obtain 100 prompts for each of these eight categories: \texttt{Animal}, \texttt{Architecture}, \texttt{Food}, \texttt{Human}, \texttt{Lifestyle}, \texttt{Plant}, \texttt{Scenery}, and \texttt{Vehicles}.

\subsection{Human Preference Annotation}  
\label{subsec:human_preference_annotation}

We perform human preference labeling on massive generated videos. The primary goal is to validate \textit{VBench evaluation's alignment with human perception in each of the 16 evaluation dimensions}, and the verification results will be detailed in Section~\ref{subsec:validating}. We also show that our human preference annotations can be useful in future tasks of finetuning generation and evaluation models to enhance alignment with human perceptions.

\smallTitle{Data Preparation.} Given a text prompt $p_{i}$, and four video generation models to be evaluated $\{A, B, C, D\}$, we use each model to generate a video, forming a ``group'' of videos $G_{i,j}=\{V_{i,A,j}, V_{i,B,j}, V_{i,C,j}, V_{i,D,j}\}$. For each prompt $p_{i}$, we sample five such groups of videos $\{G_{i,0}, G_{i,1}, G_{i,2}, G_{i,3}, G_{i,4}\}$. For each group, we pair the videos up in pair-wise combinations, yielding six pairs: 
$(V_{A}, V_{B})$, $(V_{A}, V_{C})$, $(V_{A}, V_{D})$, $(V_{B}, V_{C})$, $(V_{B}, V_{D})$, $(V_{C}, V_{D})$,
and ask human annotators to indicate their preferred video for each pair. 
Within the VBench evaluation framework, a prompt suite of $N$ prompts produces $N \times 5 \times 6$ pairwise video comparisons. The video order within each pair is randomized to ensure unbiased annotation.

\begin{table*}[htbp]
\centering
\setlength\tabcolsep{3pt}
\begin{center}
\small 
\vspace{-10pt}
\caption{\textbf{VBench Evaluation Results per Dimension.} This table compares the performance of four video generation models across each of the 16 VBench dimensions. A higher score indicates relatively better performance for a particular dimension. 
We also provide two specially built baselines, \ie,  Empirical Min and Max (the approximated achievable min and max scores for each dimension), as references.
}
\vspace{-6pt}
\resizebox{0.8\linewidth}{!}{
\begin{tabular}{c|c|c|c|c|c|c|c|c}
\Xhline{1pt}
\textbf{Models}   & \textbf{\Centerstack{Subject\\Consistency}} & \textbf{\Centerstack{Background\\Consistency}} & 
\textbf{\Centerstack{Temporal\\Flickering}} & \textbf{\Centerstack{Motion\\Smoothness}} & \textbf{\Centerstack{Dynamic\\Degree}} & \textbf{\Centerstack{Aesthetic\\Quality}} &  \textbf{\Centerstack{Imaging\\Quality}} & \textbf{\Centerstack{Object\\Class}} \\ \Xhline{1pt}
LaVie~\cite{wang2023lavie}        & 91.41\% & \textbf{97.47\%} & \textbf{98.30\%} & 96.38\% & 49.72\% & \textbf{54.94\%} & \textbf{61.90\%} & \textbf{91.82\%} \\ 
ModelScope~\cite{luo2023videofusion, wang2023modelscope}  & 89.87\% & 95.29\% & 98.28\% & 95.79\% & 66.39\% & 52.06\% & 58.57\% & 82.25\% \\ 
VideoCrafter~\cite{he2022lvdm} & 86.24\% & 92.88\% & 97.60\% & 91.79\% & \textbf{89.72\%} & 44.41\% & 57.22\% & 87.34\% \\ 
CogVideo~\cite{hong2022cogvideo}  & \textbf{92.19\%} & 95.42\% & 97.64\% & \textbf{96.47\%} & 42.22\% & 38.18\% & 41.03\% & 73.40\% \\ \hline
Empirical Min  & 14.62\% & 26.15\% & 62.93\% & 70.60\% & 0.00\% & 0.00\% & 0.00\% & 0.00\%  \\
Empirical Max  & 100.00\% & 100.00\% & 100.00\% & 99.75\% & 100.00\% & 100.00\% & 100.00\% & 100.00\% \\ \hline
\hline
\textbf{Models}   & \textbf{\Centerstack{Multiple\\Objects}} & \textbf{\Centerstack{Human\\Action}} & \textbf{Color} & \textbf{\Centerstack{Spatial\\Relationship}} & \textbf{Scene} & \textbf{\Centerstack{Appearance\\Style}} & \textbf{\Centerstack{Temporal\\Style}} & \textbf{\Centerstack{Overall\\Consistency}} \\
\Xhline{1pt}
LaVie~\cite{wang2023lavie}        & 33.32\% & \textbf{96.80\%} & \textbf{86.39\%} & 34.09\% & \textbf{52.69\%} & \textbf{23.56\%} & \textbf{25.93\%} & \textbf{26.41\%} \\ 
ModelScope~\cite{luo2023videofusion, wang2023modelscope}  & \textbf{38.98\%} & 92.40\% & 81.72\% & 33.68\% & 39.26\% & 23.39\% & 25.37\% & 25.67\% \\ 
VideoCrafter~\cite{he2022lvdm} & 25.93\% & 93.00\% & 78.84\% & \textbf{36.74\%} & 43.36\% & 21.57\% & 25.42\% & 25.21\% \\ 
CogVideo~\cite{hong2022cogvideo}  & 18.11\% & 78.20\% & 79.57\% & 18.24\% & 28.24\% & 22.01\% & 7.80\% & 7.70\% \\ \hline
Empirical  Min  & 0.00\% & 0.00\% & 0.00\% & 0.00\% & 0.00\% & 0.09\% & 0.00\% & 0.00\% \\
Empirical Max  & 100.00\% & 100.00\% & 100.00\% & 100.00\% & 82.22\% & 28.55\% & 36.40\% & 36.40\% \\
\Xhline{1pt}
\end{tabular}
}
\vspace{-20pt}
\label{tab:raw_metrics}
\end{center}
\end{table*}

\smallTitle{Human Labeling Rules.} Specifically, the human annotators are asked to only consider the specific evaluation dimension of interest and select the preferred video. For example, in Figure~\ref{fig:fig_paper_interface}, for the \textit{Appearance Style} dimension, the question is \textit{``Is the style of the video in the Van Gogh style?''}, and human annotators are instructed to only focus on whether the generated video's style belongs to the Van Gogh style and should not consider other quality aspects of the generated video, such as potential issues like the degree of temporal flickering. In the example in this figure, video A resembles the Van Gogh better than video B, and the annotator is expected to select ``A is better". For every dimension, we carefully prepare instructions and train the human annotators to understand the definition of the dimension, and perform multiple quality assurance protocols via a pre-labeling trial, and two rounds of post-labeling checks (see more details in the Supplementary File).
 
\smallTitle{Annotations for VLM Tuning.} 
We map VBench evaluation scores from various dimensions to the scale of 0-10 and combine them with human preference annotations to form the instruction data, which is then used to fine-tune the pre-trained VideoChat~\cite{2023videochat} model to demonstrate improved evaluation capabilities. For implementation details and tuning results, please refer to the Supplementary File.

\section{Experiments}
\label{sec:experiments}
 
We adopt the video generation models LaVie~\cite{wang2023lavie}, ModelScope~\cite{wang2023modelscope, luo2023videofusion}, VideoCrafter~\cite{he2022lvdm}, and CogVideo~\cite{hong2022cogvideo} for VBench evaluation, and more will be added as they become open-sourced. Details of the models and sampling procedures are in the Supplementary File.

\subsection{Per-Dimension Evaluation}
\vspace{-3pt}

\label{subsec:exp_and_discuss_per_dimension}

For every dimension, we calculate the VBench scores using the evaluation method suite described in Section~\ref{subsec:evaluation_dimension_suite}, and show the results using Figure~\ref{fig:fig_paper_radar_big} and Table~\ref{tab:raw_metrics}.
We additionally designed three reference baselines, namely \textit{Empirical Max}, \textit{Empirical Min}, and \textit{WebVid-Avg}. The first two approximate the maximum / minimum scores that videos might be able to achieve, and \textit{WebVid-Avg} reflects the WebVid-10M~\cite{bain2021frozen} dataset quality in each VBench dimension.

\smallTitle{Empirical Max.} For most dimensions, to approximate the maximum achievable values, we first retrieve WebVid-10M~\cite{bain2021frozen}  videos according to our prompt suites. 
We use CLIP~\cite{radford2021clip} to extract text features of both WebVid-10M's captions and our prompts. For each prompt, we retrieve the top-5 WebVid-10M videos according to text feature similarity with the given prompt. 
Given that the generated videos are usually 2 seconds in length, we randomly select a 2-second segment from each retrieved video and sample frames at 8 frames per second (FPS). 
For each dimension, we use the retrieved videos according to its prompt suite and report the highest-scoring video's result as \textit{Empirical Max}. 

\smallTitle{Empirical Min.} To approximate the minimum achievable values, we use randomly generated 2-second Gaussian noise clips to calculate results for the \textit{``Video-Condition Consistency''} dimensions. 
For most \textit{``Video Quality''} dimensions, we select frames from real videos and design frame concatenation for each dimension, approximating the minimum score achievable for each VBench dimension.

\smallTitle{WebVid-Avg.}  Similar to \textit{Empirical Max}, we compute the average for each dimension on retrieved WebVid-10M~\cite{bain2021frozen} videos. This baseline could reflect the average per-dimension quality of the commonly used video generation training dataset WebVid-10M, and provide a reference for model performances.
The comparison against \textit{WebVid-Avg} and \textit{Empirical Max} is visualized in Figure~\ref{fig:fig_paper_radar_sd} (b).

 \begin{figure*}[t]
  \centering
   \vspace{-15pt}
   \includegraphics[width=0.99\linewidth]{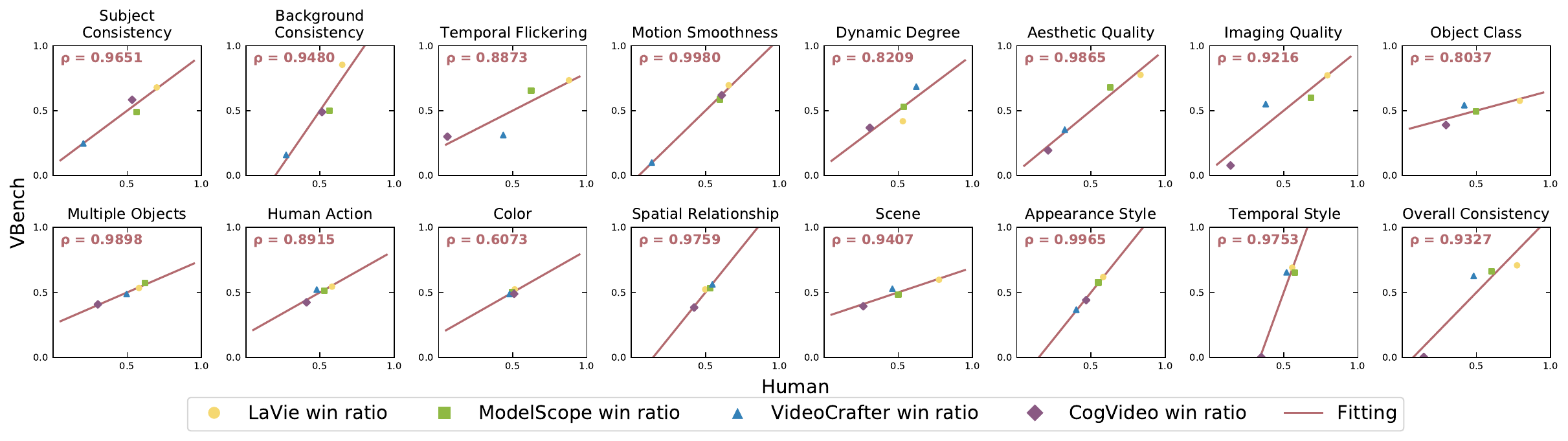}
   \vspace{-10pt}
   \caption{
   \textbf{Validate VBench's Human Alignment.} 
   Our experiments show that \textbf{\textit{VBench evaluations across all dimensions closely match human perceptions. }}
   Each plot shows the alignment verification result of a specific VBench dimension. In each plot, a dot represents the human preference win ratio (horizontal axis) and VBench evaluation win ratio (vertical axis) for a particular video generation model. We linearly fit a straight line to visualize the correlation, and calculate the Spearman's correlation coefficient ($\rho$) for each dimension. 
   }
   \label{fig:win_ratio_dot}
   \vspace{-10pt}
\end{figure*}

\subsection{Validating Human Alignment of VBench}
\label{subsec:validating}
\vspace{-3pt}

To validate that our evaluation method can faithfully reflect human perception, we performed a large-scale human annotation for each dimension, as mentioned in Section~\ref{subsec:human_preference_annotation}. We show the correlation between VBench evaluation results and human preference annotations in Figure~\ref{fig:win_ratio_dot}.

\smallTitle{Win Ratio.} Given the human labels, we calculate the win ratio of each model. During pairwise comparisons, if a model's video is selected as better, then the model scores 1 and the other model scores 0. If there is a tie, then both models score 0.5. For each model, the win ratio is calculated as the total score divided by the total number of pairs-wise comparisons participated.

\smallTitle{Per-Dimension Evaluation.} For each evaluation dimension, we calculate the model win ratio based on (1) VBench evaluation results, and (2) human annotation results, respectively, and compute their correlations, as shown in Figure~\ref{fig:win_ratio_dot}. We observe that \textit{VBench's per-dimension evaluation results are highly correlated with human preference annotations}.

\subsection{Per-Category Evaluation}
We evaluate the T2V models across eight different content categories, by generating videos based on \textit{Prompt Suite per Category} described in Section~\ref{subsec:prompt_suite}, and then calculating their performance across different evaluation dimensions. 
Figure~\ref{fig:fig_paper_radar_per_class} visualizes the evaluation results of each model in terms of the eight content categories.

\subsection{Video Generation V.S. Image Generation}
\vspace{-3pt}

\begin{figure}[t]
  \centering
   \includegraphics[width=1.01\linewidth]{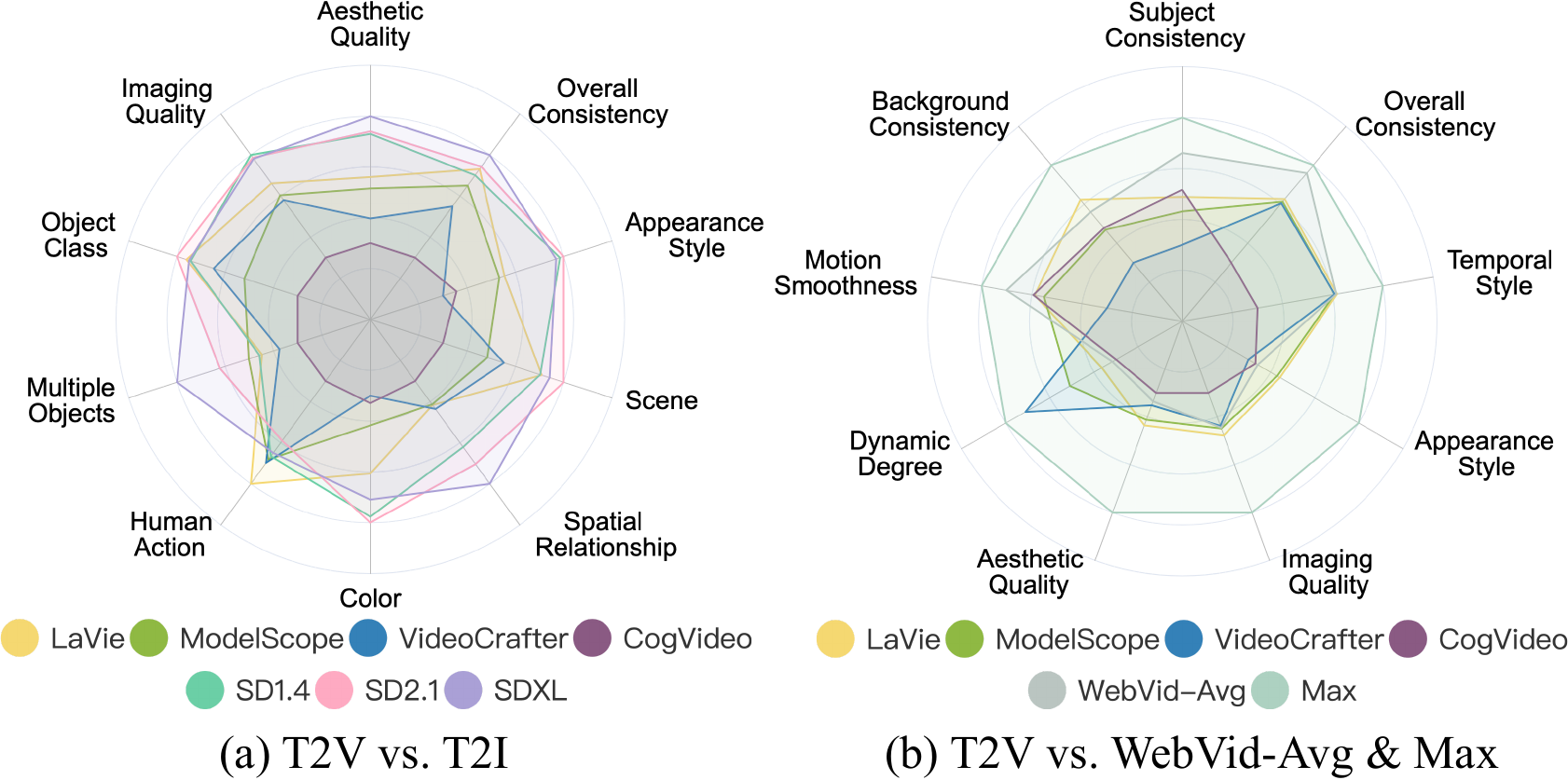}
   \vspace{-20pt}
   \caption{\textbf{More Comparisons of Video Generation Models with Other Models and Baselines.} We use VBench to evaluate other models and baselines for further comparative analysis of T2V models. \textbf{(a)} Comparison with text-to-image (T2I) generation models.
  \textbf{(b)} Comparison with \textit{WebVid-Avg} and \textit{Empirical Max} baselines.
   See the Supplementary File for comprehensive numerical results and details on normalization methods.
   }
   \label{fig:fig_paper_radar_sd}
   \vspace{-10pt}
\end{figure}

We conduct a comparative analysis of the frame-wise generation capability exhibited by text-to-video (T2V) models and text-to-image (T2I) models with two primary objectives: first, to assess the extent to which T2V models have successfully inherited the frame-wise generative capability of the T2I models; and second, to investigate the frame-wise generation capability gap between existing T2I and T2V models.
As an initial exploration into this problem, we compare video generation models with three image generation models, namely Stable Diffusion (SD) 1.4~\cite{rombach2022ldm}, SD2.1~\cite{rombach2022ldm}, and SDXL~\cite{podell2023sdxl}. We choose 10 VBench dimensions that can encompass frame-wise generation capabilities, and sample frames from all the image and video generation models according to \textit{Prompt Suite per Evaluation Dimension} described in Section~\ref{subsec:prompt_suite}. Figure~\ref{fig:fig_paper_radar_sd} (a) visualizes the evaluation results of T2V versus T2I models.

\section{Insights and Discussions}
\label{sec:insights}

In this section, we discuss the observations and insights we draw from our comprehensive evaluation experiments.

\begin{figure*}[t]
  \centering
  \vspace{-10pt}
   \includegraphics[width=0.99\linewidth]{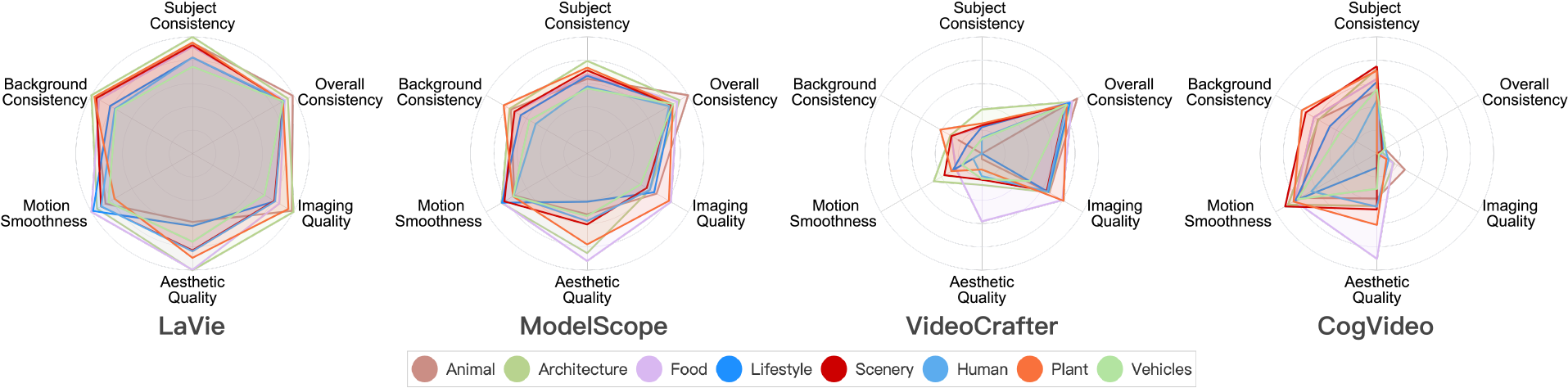}
   \vspace{-5pt}
   \caption{
   \textbf{VBench Results across Eight Content Categories} (best viewed in color). For each chart, we plot the VBench evaluation results across eight different content categories, benchmarked by our \textit{Prompt Suite per Category}. The results are linearly normalized between 0 and 1 for better visibility across categories. See the Supplementary File for comprehensive numerical results, and normalization details. }
   \label{fig:fig_paper_radar_per_class}
   \vspace{-10pt}
\end{figure*}

\smallTitle{$\boldsymbol{\cdot}$ Trade-off across Ability Dimensions.} We have noticed a trade-off in video generation models between \textbf{\textit{1)}} temporal consistency (\textit{Subject Consistency}, \textit{Background Consistency}, \textit{Temporal Flickering}, \textit{Motion Smoothness}) and \textbf{\textit{2)}} \textit{Dynamic Degree}. Models strong in temporal consistency often have a lower \textit{Dynamic Degree}, as these two aspects are somewhat complementary (see Figure~\ref{fig:fig_paper_radar_big} and Table~\ref{tab:raw_metrics}). For example, LaVie excels in \textit{Background Consistency} and \textit{Temporal Flickering} but has a low \textit{Dynamic Degree}, probably because generating relatively static scenes can ``cheat'' to get high temporal consistency scores. Conversely, VideoCrafter shows a high \textit{Dynamic Degree} but suffers from poor performance in all temporal consistency dimensions. This trend highlights the current challenge for models to achieve temporal consistency with dynamic content of large motions. Future research should focus on enhancing both aspects simultaneously, as improving only one might indicate compromising the other.

\noindent\textbf{$\boldsymbol{\cdot}$ Uncovering Hidden Potential of T2V Models in Specific Content Categories.} 
Our analysis reveals that the capabilities of some models vary significantly across different content types. 
For instance, for \textit{Aesthetic Quality}, CogVideo scores well for \texttt{Food} (see Figure~\ref{fig:fig_paper_radar_per_class} rightmost chart),
whereas it underperforms in others like \texttt{Animal} and \texttt{Vehicles}. The average results across various prompts might suggest a lower overall \textit{``Aesthetic Quality''} (as seen in Figure~\ref{fig:fig_paper_radar_big}), but CogVideo demonstrates relatively strong aesthetics in at least the \texttt{Food} category. This suggests that with tailored training data and strategies, CogVideo could potentially match other models in aesthetics by improving such ability in other content types. Therefore, we recommend \textit{evaluating video generation models not just based on ability dimensions but also considering specific content categories to uncover their hidden potential}.

\smallTitle{$\boldsymbol{\cdot}$ Bottleneck in Temporally Complex Categories Affecting Spatial and Temporal Performance.}
For spatially complex categories (\eg, \texttt{Animal}, \texttt{LifeStyle}, \texttt{Human}, \texttt{Vehicles}), models all perform relatively poorly mainly in \textit{Aesthetic Quality} (shown in Figure~\ref{fig:fig_paper_radar_per_class}). This is likely due to the challenges in synthesizing harmonious color schemes, articulated structures, and appealing layouts amidst complex elements. On the other hand, for categories involving complex and intense motions like \texttt{Human} and \texttt{Vehicle} (see their \textit{Dynamic Degree} in Supplementary File), performance is relatively poor across \textit{all dimensions}. 
This suggests that motion complexity and dynamic intensity significantly hinder synthesis, impacting both spatial and temporal dimensions,
probably because poor temporal modeling results in distorted and blurred imagery. This highlights the need for improved handling of dynamic motions in video generation models.

\smallTitle{$\boldsymbol{\cdot}$ Challenges of Data Quantity in Handling Complex Categories like Human.}
The WebVid-10M dataset \cite{bain2021frozen} allocates 26\% of its content to the \texttt{Human} category, which is the largest share among the eight categories (see statistics in Supplementary File).
However, the \texttt{Human} category exhibits one of the poorest results among eight categories (see Figure~\ref{fig:fig_paper_radar_per_class}). 
This suggests that merely increasing data volume may not significantly enhance performance in complex categories like \texttt{Human}.
A potential approach could involve integrating human-related priors or controls, such as skeletons, to better capture the articulated nature of human appearances and movements.

\smallTitle{$\boldsymbol{\cdot}$ Prioritizing Data Quality Over Quantity in Large-Scale Datasets.}
For \textit{Aesthetic Quality}, Figure~\ref{fig:fig_paper_radar_per_class} shows that the \texttt{Food} category almost always tends to have the highest scores among all categories.
This is corroborated by the WebVid-10M dataset \cite{bain2021frozen}, where \texttt{Food} ranks highest in \textit{Aesthetic Quality} according to VBench evaluation (refer to Supplementary File for more details), despite comprising just 11\% of the total data.
This observation suggests that at million scales, data quality might hold greater importance than quantity. Furthermore, \textit{VBench's evaluation dimensions can be potentially useful for cleaning datasets in specified quality dimensions}.

\smallTitle{$\boldsymbol{\cdot}$ Compositionality: T2I versus T2V.} As shown in Figure~\ref{fig:fig_paper_radar_sd} (a), T2V models significantly underperform in \textit{Multiple Objects} and \textit{Spatial Relationship} compared to T2I models (especially SDXL~\cite{podell2023sdxl}), which highlights the need to enhance compositionality (\ie, correctly composing multiple objects in the same frame). We believe possible solutions might be: \textit{1)} curating training data incorporating multiple objects with corresponding captions explicitly depicting this compositionality, or \textit{2)} adding intermediate spatial control modules or modalities during video synthesis. Furthermore, the disparity of the text encoders might also account for the performance gap. As T2I models leverage bigger (OpenCLIP ViT-H for SD2.1~\cite{rombach2022ldm}) or more sophisticated (CLIP ViT-L \& OpenCLIP ViT-G for SDXL~\cite{podell2023sdxl}) text encoders compared with T2V models (\eg, CLIP ViT-L alone for LaVie), more representative text embeddings could be featuring more accurate object composition comprehension.
\section{Conclusion}

With the growing focus on video generation, comprehensive evaluation of these models is essential to assess current advancements and guide future research.
In this work, we take the first step forward and propose \textbf{\textit{VBench}}, a comprehensive benchmark suite for evaluating video generation models. 
With its \textit{multi-dimensional, human-aligned, and insight-rich} properties, VBench could play vital roles for evaluating future video generation models and inspiring further advancements in video generation.
We believe that VBench is a significant contribution to the video generation and evaluation community.

\noindent \textbf{Limitations and Future Work}. 
We plan to expand VBench to include more models when they become available and extend the evaluations to additional video generation tasks, like image-to-video.

\noindent\textbf{Potential Negative Societal Impacts}.
We also recognize the importance of considering ethical aspects in future iterations of VBench. While VBench currently does not assess safety and equality dimensions, we urge users to exercise caution with open-sourced video generation models.

\smallTitle{Acknowledgement.} We would like to thank Shangchen Zhou, Jianyi Wang, and Ruicheng Feng for their helpful suggestions.

{
    \small
    \bibliographystyle{ieeenat_fullname}
    \bibliography{main}

\begin{thebibliography}{133}
\providecommand{\natexlab}[1]{#1}
\providecommand{\url}[1]{\texttt{#1}}
\expandafter\ifx\csname urlstyle\endcsname\relax
  \providecommand{\doi}[1]{doi: #1}\else
  \providecommand{\doi}{doi: \begingroup \urlstyle{rm}\Url}\fi

\bibitem[Gen(2023)]{Gen2}
Gen-2.
\newblock Accessed September 25, 2023 [Online] \url{https://research.runwayml.com/gen2}, 2023.

\bibitem[mor(2023)]{morphstudio}
Morph studio.
\newblock Accessed September 25, 2023 [Online] \url{https://www.morphstudio.com/}, 2023.

\bibitem[pik(2023)]{pikalab}
Pika labs.
\newblock Accessed September 25, 2023 [Online] \url{https://www.pika.art/}, 2023.

\bibitem[zer(2023)]{zeroscope}
Zeroscope-xl.
\newblock Accessed September 25, 2023 [Online] \url{https://huggingface.co/cerspense/zeroscope_v2_XL}, 2023.

\bibitem[Bain et~al.(2021)Bain, Nagrani, Varol, and Zisserman]{bain2021frozen}
Max Bain, Arsha Nagrani, G{\"u}l Varol, and Andrew Zisserman.
\newblock Frozen in time: A joint video and image encoder for end-to-end retrieval.
\newblock In \emph{ICCV}, 2021.

\bibitem[Bakr et~al.(2023)Bakr, Sun, Shen, Khan, Li, and Elhoseiny]{bakr2023hrsbench}
Eslam~Mohamed Bakr, Pengzhan Sun, Xiaoqian Shen, Faizan~Farooq Khan, Li~Erran Li, and Mohamed Elhoseiny.
\newblock Hrs-bench: Holistic, reliable and scalable benchmark for text-to-image models, 2023.

\bibitem[Basu et~al.(2023)Basu, Saberi, Bhardwaj, Chegini, Massiceti, Sanjabi, Hu, and Feizi]{basu2023editval}
Samyadeep Basu, Mehrdad Saberi, Shweta Bhardwaj, Atoosa~Malemir Chegini, Daniela Massiceti, Maziar Sanjabi, Shell~Xu Hu, and Soheil Feizi.
\newblock Editval: Benchmarking diffusion based text-guided image editing methods.
\newblock \emph{arXiv preprint arXiv:2310.02426}, 2023.

\bibitem[Blattmann et~al.(2023)Blattmann, Rombach, Ling, Dockhorn, Kim, Fidler, and Kreis]{blattmann2023videoldm}
Andreas Blattmann, Robin Rombach, Huan Ling, Tim Dockhorn, Seung~Wook Kim, Sanja Fidler, and Karsten Kreis.
\newblock Align your latents: High-resolution video synthesis with latent diffusion models.
\newblock In \emph{CVPR}, 2023.

\bibitem[Brock et~al.(2018)Brock, Donahue, and Simonyan]{brock2018biggan}
Andrew Brock, Jeff Donahue, and Karen Simonyan.
\newblock Large scale {GAN} training for high fidelity natural image synthesis.
\newblock \emph{arXiv preprint arXiv:1809.11096}, 2018.

\bibitem[Caron et~al.(2021)Caron, Touvron, Misra, J\'egou, Mairal, Bojanowski, and Joulin]{caron2021emerging}
Mathilde Caron, Hugo Touvron, Ishan Misra, Herv\'e J\'egou, Julien Mairal, Piotr Bojanowski, and Armand Joulin.
\newblock Emerging properties in self-supervised vision transformers.
\newblock In \emph{ICCV}, 2021.

\bibitem[Ceylan et~al.(2023)Ceylan, Huang, and Mitra]{ceylan2023pix2video}
Duygu Ceylan, Chun-Hao~P Huang, and Niloy~J Mitra.
\newblock Pix2video: Video editing using image diffusion.
\newblock In \emph{ICCV}, 2023.

\bibitem[Chai et~al.(2023)Chai, Guo, Wang, and Lu]{chai2023stablevideo}
Wenhao Chai, Xun Guo, Gaoang Wang, and Yan Lu.
\newblock Stablevideo: Text-driven consistency-aware diffusion video editing.
\newblock \emph{arXiv preprint arXiv:2308.09592}, 2023.

\bibitem[Chen et~al.(2023{\natexlab{a}})Chen, Xia, He, Zhang, Cun, Yang, Xing, Liu, Chen, Wang, Weng, and Shan]{chen2023videocrafter1}
Haoxin Chen, Menghan Xia, Yingqing He, Yong Zhang, Xiaodong Cun, Shaoshu Yang, Jinbo Xing, Yaofang Liu, Qifeng Chen, Xintao Wang, Chao Weng, and Ying Shan.
\newblock Videocrafter1: Open diffusion models for high-quality video generation.
\newblock \emph{arXiv preprint arXiv:2310.19512}, 2023{\natexlab{a}}.

\bibitem[Chen et~al.(2023{\natexlab{b}})Chen, Lin, Tseng, Lin, and Yang]{chen2023mcdiff}
Tsai-Shien Chen, Chieh~Hubert Lin, Hung-Yu Tseng, Tsung-Yi Lin, and Ming-Hsuan Yang.
\newblock Motion-conditioned diffusion model for controllable video synthesis.
\newblock \emph{arXiv preprint arXiv:2304.14404}, 2023{\natexlab{b}}.

\bibitem[Chen et~al.(2023{\natexlab{c}})Chen, Wu, Xie, Wu, Li, Xia, Xiao, and Lin]{chen2023controlavideo}
Weifeng Chen, Jie Wu, Pan Xie, Hefeng Wu, Jiashi Li, Xin Xia, Xuefeng Xiao, and Liang Lin.
\newblock Control-a-video: Controllable text-to-video generation with diffusion models, 2023{\natexlab{c}}.

\bibitem[Chen et~al.(2023{\natexlab{d}})Chen, Wang, Zhang, Zhuang, Ma, Yu, Wang, Lin, Qiao, and Liu]{chen2023seine}
Xinyuan Chen, Yaohui Wang, Lingjun Zhang, Shaobin Zhuang, Xin Ma, Jiashuo Yu, Yali Wang, Dahua Lin, Yu Qiao, and Ziwei Liu.
\newblock Seine: Short-to-long video diffusion model for generative transition and prediction.
\newblock \emph{arXiv preprint arXiv:2310.20700}, 2023{\natexlab{d}}.

\bibitem[Chu et~al.(2023)Chu, Lin, and Chen]{chu2023video}
Ernie Chu, Shuo-Yen Lin, and Jun-Cheng Chen.
\newblock Video controlnet: Towards temporally consistent synthetic-to-real video translation using conditional image diffusion models, 2023.

\bibitem[Couairon et~al.(2023)Couairon, Rambour, Haugeard, and Thome]{couairon2023videdit}
Paul Couairon, Cl{\'e}ment Rambour, Jean-Emmanuel Haugeard, and Nicolas Thome.
\newblock Videdit: Zero-shot and spatially aware text-driven video editing.
\newblock \emph{arXiv preprint arXiv:2306.08707}, 2023.

\bibitem[Dhariwal and Nichol(2021)]{dhariwal2021beatgan}
Prafulla Dhariwal and Alexander Nichol.
\newblock Diffusion models beat {GAN}s on image synthesis.
\newblock In \emph{NeurIPS}, 2021.

\bibitem[Ding et~al.(2022)Ding, Zheng, Hong, and Tang]{ding2022cogview2}
Ming Ding, Wendi Zheng, Wenyi Hong, and Jie Tang.
\newblock Cogview2: Faster and better text-to-image generation via hierarchical transformers.
\newblock In \emph{NeurIPS}, 2022.

\bibitem[Esser et~al.(2020)Esser, Rombach, and Ommer]{esser2020note}
Patrick Esser, Robin Rombach, and Bj{\"o}rn Ommer.
\newblock A note on data biases in generative models.
\newblock In \emph{NeurIPS Workshop}, 2020.

\bibitem[Esser et~al.(2021)Esser, Rombach, and Ommer]{esser2021vqgan}
Patrick Esser, Robin Rombach, and Bjorn Ommer.
\newblock Taming transformers for high-resolution image synthesis.
\newblock In \emph{CVPR}, 2021.

\bibitem[Esser et~al.(2023)Esser, Chiu, Atighehchian, Granskog, and Germanidis]{esser2023structure}
Patrick Esser, Johnathan Chiu, Parmida Atighehchian, Jonathan Granskog, and Anastasis Germanidis.
\newblock Structure and content-guided video synthesis with diffusion models, 2023.

\bibitem[Fang et~al.(2020)Fang, Zhu, Zeng, Ma, and Wang]{Fang2020spaq}
Yuming Fang, Hanwei Zhu, Yan Zeng, Kede Ma, and Zhou Wang.
\newblock Perceptual quality assessment of smartphone photography.
\newblock In \emph{CVPR}, 2020.

\bibitem[Fu et~al.(2022)Fu, Li, Jiang, Lin, Qian, Loy, Wu, and Liu]{fu2022stylegan}
Jianglin Fu, Shikai Li, Yuming Jiang, Kwan-Yee Lin, Chen Qian, Chen~Change Loy, Wayne Wu, and Ziwei Liu.
\newblock Stylegan-human: A data-centric odyssey of human generation.
\newblock In \emph{ECCV}, 2022.

\bibitem[Fu et~al.(2023{\natexlab{a}})Fu, Li, Jiang, Lin, Wu, and Liu]{fu2023unitedhuman}
Jianglin Fu, Shikai Li, Yuming Jiang, Kwan-Yee Lin, Wayne Wu, and Ziwei Liu.
\newblock Unitedhuman: Harnessing multi-source data for high-resolution human generation.
\newblock In \emph{ICCV}, 2023{\natexlab{a}}.

\bibitem[Fu et~al.(2023{\natexlab{b}})Fu, Yu, Zhang, Fu, Su, Wang, and Bell]{fu2023tell}
Tsu-Jui Fu, Licheng Yu, Ning Zhang, Cheng-Yang Fu, Jong-Chyi Su, William~Yang Wang, and Sean Bell.
\newblock Tell me what happened: Unifying text-guided video completion via multimodal masked video generation, 2023{\natexlab{b}}.

\bibitem[Ge et~al.(2023)Ge, Nah, Liu, Poon, Tao, Catanzaro, Jacobs, Huang, Liu, and Balaji]{ge2023pyoco}
Songwei Ge, Seungjun Nah, Guilin Liu, Tyler Poon, Andrew Tao, Bryan Catanzaro, David Jacobs, Jia-Bin Huang, Ming-Yu Liu, and Yogesh Balaji.
\newblock Preserve your own correlation: A noise prior for video diffusion models.
\newblock In \emph{ICCV}, 2023.

\bibitem[Geyer et~al.(2023)Geyer, Bar-Tal, Bagon, and Dekel]{tokenflow2023}
Michal Geyer, Omer Bar-Tal, Shai Bagon, and Tali Dekel.
\newblock Tokenflow: Consistent diffusion features for consistent video editing.
\newblock \emph{arXiv preprint arxiv:2307.10373}, 2023.

\bibitem[Goodfellow et~al.(2014)Goodfellow, Pouget-Abadie, Mirza, Xu, Warde-Farley, Ozair, Courville, and Bengio]{goodfellow2014gan}
Ian~J Goodfellow, Jean Pouget-Abadie, Mehdi Mirza, Bing Xu, David Warde-Farley, Sherjil Ozair, Aaron~C Courville, and Yoshua Bengio.
\newblock Generative adversarial nets.
\newblock In \emph{NeurIPS}, 2014.

\bibitem[Gu et~al.(2022)Gu, Chen, Bao, Wen, Zhang, Chen, Yuan, and Guo]{gu2022vqdiffusion}
Shuyang Gu, Dong Chen, Jianmin Bao, Fang Wen, Bo Zhang, Dongdong Chen, Lu Yuan, and Baining Guo.
\newblock Vector quantized diffusion model for text-to-image synthesis.
\newblock In \emph{CVPR}, 2022.

\bibitem[Gu et~al.(2023)Gu, Wen, Song, and Gao]{gu2023seer}
Xianfan Gu, Chuan Wen, Jiaming Song, and Yang Gao.
\newblock Seer: Language instructed video prediction with latent diffusion models.
\newblock \emph{arXiv preprint arXiv:2303.14897}, 2023.

\bibitem[Guo et~al.(2023)Guo, Yang, Rao, Wang, Qiao, Lin, and Dai]{guo2023animatediff}
Yuwei Guo, Ceyuan Yang, Anyi Rao, Yaohui Wang, Yu Qiao, Dahua Lin, and Bo Dai.
\newblock Animatediff: Animate your personalized text-to-image diffusion models without specific tuning.
\newblock \emph{arXiv preprint arXiv:2307.04725}, 2023.

\bibitem[Harvey et~al.(2022)Harvey, Naderiparizi, Masrani, Weilbach, and Wood]{harvey2022fdm}
William Harvey, Saeid Naderiparizi, Vaden Masrani, Christian Weilbach, and Frank Wood.
\newblock Flexible diffusion modeling of long videos.
\newblock \emph{arXiv preprint arXiv:2205.11495}, 2022.

\bibitem[He et~al.(2022)He, Yang, Zhang, Shan, and Chen]{he2022lvdm}
Yingqing He, Tianyu Yang, Yong Zhang, Ying Shan, and Qifeng Chen.
\newblock Latent video diffusion models for high-fidelity video generation with arbitrary lengths.
\newblock \emph{arXiv preprint arXiv:2211.13221}, 2022.

\bibitem[He et~al.(2023)He, Xia, Chen, Cun, Gong, Xing, Zhang, Wang, Weng, Shan, et~al.]{he2023animate}
Yingqing He, Menghan Xia, Haoxin Chen, Xiaodong Cun, Yuan Gong, Jinbo Xing, Yong Zhang, Xintao Wang, Chao Weng, Ying Shan, et~al.
\newblock Animate-a-story: Storytelling with retrieval-augmented video generation.
\newblock \emph{arXiv preprint arXiv:2307.06940}, 2023.

\bibitem[Heusel et~al.(2017)Heusel, Ramsauer, Unterthiner, Nessler, and Hochreiter]{heusel2017fid}
Martin Heusel, Hubert Ramsauer, Thomas Unterthiner, Bernhard Nessler, and Sepp Hochreiter.
\newblock {GANs} trained by a two time-scale update rule converge to a local nash equilibrium.
\newblock In \emph{NeurIPS}, 2017.

\bibitem[Ho et~al.(2020)Ho, Jain, and Abbeel]{ho2020ddpm}
Jonathan Ho, Ajay Jain, and Pieter Abbeel.
\newblock Denoising diffusion probabilistic models.
\newblock In \emph{NeurIPS}, 2020.

\bibitem[Ho et~al.(2022{\natexlab{a}})Ho, Chan, Saharia, Whang, Gao, Gritsenko, Kingma, Poole, Norouzi, Fleet, et~al.]{ho2022imagenvideo}
Jonathan Ho, William Chan, Chitwan Saharia, Jay Whang, Ruiqi Gao, Alexey Gritsenko, Diederik~P Kingma, Ben Poole, Mohammad Norouzi, David~J Fleet, et~al.
\newblock Imagen video: High definition video generation with diffusion models.
\newblock \emph{arXiv preprint arXiv:2210.02303}, 2022{\natexlab{a}}.

\bibitem[Ho et~al.(2022{\natexlab{b}})Ho, Salimans, Gritsenko, Chan, Norouzi, and Fleet]{ho2022videoDM}
Jonathan Ho, Tim Salimans, Alexey Gritsenko, William Chan, Mohammad Norouzi, and David~J Fleet.
\newblock Video diffusion models.
\newblock \emph{arXiv preprint arXiv:2204.03458}, 2022{\natexlab{b}}.

\bibitem[Hong et~al.(2022)Hong, Ding, Zheng, Liu, and Tang]{hong2022cogvideo}
Wenyi Hong, Ming Ding, Wendi Zheng, Xinghan Liu, and Jie Tang.
\newblock {CogVideo}: Large-scale pretraining for text-to-video generation via transformers.
\newblock \emph{arXiv preprint arXiv:2205.15868}, 2022.

\bibitem[Hu and Xu(2023)]{hu2023videocontrolnet}
Zhihao Hu and Dong Xu.
\newblock Videocontrolnet: A motion-guided video-to-video translation framework by using diffusion model with controlnet.
\newblock \emph{arXiv preprint arXiv:2307.14073}, 2023.

\bibitem[Huang et~al.(2023{\natexlab{a}})Huang, Sigal, Yi, Wang, and Lee]{huang2023inve}
Jiahui Huang, Leonid Sigal, Kwang~Moo Yi, Oliver Wang, and Joon-Young Lee.
\newblock Inve: Interactive neural video editing, 2023{\natexlab{a}}.

\bibitem[Huang et~al.(2023{\natexlab{b}})Huang, Sun, Xie, Li, and Liu]{huang2023t2icompbench}
Kaiyi Huang, Kaiyue Sun, Enze Xie, Zhenguo Li, and Xihui Liu.
\newblock T2i-compbench: A comprehensive benchmark for open-world compositional text-to-image generation.
\newblock \emph{arXiv preprint arXiv: 2307.06350}, 2023{\natexlab{b}}.

\bibitem[Huang et~al.(2023{\natexlab{c}})Huang, Zhang, Ma, Tian, Feng, Zhang, Li, Guo, and Zhang]{huang2023tag2text}
Xinyu Huang, Youcai Zhang, Jinyu Ma, Weiwei Tian, Rui Feng, Yuejie Zhang, Yaqian Li, Yandong Guo, and Lei Zhang.
\newblock Tag2text: Guiding vision-language model via image tagging.
\newblock \emph{arXiv preprint arXiv:2303.05657}, 2023{\natexlab{c}}.

\bibitem[Huang et~al.(2023{\natexlab{d}})Huang, Chan, Jiang, and Liu]{huang2023collaborative}
Ziqi Huang, Kelvin~C.K. Chan, Yuming Jiang, and Ziwei Liu.
\newblock Collaborative diffusion for multi-modal face generation and editing.
\newblock In \emph{CVPR}, 2023{\natexlab{d}}.

\bibitem[Huang et~al.(2023{\natexlab{e}})Huang, Wu, Jiang, Chan, and Liu]{huang2023reversion}
Ziqi Huang, Tianxing Wu, Yuming Jiang, Kelvin~C.K. Chan, and Ziwei Liu.
\newblock {ReVersion}: Diffusion-based relation inversion from images.
\newblock \emph{arXiv preprint arXiv:2303.13495}, 2023{\natexlab{e}}.

\bibitem[Jiang et~al.(2021)Jiang, Huang, Pan, Loy, and Liu]{jiang2021talktoedit}
Yuming Jiang, Ziqi Huang, Xingang Pan, Chen~Change Loy, and Ziwei Liu.
\newblock {Talk-to-Edit}: Fine-grained facial editing via dialog.
\newblock In \emph{ICCV}, 2021.

\bibitem[Jiang et~al.(2022)Jiang, Yang, Qju, Wu, Loy, and Liu]{jiang2022text2human}
Yuming Jiang, Shuai Yang, Haonan Qju, Wayne Wu, Chen~Change Loy, and Ziwei Liu.
\newblock Text2human: Text-driven controllable human image generation.
\newblock \emph{ACM TOG}, 2022.

\bibitem[Jiang et~al.(2023)Jiang, Yang, Koh, Wu, Loy, and Liu]{jiang2023text2performer}
Yuming Jiang, Shuai Yang, Tong~Liang Koh, Wayne Wu, Chen~Change Loy, and Ziwei Liu.
\newblock {Text2Performer}: Text-driven human video generation.
\newblock In \emph{ICCV}, 2023.

\bibitem[Karras et~al.(2023)Karras, Holynski, Wang, and Kemelmacher-Shlizerman]{dreampose_2023}
Johanna Karras, Aleksander Holynski, Ting-Chun Wang, and Ira Kemelmacher-Shlizerman.
\newblock Dreampose: Fashion image-to-video synthesis via stable diffusion.
\newblock \emph{arXiv preprint arXiv:2304.06025}, 2023.

\bibitem[Karras et~al.(2018)Karras, Aila, Laine, and Lehtinen]{karras2018pggan}
Tero Karras, Timo Aila, Samuli Laine, and Jaakko Lehtinen.
\newblock Progressive growing of {GAN}s for improved quality, stability, and variation.
\newblock In \emph{ICLR}, 2018.

\bibitem[Karras et~al.(2019)Karras, Laine, and Aila]{karras2019stylegan1}
Tero Karras, Samuli Laine, and Timo Aila.
\newblock A style-based generator architecture for generative adversarial networks.
\newblock In \emph{CVPR}, 2019.

\bibitem[Karras et~al.(2020)Karras, Laine, Aittala, Hellsten, Lehtinen, and Aila]{karras2020stylegan2}
Tero Karras, Samuli Laine, Miika Aittala, Janne Hellsten, Jaakko Lehtinen, and Timo Aila.
\newblock Analyzing and improving the image quality of {StyleGAN}.
\newblock In \emph{CVPR}, 2020.

\bibitem[Karras et~al.(2021)Karras, Aittala, Laine, H{\"a}rk{\"o}nen, Hellsten, Lehtinen, and Aila]{karras2021stylegan3}
Tero Karras, Miika Aittala, Samuli Laine, Erik H{\"a}rk{\"o}nen, Janne Hellsten, Jaakko Lehtinen, and Timo Aila.
\newblock Alias-free generative adversarial networks.
\newblock In \emph{NeurIPS}, 2021.

\bibitem[Kay et~al.(2017)Kay, Carreira, Simonyan, Zhang, Hillier, Vijayanarasimhan, Viola, Green, Back, Natsev, et~al.]{kay2017kinetics}
Will Kay, Joao Carreira, Karen Simonyan, Brian Zhang, Chloe Hillier, Sudheendra Vijayanarasimhan, Fabio Viola, Tim Green, Trevor Back, Paul Natsev, et~al.
\newblock The kinetics human action video dataset.
\newblock \emph{arXiv preprint arXiv:1705.06950}, 2017.

\bibitem[Ke et~al.(2021)Ke, Wang, Wang, Milanfar, and Yang]{Ke2021MUSIQ}
Junjie Ke, Qifei Wang, Yilin Wang, Peyman Milanfar, and Feng Yang.
\newblock {MUSIQ:} multi-scale image quality transformer.
\newblock \emph{CoRR}, abs/2108.05997, 2021.

\bibitem[Khachatryan et~al.(2023)Khachatryan, Movsisyan, Tadevosyan, Henschel, Wang, Navasardyan, and Shi]{khachatryan2023text2videozero}
Levon Khachatryan, Andranik Movsisyan, Vahram Tadevosyan, Roberto Henschel, Zhangyang Wang, Shant Navasardyan, and Humphrey Shi.
\newblock Text2video-zero: Text-to-image diffusion models are zero-shot video generators.
\newblock \emph{arXiv preprint arXiv:2303.13439}, 2023.

\bibitem[Kingma and Welling(2013)]{kingma2013vae}
Diederik~P Kingma and Max Welling.
\newblock Auto-encoding variational bayes.
\newblock \emph{arXiv preprint arXiv:1312.6114}, 2013.

\bibitem[LAION-AI(2022)]{LAIONaes}
LAION-AI.
\newblock aesthetic-predictor.
\newblock \url{https://github.com/LAION-AI/aesthetic-predictor}, 2022.

\bibitem[Lee et~al.(2023{\natexlab{a}})Lee, Yasunaga, Meng, Mai, Park, Gupta, Zhang, Narayanan, Teufel, Bellagente, et~al.]{lee2023holistic}
Tony Lee, Michihiro Yasunaga, Chenlin Meng, Yifan Mai, Joon~Sung Park, Agrim Gupta, Yunzhi Zhang, Deepak Narayanan, Hannah~Benita Teufel, Marco Bellagente, et~al.
\newblock Holistic evaluation of text-to-image models.
\newblock \emph{arXiv preprint arXiv:2311.04287}, 2023{\natexlab{a}}.

\bibitem[Lee et~al.(2023{\natexlab{b}})Lee, Jang, Chen, Qiu, and Huang]{lee2023textvideoedit}
Yao-Chih Lee, Ji-Ze Genevieve~Jang Jang, Yi-Ting Chen, Elizabeth Qiu, and Jia-Bin Huang.
\newblock Shape-aware text-driven layered video editing demo.
\newblock \emph{arXiv preprint arXiv:2301.13173}, 2023{\natexlab{b}}.

\bibitem[Li et~al.(2019)Li, Jiang, and Jiang]{vsfa}
Dingquan Li, Tingting Jiang, and Ming Jiang.
\newblock Quality assessment of in-the-wild videos.
\newblock In \emph{ACM MM}, 2019.

\bibitem[Li et~al.(2023{\natexlab{a}})Li, He, Wang, Li, Wang, Luo, Wang, Wang, and Qiao]{2023videochat}
Kunchang Li, Yinan He, Yi Wang, Yizhuo Li, Wenhai Wang, Ping Luo, Yali Wang, Limin Wang, and Yu Qiao.
\newblock Videochat: Chat-centric video understanding.
\newblock \emph{arXiv preprint arXiv:2305.06355}, 2023{\natexlab{a}}.

\bibitem[Li et~al.(2023{\natexlab{b}})Li, Wang, Li, Wang, He, Wang, and Qiao]{li2023unmasked}
Kunchang Li, Yali Wang, Yizhuo Li, Yi Wang, Yinan He, Limin Wang, and Yu Qiao.
\newblock Unmasked teacher: Towards training-efficient video foundation models.
\newblock \emph{arXiv preprint arXiv:2303.16058}, 2023{\natexlab{b}}.

\bibitem[Li et~al.(2023{\natexlab{c}})Li, Zhu, Han, Hou, Guo, and Cheng]{licvpr23amt}
Zhen Li, Zuo-Liang Zhu, Ling-Hao Han, Qibin Hou, Chun-Le Guo, and Ming-Ming Cheng.
\newblock Amt: All-pairs multi-field transforms for efficient frame interpolation.
\newblock In \emph{CVPR}, 2023{\natexlab{c}}.

\bibitem[Liew et~al.(2023)Liew, Yan, Zhang, Xu, and Feng]{liew2023magicedit}
Jun~Hao Liew, Hanshu Yan, Jianfeng Zhang, Zhongcong Xu, and Jiashi Feng.
\newblock Magicedit: High-fidelity and temporally coherent video editing.
\newblock \emph{arXiv preprint arXiv:2308.14749}, 2023.

\bibitem[Lin et~al.(2014)Lin, Maire, Belongie, Hays, Perona, Ramanan, Doll{\'a}r, and Zitnick]{lin2014mscoco}
Tsung-Yi Lin, Michael Maire, Serge Belongie, James Hays, Pietro Perona, Deva Ramanan, Piotr Doll{\'a}r, and C~Lawrence Zitnick.
\newblock Microsoft coco: Common objects in context.
\newblock In \emph{ECCV}, 2014.

\bibitem[Liu et~al.(2023{\natexlab{a}})Liu, Zhang, Li, Lin, and Jia]{liu2023videop2p}
Shaoteng Liu, Yuechen Zhang, Wenbo Li, Zhe Lin, and Jiaya Jia.
\newblock Video-p2p: Video editing with cross-attention control, 2023{\natexlab{a}}.

\bibitem[Liu et~al.(2023{\natexlab{b}})Liu, Cun, Liu, Wang, Zhang, Chen, Liu, Zeng, Chan, and Shan]{liu2023evalcrafter}
Yaofang Liu, Xiaodong Cun, Xuebo Liu, Xintao Wang, Yong Zhang, Haoxin Chen, Yang Liu, Tieyong Zeng, Raymond Chan, and Ying Shan.
\newblock Evalcrafter: Benchmarking and evaluating large video generation models.
\newblock \emph{arXiv preprint arXiv:2310.11440}, 2023{\natexlab{b}}.

\bibitem[Liu et~al.(2023{\natexlab{c}})Liu, Li, Ren, Gao, Li, Chen, Sun, and Hou]{liu2023fetv}
Yuanxin Liu, Lei Li, Shuhuai Ren, Rundong Gao, Shicheng Li, Sishuo Chen, Xu Sun, and Lu Hou.
\newblock Fetv: A benchmark for fine-grained evaluation of open-domain text-to-video generation.
\newblock In \emph{NeurIPS}, 2023{\natexlab{c}}.

\bibitem[Luo et~al.(2023)Luo, Chen, Zhang, Huang, Wang, Shen, Zhao, Zhou, and Tan]{luo2023videofusion}
Zhengxiong Luo, Dayou Chen, Yingya Zhang, Yan Huang, Liang Wang, Yujun Shen, Deli Zhao, Jingren Zhou, and Tieniu Tan.
\newblock {VideoFusion}: Decomposed diffusion models for high-quality video generation.
\newblock In \emph{CVPR}, 2023.

\bibitem[Ma et~al.(2023)Ma, He, Cun, Wang, Shan, Li, and Chen]{ma2023follow}
Yue Ma, Yingqing He, Xiaodong Cun, Xintao Wang, Ying Shan, Xiu Li, and Qifeng Chen.
\newblock Follow your pose: Pose-guided text-to-video generation using pose-free videos.
\newblock \emph{arXiv preprint arXiv:2304.01186}, 2023.

\bibitem[Mirza and Osindero(2014)]{mirza2014cgan}
Mehdi Mirza and Simon Osindero.
\newblock Conditional generative adversarial nets.
\newblock \emph{arXiv preprint arXiv:1411.1784}, 2014.

\bibitem[Moing et~al.(2023)Moing, Ponce, and Schmid]{lemoing2022waldo}
Guillaume~Le Moing, Jean Ponce, and Cordelia Schmid.
\newblock {WALDO}: Future video synthesis using object layer decomposition and parametric flow prediction.
\newblock In \emph{ICCV}, 2023.

\bibitem[Molad et~al.(2023)Molad, Horwitz, Valevski, Acha, Matias, Pritch, Leviathan, and Hoshen]{molad2023dreamix}
Eyal Molad, Eliahu Horwitz, Dani Valevski, Alex~Rav Acha, Yossi Matias, Yael Pritch, Yaniv Leviathan, and Yedid Hoshen.
\newblock Dreamix: Video diffusion models are general video editors.
\newblock \emph{arXiv preprint arXiv:2302.01329}, 2023.

\bibitem[Ni et~al.(2023)Ni, Shi, Li, Huang, and Min]{ni2023conditional}
Haomiao Ni, Changhao Shi, Kai Li, Sharon~X Huang, and Martin~Renqiang Min.
\newblock Conditional image-to-video generation with latent flow diffusion models.
\newblock In \emph{CVPR}, 2023.

\bibitem[Nichol et~al.(2021)Nichol, Dhariwal, Ramesh, Shyam, Mishkin, McGrew, Sutskever, and Chen]{nichol2021glide}
Alex Nichol, Prafulla Dhariwal, Aditya Ramesh, Pranav Shyam, Pamela Mishkin, Bob McGrew, Ilya Sutskever, and Mark Chen.
\newblock {GLIDE}: Towards photorealistic image generation and editing with text-guided diffusion models.
\newblock \emph{arXiv preprint arXiv:2112.10741}, 2021.

\bibitem[Otani et~al.(2023)Otani, Togashi, Sawai, Ishigami, Nakashima, Rahtu, Heikkil{\"a}, and Satoh]{otani2023toward}
Mayu Otani, Riku Togashi, Yu Sawai, Ryosuke Ishigami, Yuta Nakashima, Esa Rahtu, Janne Heikkil{\"a}, and Shin’ichi Satoh.
\newblock Toward verifiable and reproducible human evaluation for text-to-image generation.
\newblock In \emph{CVPR}, 2023.

\bibitem[Ouyang et~al.(2023)Ouyang, Wang, Xiao, Bai, Zhang, Zheng, Zhou, Chen, and Shen]{ouyang2023codef}
Hao Ouyang, Qiuyu Wang, Yuxi Xiao, Qingyan Bai, Juntao Zhang, Kecheng Zheng, Xiaowei Zhou, Qifeng Chen, and Yujun Shen.
\newblock Codef: Content deformation fields for temporally consistent video processing.
\newblock \emph{arXiv preprint arXiv:2308.07926}, 2023.

\bibitem[Podell et~al.(2023)Podell, English, Lacey, Blattmann, Dockhorn, M{\"u}ller, Penna, and Rombach]{podell2023sdxl}
Dustin Podell, Zion English, Kyle Lacey, Andreas Blattmann, Tim Dockhorn, Jonas M{\"u}ller, Joe Penna, and Robin Rombach.
\newblock Sdxl: Improving latent diffusion models for high-resolution image synthesis.
\newblock \emph{arXiv preprint arXiv:2307.01952}, 2023.

\bibitem[Qi et~al.(2023)Qi, Cun, Zhang, Lei, Wang, Shan, and Chen]{qi2023fatezero}
Chenyang Qi, Xiaodong Cun, Yong Zhang, Chenyang Lei, Xintao Wang, Ying Shan, and Qifeng Chen.
\newblock Fatezero: Fusing attentions for zero-shot text-based video editing.
\newblock \emph{arXiv preprint arXiv:2303.09535}, 2023.

\bibitem[Radford et~al.(2021)Radford, Kim, Hallacy, Ramesh, Goh, Agarwal, Sastry, Askell, Mishkin, Clark, et~al.]{radford2021clip}
Alec Radford, Jong~Wook Kim, Chris Hallacy, Aditya Ramesh, Gabriel Goh, Sandhini Agarwal, Girish Sastry, Amanda Askell, Pamela Mishkin, Jack Clark, et~al.
\newblock Learning transferable visual models from natural language supervision.
\newblock In \emph{ICML}, 2021.

\bibitem[Rombach et~al.(2022)Rombach, Blattmann, Lorenz, Esser, and Ommer]{rombach2022ldm}
Robin Rombach, Andreas Blattmann, Dominik Lorenz, Patrick Esser, and Bj{\"o}rn Ommer.
\newblock High-resolution image synthesis with latent diffusion models.
\newblock In \emph{CVPR}, 2022.

\bibitem[Ruiz et~al.(2023)Ruiz, Li, Jampani, Pritch, Rubinstein, and Aberman]{ruiz2022dreambooth}
Nataniel Ruiz, Yuanzhen Li, Varun Jampani, Yael Pritch, Michael Rubinstein, and Kfir Aberman.
\newblock Dreambooth: Fine tuning text-to-image diffusion models for subject-driven generation.
\newblock In \emph{CVPR}, 2023.

\bibitem[Saharia et~al.(2022)Saharia, Chan, Saxena, Li, Whang, Denton, Ghasemipour, Ayan, Mahdavi, Lopes, et~al.]{saharia2022imagen}
Chitwan Saharia, William Chan, Saurabh Saxena, Lala Li, Jay Whang, Emily Denton, Seyed Kamyar~Seyed Ghasemipour, Burcu~Karagol Ayan, S~Sara Mahdavi, Rapha~Gontijo Lopes, et~al.
\newblock Photorealistic text-to-image diffusion models with deep language understanding.
\newblock \emph{arXiv preprint arXiv:2205.11487}, 2022.

\bibitem[Salimans et~al.(2016)Salimans, Goodfellow, Zaremba, Cheung, Radford, Chen, and Chen]{salimans2016inceptionscore}
Tim Salimans, Ian Goodfellow, Wojciech Zaremba, Vicki Cheung, Alec Radford, Xi Chen, and Xi Chen.
\newblock Improved techniques for training gans.
\newblock In \emph{NeurIPS}, 2016.

\bibitem[Singer et~al.(2022)Singer, Polyak, Hayes, Yin, An, Zhang, Hu, Yang, Ashual, Gafni, et~al.]{singer2022makeavideo}
Uriel Singer, Adam Polyak, Thomas Hayes, Xi Yin, Jie An, Songyang Zhang, Qiyuan Hu, Harry Yang, Oron Ashual, Oran Gafni, et~al.
\newblock Make-a-video: Text-to-video generation without text-video data.
\newblock \emph{arXiv preprint arXiv:2209.14792}, 2022.

\bibitem[Sohl-Dickstein et~al.(2015)Sohl-Dickstein, Weiss, Maheswaranathan, and Ganguli]{sohl2015deep}
Jascha Sohl-Dickstein, Eric Weiss, Niru Maheswaranathan, and Surya Ganguli.
\newblock Deep unsupervised learning using nonequilibrium thermodynamics.
\newblock In \emph{ICML}, 2015.

\bibitem[Song et~al.(2022)Song, Chen, Zhu, and Jiang]{song2022textdriven}
Xue Song, Jingjing Chen, Bin Zhu, and Yu-Gang Jiang.
\newblock Text-driven video prediction.
\newblock \emph{arXiv preprint arXiv:2210.02872}, 2022.

\bibitem[Song et~al.(2021)Song, Sohl-Dickstein, Kingma, Kumar, Ermon, and Poole]{song2020score}
Yang Song, Jascha Sohl-Dickstein, Diederik~P Kingma, Abhishek Kumar, Stefano Ermon, and Ben Poole.
\newblock Score-based generative modeling through stochastic differential equations.
\newblock In \emph{ICLR}, 2021.

\bibitem[Soomro et~al.(2012)Soomro, Zamir, and Shah]{soomro2012ucf101}
Khurram Soomro, Amir~Roshan Zamir, and Mubarak Shah.
\newblock Ucf101: A dataset of 101 human actions classes from videos in the wild.
\newblock \emph{arXiv preprint arXiv:1212.0402}, 2012.

\bibitem[Teed and Deng(2020)]{teed2020raft}
Zachary Teed and Jia Deng.
\newblock Raft: Recurrent all-pairs field transforms for optical flow.
\newblock In \emph{ECCV}, 2020.

\bibitem[Tu et~al.(2021)Tu, Wang, Birkbeck, Adsumilli, and Bovik]{videval}
Zhengzhong Tu, Yilin Wang, Neil Birkbeck, Balu Adsumilli, and Alan~C. Bovik.
\newblock Ugc-vqa: Benchmarking blind video quality assessment for user generated content.
\newblock \emph{IEEE TIP}, 30:\penalty0 4449--4464, 2021.

\bibitem[Unterthiner et~al.(2018)Unterthiner, van Steenkiste, Kurach, Marinier, Michalski, and Gelly]{unterthiner2018fvd}
Thomas Unterthiner, Sjoerd van Steenkiste, Karol Kurach, Raphael Marinier, Marcin Michalski, and Sylvain Gelly.
\newblock Towards accurate generative models of video: A new metric \& challenges.
\newblock \emph{arXiv preprint arXiv:1812.01717}, 2018.

\bibitem[Unterthiner et~al.(2019)Unterthiner, van Steenkiste, Kurach, Marinier, Michalski, and Gelly]{unterthiner2019fvd}
Thomas Unterthiner, Sjoerd van Steenkiste, Karol Kurach, Rapha{\"e}l Marinier, Marcin Michalski, and Sylvain Gelly.
\newblock {FVD}: A new metric for video generation.
\newblock In \emph{ICLRW}, 2019.

\bibitem[Van Den~Oord et~al.(2017)Van Den~Oord, Vinyals, et~al.]{van2017vqvae}
Aaron Van Den~Oord, Oriol Vinyals, et~al.
\newblock Neural discrete representation learning.
\newblock In \emph{NeurIPS}, 2017.

\bibitem[Wang et~al.(2023{\natexlab{a}})Wang, Yuan, Chen, Zhang, Wang, and Zhang]{wang2023modelscope}
Jiuniu Wang, Hangjie Yuan, Dayou Chen, Yingya Zhang, Xiang Wang, and Shiwei Zhang.
\newblock Modelscope text-to-video technical report.
\newblock \emph{arXiv preprint arXiv:2308.06571}, 2023{\natexlab{a}}.

\bibitem[Wang et~al.(2022)Wang, Saharia, Montgomery, Pont-Tuset, Noy, Pellegrini, Onoe, Laszlo, Fleet, Soricut, et~al.]{wang2022imagenedit}
Su Wang, Chitwan Saharia, Ceslee Montgomery, Jordi Pont-Tuset, Shai Noy, Stefano Pellegrini, Yasumasa Onoe, Sarah Laszlo, David~J Fleet, Radu Soricut, et~al.
\newblock Imagen editor and {EditBench}: Advancing and evaluating text-guided image inpainting.
\newblock \emph{arXiv preprint arXiv:2212.06909}, 2022.

\bibitem[Wang et~al.(2023{\natexlab{b}})Wang, Li, Lin, Lin, Yang, Zhang, Liu, and Wang]{wang2023disco}
Tan Wang, Linjie Li, Kevin Lin, Chung-Ching Lin, Zhengyuan Yang, Hanwang Zhang, Zicheng Liu, and Lijuan Wang.
\newblock Disco: Disentangled control for referring human dance generation in real world.
\newblock \emph{arXiv preprint arXiv:2307.00040}, 2023{\natexlab{b}}.

\bibitem[Wang et~al.(2023{\natexlab{c}})Wang, Xie, Liu, Chen, Cao, Wang, and Shen]{vid2vid-zero}
Wen Wang, kangyang Xie, Zide Liu, Hao Chen, Yue Cao, Xinlong Wang, and Chunhua Shen.
\newblock Zero-shot video editing using off-the-shelf image diffusion models.
\newblock \emph{arXiv preprint arXiv:2303.17599}, 2023{\natexlab{c}}.

\bibitem[Wang et~al.(2023{\natexlab{d}})Wang, Wu, Yin, Ni, Wang, Li, Yang, Yang, Wang, Liu, Fang, and Duan]{wang2023learning}
Xiaodong Wang, Chenfei Wu, Shengming Yin, Minheng Ni, Jianfeng Wang, Linjie Li, Zhengyuan Yang, Fan Yang, Lijuan Wang, Zicheng Liu, Yuejian Fang, and Nan Duan.
\newblock Learning 3d photography videos via self-supervised diffusion on single images, 2023{\natexlab{d}}.

\bibitem[Wang et~al.(2023{\natexlab{e}})Wang, Yuan, Zhang, Chen, Wang, Zhang, Shen, Zhao, and Zhou]{2023videocomposer}
Xiang Wang, Hangjie Yuan, Shiwei Zhang, Dayou Chen, Jiuniu Wang, Yingya Zhang, Yujun Shen, Deli Zhao, and Jingren Zhou.
\newblock Videocomposer: Compositional video synthesis with motion controllability.
\newblock \emph{arXiv preprint arXiv:2306.02018}, 2023{\natexlab{e}}.

\bibitem[Wang et~al.(2023{\natexlab{f}})Wang, Chen, Ma, Zhou, Huang, Wang, Yang, He, Yu, Yang, et~al.]{wang2023lavie}
Yaohui Wang, Xinyuan Chen, Xin Ma, Shangchen Zhou, Ziqi Huang, Yi Wang, Ceyuan Yang, Yinan He, Jiashuo Yu, Peiqing Yang, et~al.
\newblock Lavie: High-quality video generation with cascaded latent diffusion models.
\newblock \emph{arXiv preprint arXiv:2309.15103}, 2023{\natexlab{f}}.

\bibitem[Wang et~al.(2023{\natexlab{g}})Wang, He, Li, Li, Yu, Ma, Chen, Wang, Luo, Liu, Wang, Wang, and Qiao]{wang2023internvid}
Yi Wang, Yinan He, Yizhuo Li, Kunchang Li, Jiashuo Yu, Xin Ma, Xinyuan Chen, Yaohui Wang, Ping Luo, Ziwei Liu, Yali Wang, Limin Wang, and Yu Qiao.
\newblock Internvid: A large-scale video-text dataset for multimodal understanding and generation.
\newblock \emph{arXiv preprint arXiv:2307.06942}, 2023{\natexlab{g}}.

\bibitem[Wang et~al.(2023{\natexlab{h}})Wang, Ma, Chen, Dantcheva, Dai, and Qiao]{wang2023leo}
Yaohui Wang, Xin Ma, Xinyuan Chen, Antitza Dantcheva, Bo Dai, and Yu Qiao.
\newblock Leo: Generative latent image animator for human video synthesis.
\newblock \emph{arXiv preprint arXiv:2305.03989}, 2023{\natexlab{h}}.

\bibitem[Wu et~al.(2022{\natexlab{a}})Wu, Liang, Ji, Yang, Fang, Jiang, and Duan]{wu2021nuwa}
Chenfei Wu, Jian Liang, Lei Ji, Fan Yang, Yuejian Fang, Daxin Jiang, and Nan Duan.
\newblock N{\"u}wa: Visual synthesis pre-training for neural visual world creation.
\newblock In \emph{ECCV}, 2022{\natexlab{a}}.

\bibitem[Wu et~al.(2022{\natexlab{b}})Wu, Chen, Hou, Liao, Wang, Sun, Yan, and Lin]{wu2022fastvqa}
Haoning Wu, Chaofeng Chen, Jingwen Hou, Liang Liao, Annan Wang, Wenxiu Sun, Qiong Yan, and Weisi Lin.
\newblock Fast-vqa: Efficient end-to-end video quality assessment with fragment sampling.
\newblock In \emph{ECCV}, 2022{\natexlab{b}}.

\bibitem[Wu et~al.(2022{\natexlab{c}})Wu, Chen, Liao, Hou, Sun, Yan, Gu, and Lin]{wu2022fastervqa}
Haoning Wu, Chaofeng Chen, Liang Liao, Jingwen Hou, Wenxiu Sun, Qiong Yan, Jinwei Gu, and Weisi Lin.
\newblock Neighbourhood representative sampling for efficient end-to-end video quality assessment.
\newblock \emph{arXiv preprint arXiv:2210.05357}, 2022{\natexlab{c}}.

\bibitem[Wu et~al.(2023{\natexlab{a}})Wu, Chen, Liao, Hou, Sun, Yan, and Lin]{wu2023discovqa}
Haoning Wu, Chaofeng Chen, Liang Liao, Jingwen Hou, Wenxiu Sun, Qiong Yan, and Weisi Lin.
\newblock Discovqa: Temporal distortion-content transformers for video quality assessment.
\newblock \emph{IEEE Transactions on Circuits and Systems for Video Technology}, pages 1--1, 2023{\natexlab{a}}.

\bibitem[Wu et~al.(2023{\natexlab{b}})Wu, Liao, Chen, Hou, Zhang, Wang, Sun, Yan, and Lin]{wu2023bvqi}
Haoning Wu, Liang Liao, Chaofeng Chen, Jingwen~Hou Hou, Erli Zhang, Annan Wang, Wenxiu~Sun Sun, Qiong Yan, and Weisi Lin.
\newblock Exploring opinion-unaware video quality assessment with semantic affinity criterion.
\newblock In \emph{ICME}, 2023{\natexlab{b}}.

\bibitem[Wu et~al.(2023{\natexlab{c}})Wu, Liao, Wang, Chen, Hou, Zhang, Sun, Yan, and Lin]{wu2023bvqiplus}
Haoning Wu, Liang Liao, Annan Wang, Chaofeng Chen, Jingwen~Hou Hou, Erli Zhang, Wenxiu~Sun Sun, Qiong Yan, and Weisi Lin.
\newblock Towards robust text-prompted semantic criterion for in-the-wild video quality assessment.
\newblock \emph{arXiv preprint arXiv:2304.14672}, 2023{\natexlab{c}}.

\bibitem[Wu et~al.(2023{\natexlab{d}})Wu, Zhang, Liao, Chen, Hou, Wang, Sun, Yan, and Lin]{wu2023dover}
Haoning Wu, Erli Zhang, Liang Liao, Chaofeng Chen, Jingwen~Hou Hou, Annan Wang, Wenxiu~Sun Sun, Qiong Yan, and Weisi Lin.
\newblock Exploring video quality assessment on user generated contents from aesthetic and technical perspectives.
\newblock In \emph{ICCV}, 2023{\natexlab{d}}.

\bibitem[Wu et~al.(2023{\natexlab{e}})Wu, Zhang, Liao, Chen, Hou, Wang, Sun, Yan, and Lin]{wu2023explainablevqa}
Haoning Wu, Erli Zhang, Liang Liao, Chaofeng Chen, Jingwen~Hou Hou, Annan Wang, Wenxiu~Sun Sun, Qiong Yan, and Weisi Lin.
\newblock Towards explainable video quality assessment: A database and a language-prompted approach.
\newblock In \emph{ACM MM}, 2023{\natexlab{e}}.

\bibitem[Wu et~al.(2022{\natexlab{d}})Wu, Wang, Yang, Gan, Liu, Yuan, and Wang]{wu2022grit}
Jialian Wu, Jianfeng Wang, Zhengyuan Yang, Zhe Gan, Zicheng Liu, Junsong Yuan, and Lijuan Wang.
\newblock Grit: A generative region-to-text transformer for object understanding.
\newblock \emph{arXiv preprint arXiv:2212.00280}, 2022{\natexlab{d}}.

\bibitem[Wu et~al.(2022{\natexlab{e}})Wu, Ge, Wang, Lei, Gu, Hsu, Shan, Qie, and Shou]{wu2022tuneavideo}
Jay~Zhangjie Wu, Yixiao Ge, Xintao Wang, Stan~Weixian Lei, Yuchao Gu, Wynne Hsu, Ying Shan, Xiaohu Qie, and Mike~Zheng Shou.
\newblock Tune-a-video: One-shot tuning of image diffusion models for text-to-video generation.
\newblock \emph{arXiv preprint arXiv:2212.11565}, 2022{\natexlab{e}}.

\bibitem[Xing et~al.(2023{\natexlab{a}})Xing, Xia, Liu, Zhang, Zhang, He, Liu, Chen, Cun, Wang, Shan, and Wong]{xing2023make}
Jinbo Xing, Menghan Xia, Yuxin Liu, Yuechen Zhang, Yong Zhang, Yingqing He, Hanyuan Liu, Haoxin Chen, Xiaodong Cun, Xintao Wang, Ying Shan, and Tien-Tsin Wong.
\newblock Make-your-video: Customized video generation using textual and structural guidance.
\newblock \emph{arXiv preprint arXiv:2306.00943}, 2023{\natexlab{a}}.

\bibitem[Xing et~al.(2023{\natexlab{b}})Xing, Xia, Zhang, Chen, Wang, Wong, and Shan]{xing2023dynamicrafter}
Jinbo Xing, Menghan Xia, Yong Zhang, Haoxin Chen, Xintao Wang, Tien-Tsin Wong, and Ying Shan.
\newblock Dynamicrafter: Animating open-domain images with video diffusion priors.
\newblock \emph{arXiv preprint arXiv:2310.12190}, 2023{\natexlab{b}}.

\bibitem[Xing et~al.(2023{\natexlab{c}})Xing, Dai, Hu, Wu, and Jiang]{xing2023simda}
Zhen Xing, Qi Dai, Han Hu, Zuxuan Wu, and Yu-Gang Jiang.
\newblock Simda: Simple diffusion adapter for efficient video generation.
\newblock \emph{arXiv preprint arXiv:2308.09710}, 2023{\natexlab{c}}.

\bibitem[Xu et~al.(2016)Xu, Mei, Yao, and Rui]{xu2016msr}
Jun Xu, Tao Mei, Ting Yao, and Yong Rui.
\newblock Msr-vtt: A large video description dataset for bridging video and language.
\newblock In \emph{CVPR}, 2016.

\bibitem[Yang et~al.(2023)Yang, Zhou, Liu, and Loy]{yang2023rerender}
Shuai Yang, Yifan Zhou, Ziwei Liu, and Chen~Change Loy.
\newblock Rerender a video: Zero-shot text-guided video-to-video translation.
\newblock \emph{arXiv preprint arXiv:2306.07954}, 2023.

\bibitem[Yin et~al.(2023)Yin, Wu, Liang, Shi, Li, Ming, and Duan]{yin2023dragnuwa}
Shengming Yin, Chenfei Wu, Jian Liang, Jie Shi, Houqiang Li, Gong Ming, and Nan Duan.
\newblock Dragnuwa: Fine-grained control in video generation by integrating text, image, and trajectory, 2023.

\bibitem[Yu et~al.(2023)Yu, Cheng, Sohn, Lezama, Zhang, Chang, Hauptmann, Yang, Hao, Essa, and Jiang]{yu2023magvit}
Lijun Yu, Yong Cheng, Kihyuk Sohn, José Lezama, Han Zhang, Huiwen Chang, Alexander~G. Hauptmann, Ming-Hsuan Yang, Yuan Hao, Irfan Essa, and Lu Jiang.
\newblock Magvit: Masked generative video transformer, 2023.

\bibitem[Zhang et~al.(2023{\natexlab{a}})Zhang, Wu, Liu, Zhao, Ran, Gu, Gao, and Shou]{zhang2023show1}
David~Junhao Zhang, Jay~Zhangjie Wu, Jia-Wei Liu, Rui Zhao, Lingmin Ran, Yuchao Gu, Difei Gao, and Mike~Zheng Shou.
\newblock Show-1: Marrying pixel and latent diffusion models for text-to-video generation.
\newblock \emph{arXiv preprint arXiv:2309.15818}, 2023{\natexlab{a}}.

\bibitem[Zhang et~al.(2023{\natexlab{b}})Zhang, Yan, Xu, Feng, and Liew]{zhang2023magicavatar}
Jianfeng Zhang, Hanshu Yan, Zhongcong Xu, Jiashi Feng, and Jun~Hao Liew.
\newblock Magicavatar: Multi-modal avatar generation and animation.
\newblock In \emph{arXiv}, 2023{\natexlab{b}}.

\bibitem[Zhang et~al.(2023{\natexlab{c}})Zhang, Wei, Jiang, Zhang, Zuo, and Tian]{zhang2023controlvideo}
Yabo Zhang, Yuxiang Wei, Dongsheng Jiang, Xiaopeng Zhang, Wangmeng Zuo, and Qi Tian.
\newblock Controlvideo: Training-free controllable text-to-video generation.
\newblock \emph{arXiv preprint arXiv:2305.13077}, 2023{\natexlab{c}}.

\bibitem[Zhang et~al.(2023{\natexlab{d}})Zhang, Li, Nie, Han, Guo, and Liu]{zhang2023consistent}
Zicheng Zhang, Bonan Li, Xuecheng Nie, Congying Han, Tiande Guo, and Luoqi Liu.
\newblock Towards consistent video editing with text-to-image diffusion models, 2023{\natexlab{d}}.

\bibitem[Zhao et~al.(2023{\natexlab{a}})Zhao, Wang, Bao, Li, and Zhu]{zhao2023controlvideo}
Min Zhao, Rongzhen Wang, Fan Bao, Chongxuan Li, and Jun Zhu.
\newblock Controlvideo: Adding conditional control for one shot text-to-video editing.
\newblock \emph{arXiv preprint arXiv:2305.17098}, 2023{\natexlab{a}}.

\bibitem[Zhao et~al.(2023{\natexlab{b}})Zhao, Xie, Hong, Li, and Lee]{zhao2023makeaprotagonist}
Yuyang Zhao, Enze Xie, Lanqing Hong, Zhenguo Li, and Gim~Hee Lee.
\newblock Make-a-protagonist: Generic video editing with an ensemble of experts.
\newblock \emph{arXiv preprint arXiv:2305.08850}, 2023{\natexlab{b}}.

\bibitem[Zheng et~al.(2023)Zheng, Chiang, Sheng, Zhuang, Wu, Zhuang, Lin, Li, Li, Xing, et~al.]{zheng2023judging}
Lianmin Zheng, Wei-Lin Chiang, Ying Sheng, Siyuan Zhuang, Zhanghao Wu, Yonghao Zhuang, Zi Lin, Zhuohan Li, Dacheng Li, Eric Xing, et~al.
\newblock Judging llm-as-a-judge with mt-bench and chatbot arena.
\newblock \emph{arXiv preprint arXiv:2306.05685}, 2023.

\bibitem[Zhou et~al.(2014)Zhou, Lapedriza, Xiao, Torralba, and Oliva]{zhou2014learning}
Bolei Zhou, Agata Lapedriza, Jianxiong Xiao, Antonio Torralba, and Aude Oliva.
\newblock Learning deep features for scene recognition using places database.
\newblock In \emph{NeurIPS}, 2014.

\bibitem[Zhou et~al.(2022)Zhou, Wang, Yan, Lv, Zhu, and Feng]{zhou2022magicvideo}
Daquan Zhou, Weimin Wang, Hanshu Yan, Weiwei Lv, Yizhe Zhu, and Jiashi Feng.
\newblock Magicvideo: Efficient video generation with latent diffusion models.
\newblock \emph{arXiv preprint arXiv:2211.11018}, 2022.

\bibitem[Zhou et~al.(2023)Zhou, Wang, Yan, Lv, Zhu, and Feng]{zhou2023magicvideo}
Daquan Zhou, Weimin Wang, Hanshu Yan, Weiwei Lv, Yizhe Zhu, and Jiashi Feng.
\newblock Magicvideo: Efficient video generation with latent diffusion models, 2023.

\end{thebibliography}
}

\clearpage
\maketitlesupplementary

\renewcommand\thesection{\Alph{section}}
\renewcommand\thefigure{A\arabic{figure}}
\renewcommand\thetable{A\arabic{table}}

In this \textbf{\textit{supplementary file}}, 
we provide more details on \textit{Evaluation Dimension Suite} and \textit{Evaluation Method Suite} in Section~\ref{suppsec:dimension}, 
and elaborate on \textit{Prompt Suite} details in Section~\ref{suppsec:prompt}. 
We then provide further explanations on
\textit{Human Preference Annotations} in Section~\ref{suppsec:human}, 
and more implementation details on our experiments and visualizations in Section~\ref{suppsec:implementation}. 
The potential societal impacts of our work are discussed in Section~\ref{suppsec:impact}. 
We also discuss our limitations in Section~\ref{suppsec:limitation}. 
Finally, in Section~\ref{suppsec:additional}, we provide additional experimental results used to support the visualizations and insights in the main paper.

A \textbf{\textit{demo video}} is also provided along with this supplementary file to illustrate VBench and show video examples of each dimension.

\section{More Details on Evaluation Dimension and Method Suite}
\label{suppsec:dimension}

\subsection{Video Quality}

\begin{figure}[htbp]
\centering
\includegraphics[width=0.99\linewidth]{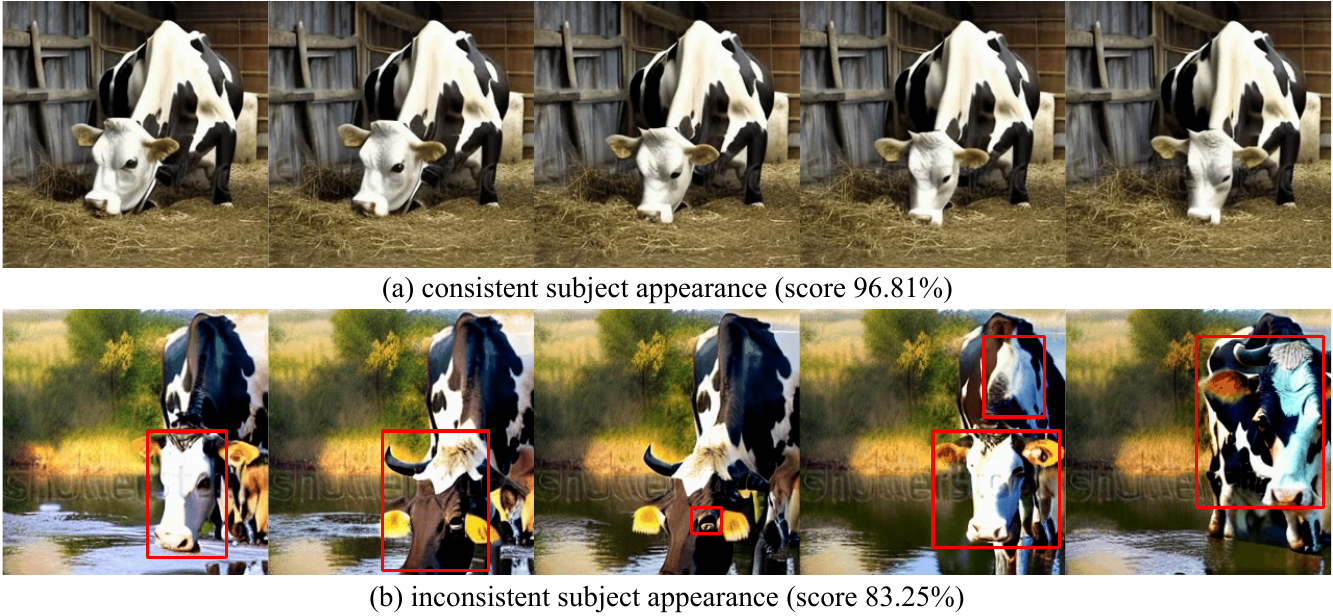}
\caption{\textbf{Visualization of \textit{Subject Consistency}}. We demonstrate different degrees of subject consistency, as indicated by our \textit{Subject Consistency} score (the larger the better) \textbf{(a)} The cow has a relatively consistent look throughout across different frames. \textbf{(b)} The cow shows inconsistency in its appearance over time. The red boxes indicate areas of subject inconsistency.}
\label{fig:subject}
\end{figure}

\smallTitle{Subject Consistency.} 
When there is a subject (\eg, a cow, a person, a car, or a cat) in the video, it is important that the subject looks consistent throughout the video (\ie, whether it is still the same thing or the same person). For example, in Figure~\ref{fig:subject}, the cow in the top row remains consistent across different frames, while the cow in the bottom row shows changes in appearance between frames. To evaluate subject consistency,  we employ DINO~\cite{caron2021emerging} to extract features from each frame to represent the subject. Since DINO is not trained to disregard the differences within subjects of the same class~\cite{ruiz2022dreambooth}, its feature extraction is particularly sensitive to the identity variations of the subject within the video, thereby making it a suitable tool for evaluating subject consistency.
Specifically, for each video, the subject consistency score is calculated as:

\begin{gather}
    S_{subject} = \frac{1}{T-1}\sum_{t=2}^{T} {\it \frac{1}{2}}(\langle d_1\cdot d_t \rangle + \langle d_{t-1} \cdot d_t\rangle),
    \label{eq:subject_consistency}
\end{gather}
where $d_{i}$ is the DINO image feature of the $i^{th}$ frame, normalized to unit length, and $\langle \cdot \rangle$ is the dot product operation for calculating cosine similarity. For each frame, we calculate the cosine similarity with the first frame and its preceding frame, take the average, and then compute the mean over all the non-starting video frames. We average the score $S_{subject}$ for all the videos generated by one model as the final score of the model.

\begin{figure}[htbp]
\centering
\includegraphics[width=0.99\linewidth]{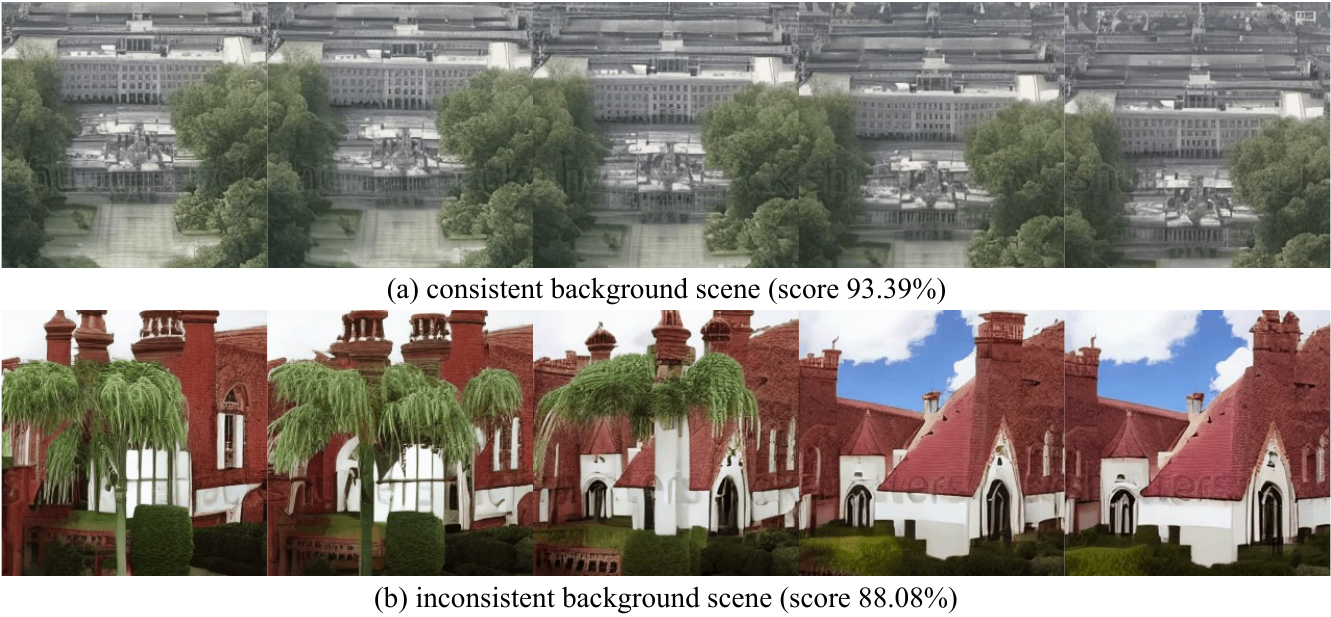}
\caption{\textbf{Visualization of \textit{Background Consistency}}. We showcase varying levels of background consistency, as indicated by
our \textit{Background Consistency} metrics (larger values denote better consistency) \textbf{(a)} The background scene maintains a high degree of consistency (\ie, still the same scene) across different frames. \textbf{(b)} The background exhibits noticeable distortion and abrupt changes over time.}
\vspace{-5pt}

\label{fig:background}
\end{figure}

\smallTitle{Background Consistency.}
Beyond the focus on the foreground subject, maintaining a consistent background scene across different frames is equally important. 
For example, in Figure~\ref{fig:background}, in the top row, the scene maintains a consistent appearance as the camera moves, while in the bottom row, the entire scene undergoes significant changes within a few frames.  
For each video frame, we employ the CLIP~\cite{radford2021clip} image encoder to extract its feature vector. We then compute the background consistency metric, which is similar to the method used for \textit{Subject Consistency}:

\begin{equation}
    S_{background} = \frac{1}{T-1}\sum_{t=2}^{T} {\it \frac{1}{2}}(\langle c_1 \cdot c_t \rangle + \langle c_{t-1} \cdot c_t \rangle),
\label{eq:background_consistency}
\end{equation}
where $c_{i}$ represents the CLIP image feature of the $i^{th}$ frame, normalized to unit length.

\begin{figure}[htbp]
\centering
\includegraphics[width=0.99\linewidth]{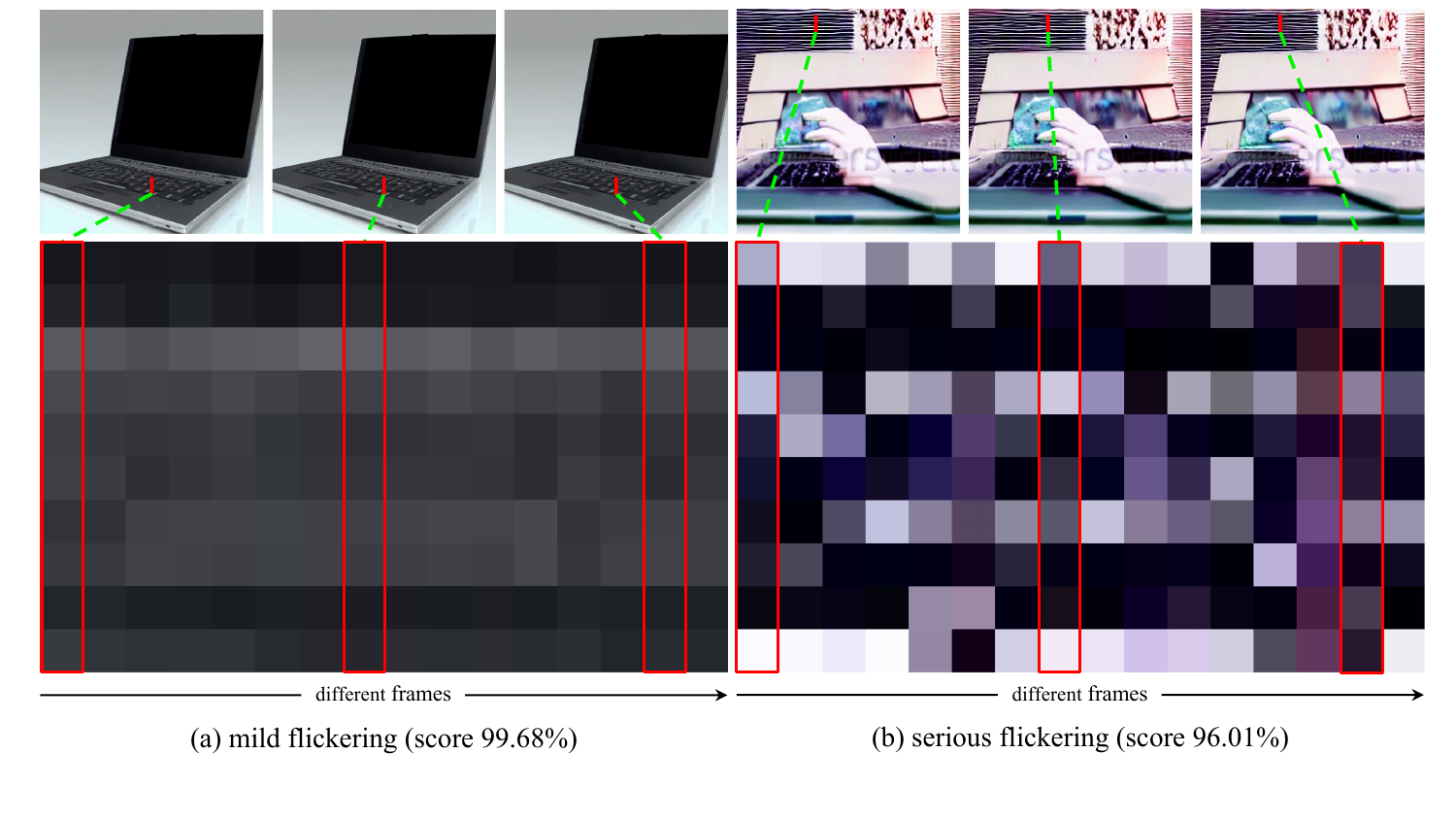}
\vspace{-5pt}
\caption{
\textbf{Visualization of \textit{Temporal Flickering}}. We demonstrate different degrees of temporal flickering, with a mild occurrence in (a), and a severe occurrence in (b), both reflected by our flicker score metrics (the larger the better). To visualize temporal flickering, given a generated video (\textbf{\textit{top row}}), we extract a small segment of pixels (marked as the red segment) from each frame at the same location and stack them in frame order (\textbf{\textit{bottom row}}). \textbf{(a)} Pixel values do not vary abruptly, and the video suffers less from flickering. \textbf{(b)} Pixel values vary abruptly and frequently across different frames, showing strong temporal flickering. Our evaluation metrics also give a lower score.}
\label{fig:flicker}
\end{figure}

\begin{table}[htbp]
    \caption{\textbf{\textit{Dynamic Degree} on Three Benchmarks.} We report the \textit{Dynamic Degree} metrics on three \textit{Temporal Flickering} benchmarks. We use videos from the Subject Consistency dimension as the ``Dynamic Benchmark'', videos from the Background Consistency dimension as the ``Semi-Dynamic Benchmark'', and videos from the temporal flickering dimension as the ``Static Benchmark''.}
    \footnotesize
    \centering
    \resizebox{0.47\textwidth}{!}{
    \begin{tabular}{c|c|c|c}
        \Xhline{1pt}  
        \textbf{Models}   & \textbf{\Centerstack{Static\\Benchmark}} & 
        \textbf{\Centerstack{Semi-Dynamic\\Benchmark}} & 
        \textbf{\Centerstack{Dynamic\\Benchmark}}
        \\ \Xhline{1pt}
        LaVie~\cite{wang2023lavie}   & 0.00\% & 6.51\% & 49.72\%  \\ 
        ModelScope~\cite{luo2023videofusion, wang2023modelscope}  & 0.00\% & 33.72\% & 66.39\% \\ 
        VideoCrafter~\cite{he2022lvdm} & 0.00\% & 51.63\% & 89.72\% \\ 
        CogVideo~\cite{hong2022cogvideo}  & 0.00\% & 14.19\% & 42.22\%\\ \hline
        \Xhline{1pt}  
    \end{tabular}
    }
    \centering
  \label{tab:flicker-dynamic}
\end{table}

\smallTitle{Temporal Flickering.} 
For real videos, temporal flickering is usually a result of frequent lighting variation, or shaky camera motions during the video capture process. However, for generated videos, temporal flickering is an intrinsic property of the video generation model, usually caused by imperfect temporal consistency at local and high-frequency details. In generated videos, temporal inconsistency can be attributed to various types of issues, including temporal flickering, unnatural motions, subject inconsistency \etc. To disentangle the evaluation of temporal flickering from other aspects, we use static video scenes (\ie, no apparent motions) as the test cases (We use carefully designed prompts to generate static scenes for video sampling. To further ensure that the evaluation is conducted on static videos without apparent motions, we employ an optical flow estimator~\cite{teed2020raft} to filter out videos and only keep the static videos).
\begin{figure}[t]
\centering
\includegraphics[width=0.99\linewidth]{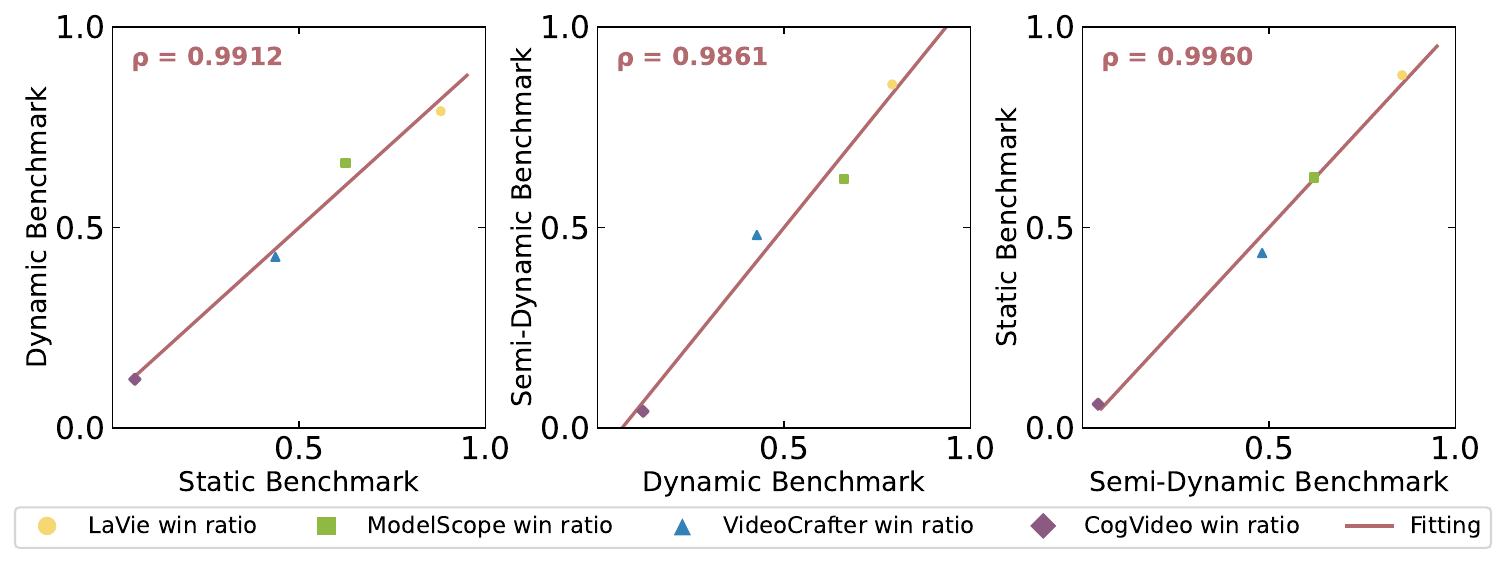}
\vspace{-5pt}
\caption{\textbf{Temporal Flickering Human Preference across Different Dynamic Degrees.} In each plot, a dot represents the human preference win ratio, where the horizontal and vertical axes correspond to two different benchmarks with different dynamic degrees. We linearly fit a straight line to visualize the correlation and calculate the correlation ($\rho$) for each dimension. We observe that the human preferences in terms of temporal flickering on these three benchmarks have high mutual correlations of around 99\%.}
\label{fig:flicker_validation}
\end{figure}
We calculate the frame-by-frame temporal flickering degree with the following formula:
\begin{gather}
    S_{flicker} = \frac{1}{N}\sum_{i=1}^N(\frac{1}{T-1}\sum_{t=1}^{T-1} {\it MAE}(f_i^t, f_i^{t+1})),
    \label{eq:flicker_degree}
\end{gather}
where N is the number of videos generated by a model, T is the number of frames per video, $f_i^t$ is the frame $t$ in video $i$, and MAE is the Mean Absolute Error between two consecutive frames over all pixel locations.
We then normalize the temporal flickering degree to $[0, 1]$ as follows: 
\begin{gather}
    S_{flicker-norm} = \frac{255 - S_{flicker}}{255},
    \label{eq:flicker_score}
\end{gather}
where a higher score implies less flickering, and thus better video perceptual quality in terms of temporal flickering.

To verify that the strength of motions (\ie, large motion or small motion) in videos does not significantly impact the model's ranking in terms of temporal flickering, we conduct separate human evaluations for the level of temporal flickering on videos with different dynamic degrees, and show in Figure~\ref{fig:flicker_validation} that model ranking in terms of temporal flickering does not vary based on the dynamic degree of test videos. For videos of high dynamic degrees, we use videos from the \textit{Subject Consistency} dimension's prompt suite, and term as the ``Dynamic Benchmark''. For videos that exhibit lower dynamic degrees but remain non-static, we use videos sampled from the \textit{Background Consistency} dimension's prompt suite, and label them as the ``Semi-Dynamic Benchmark''. Additionally, the ``Static Benchmark'' refers to the videos sampled from the prompt suite for the \textit{Temporal Flickering} dimension. We show the dynamic degree of videos in these three benchmarks in Table ~\ref{tab:flicker-dynamic}. In Figure~\ref{fig:flicker_validation}, we show that the human win ratio in terms of temporal flickering on three benchmarks is almost perfectly correlated with each other, with a correlation of around 99\% between any two benchmarks. Therefore, we believe the degree of motion is disentangled with the temporal flickering ranking in video generative models, and we use the ``Static Benchmark'' for easier and more focused evaluation on \textit{Temporal Flickering}.

\begin{figure}[htbp]
\centering
\includegraphics[width=0.99\linewidth]{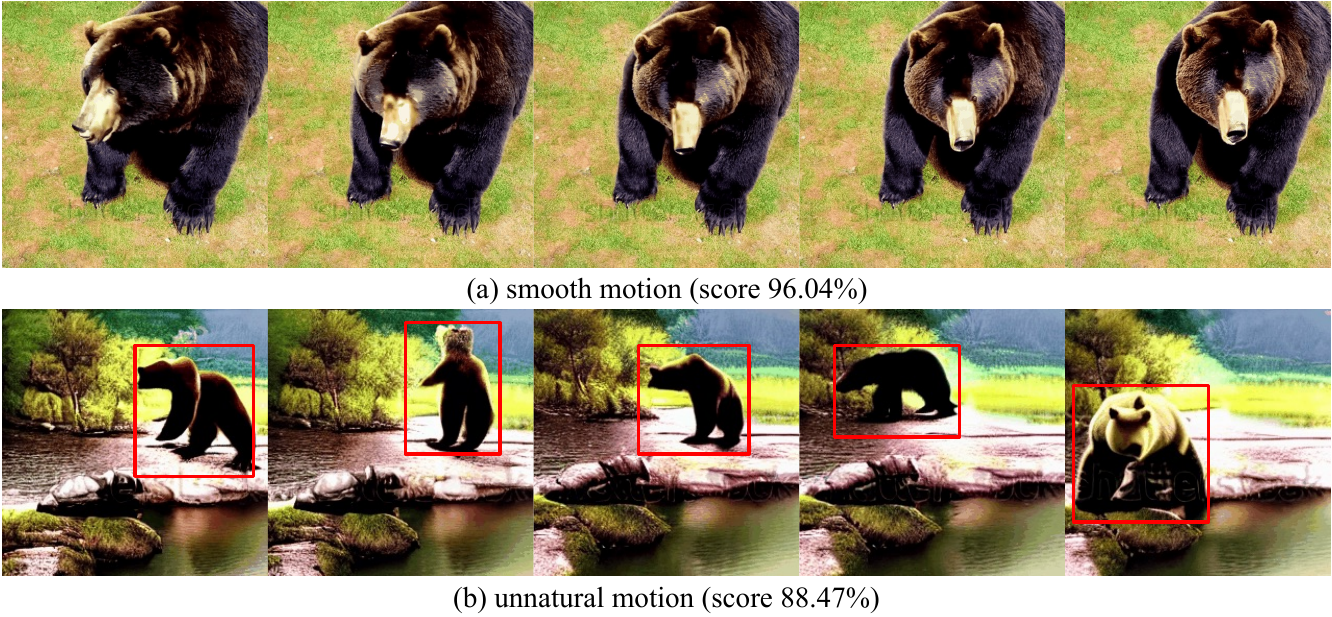}
\vspace{-5pt}
\caption{\textbf{Visualization of \textit{Motion Smoothness}}. We investigate various levels of motion smoothness, ranging from being smooth as depicted in (a) to highly erratic as depicted in (b), as indicated by our motion score metrics (larger values denote better smoothness). The red boxes indicate areas of discontinuous motion.}
\label{fig:motion}
\end{figure}

\smallTitle{Motion Smoothness.} 
To evaluate whether the motion in the generated video is smooth and follows the physical law of the real world, we make use of the frame-by-frame motion prior to video frame interpolation models. Specifically, video frame interpolation models usually assume real-world motions within a very short time period (\ie, a few consecutive frames) to be linear or quadratic and synthesize the non-existing intermediate frames based on this assumption. Given a generated video consisting of frames $[f_0, f_1, f_2, f_3, f_4 ..., f_{2n-2}, f_{2n-1}, f_{2n}]$, we manually drop the odd-number frames to obtain a lower-frame-rate video $[f_0, f_2, f_4 ..., f_{2n-2}, f_{2n}]$, and use video frame interpolation~\cite{licvpr23amt} to infer the dropped frames $[\hat{f}_1, \hat{f}_3, ..., \hat{f}_{2n-1}]$. We then compute the Mean Absolute Error (MAE) between the reconstructed frames and the original dropped frames. The calculated MAE is normalized in the same way as Equation~\ref{eq:flicker_score}, so that the final score falls into $[0, 1]$, with a larger number implying smoother motion.

\begin{figure}[htbp]
\centering
\includegraphics[width=0.99\linewidth]{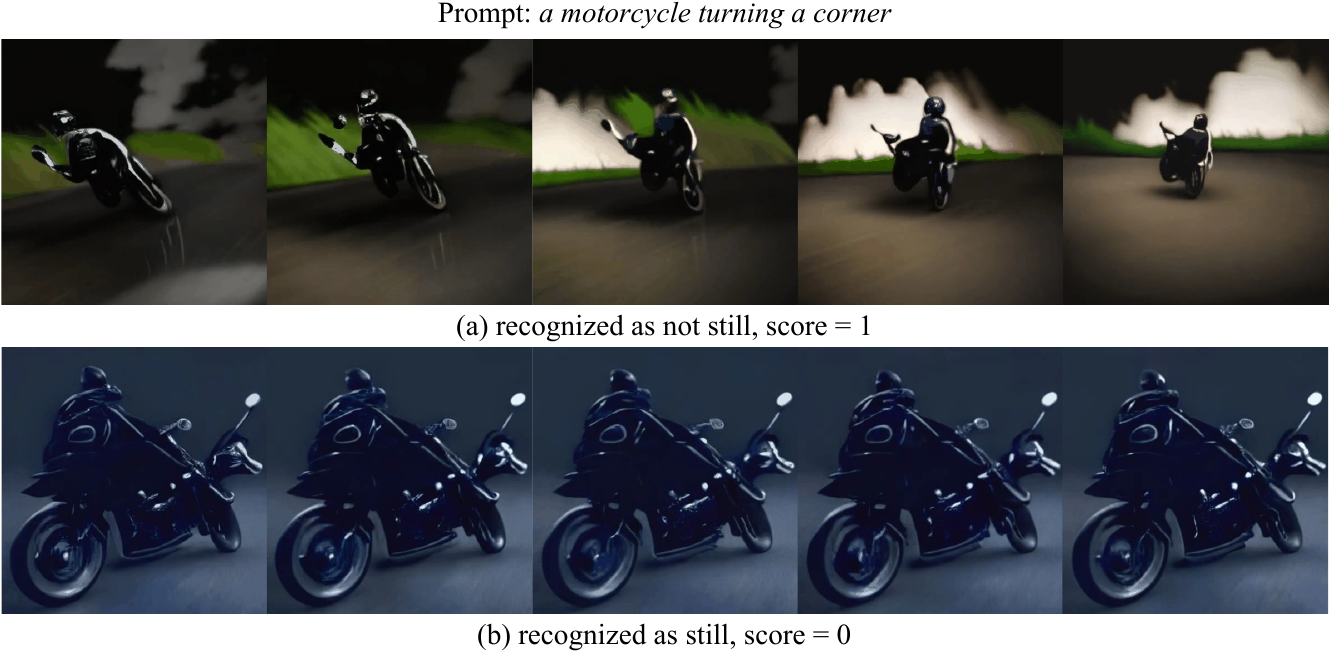}
\vspace{-5pt}
\caption{\textbf{Visualization of \textit{Dynamic Degree}}. We present generated examples of different degrees of motion. \textbf{(a)} In the video, there is obvious motion of the camera and the object, which is identified as dynamic. \textbf{(b)} The video remains almost unchanged from the start to the end and is identified as static.}
\label{fig:dynamic}
\end{figure}

\smallTitle{Dynamic Degree.} 
Based on our observations, some models tend to generate static videos even when the prompt includes descriptions of movement. This results in a noticeable advantage for these models in evaluations of other temporal consistency dimensions, leading to unfair comparisons. This dimension is designed to assess the extent to which models tend to generate non-static videos. We use RAFT~\cite{teed2020raft} to estimate optical flow strengths between consecutive frames of a generated video. We then take the average of the largest 5$\%$ optical flows (considering the movement of small objects in the video) as the basis to determine whether the video is static. The final dynamic degree score is calculated by measuring the proportion of non-static videos generated by the model.

\begin{figure}[htbp]
\centering
\includegraphics[width=0.99\linewidth]{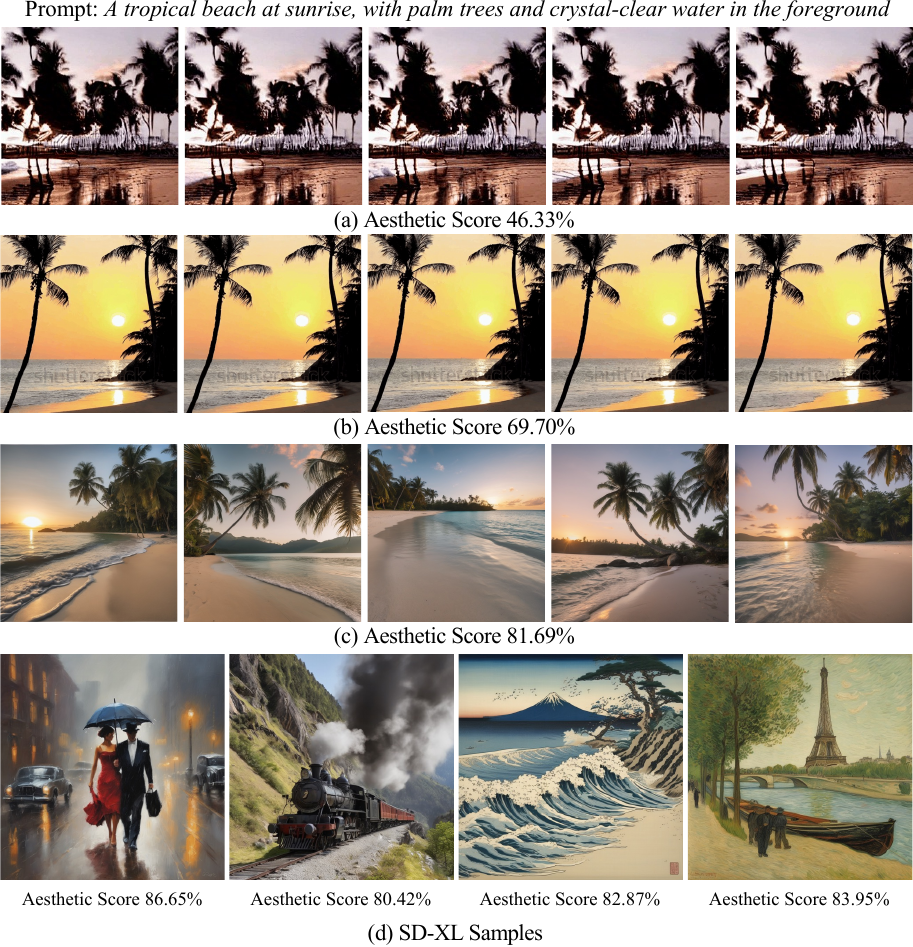}
\vspace{-5pt}
\caption{\textbf{Visualization of \textit{Aesthetic Quality}}. We demonstrate video frames with varying degrees of aesthetic quality in (a), (b), and (c), which are effectively reflected by our aesthetic score metrics (higher indicating better). In (d), we showcase images with high aesthetic scores sampled from SDXL~\cite{podell2023sdxl}.}

\label{fig:aesthetic}
\end{figure}

\smallTitle{Aesthetic Quality.} 
Aesthetic Quality takes photographic layout rules, the richness and harmonies of colors, the artistic quality of the subjects, etc into account. We adopt an image aesthetic quality predictor to evaluate the generated videos frame by frame. We use the LAION aesthetic predictor~\cite{LAIONaes} to give a 0-10 rating for each frame, linearly normalize the score to 0-1, and calculate the average score of all synthetic frames as the final video aesthetic score.

\smallTitle{Imaging Quality.} 
Imaging quality mainly considers the low-level distortions presented in the generated video frames \textit{(e.g., over-exposure, noise, blur)}. We use the MUSIQ~\cite{Ke2021MUSIQ} image quality predictor trained on the SPAQ~\cite{Fang2020spaq} dataset, which is capable of handling variable-sized aspect ratios and resolutions. The frame-wise score is linearly normalized to $[0,1]$ by dividing 100, and the final score is then calculated by averaging the frame-wise scores across the entire video sequence.

\begin{figure}[htbp]
\centering
\includegraphics[width=0.99\linewidth]{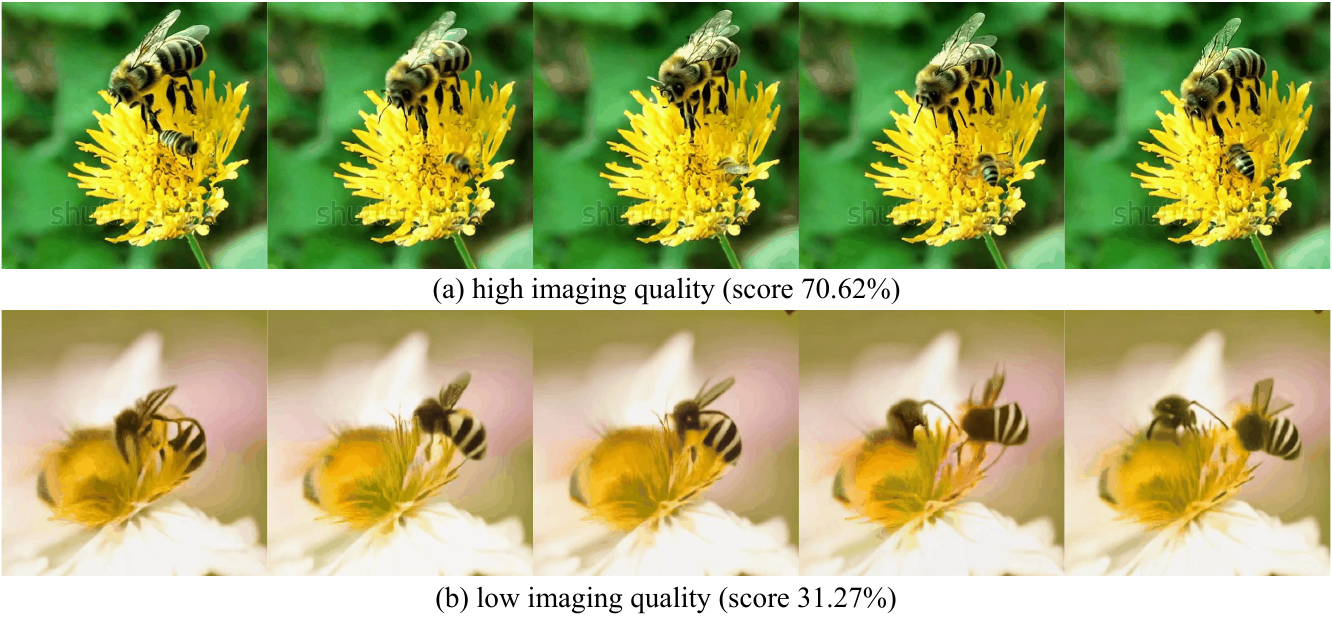}
\vspace{-5pt}
\caption{\textbf{Visualization of \textit{Imaging Quality}}. We present examples of generated videos with high imaging quality scores in (a), and low imaging quality scores (where the video is blurry and over-exposed) in (b).}
\end{figure}

\subsection{Video-Condition Consistency}

\begin{figure}[htbp]
\centering
\includegraphics[width=0.99\linewidth]{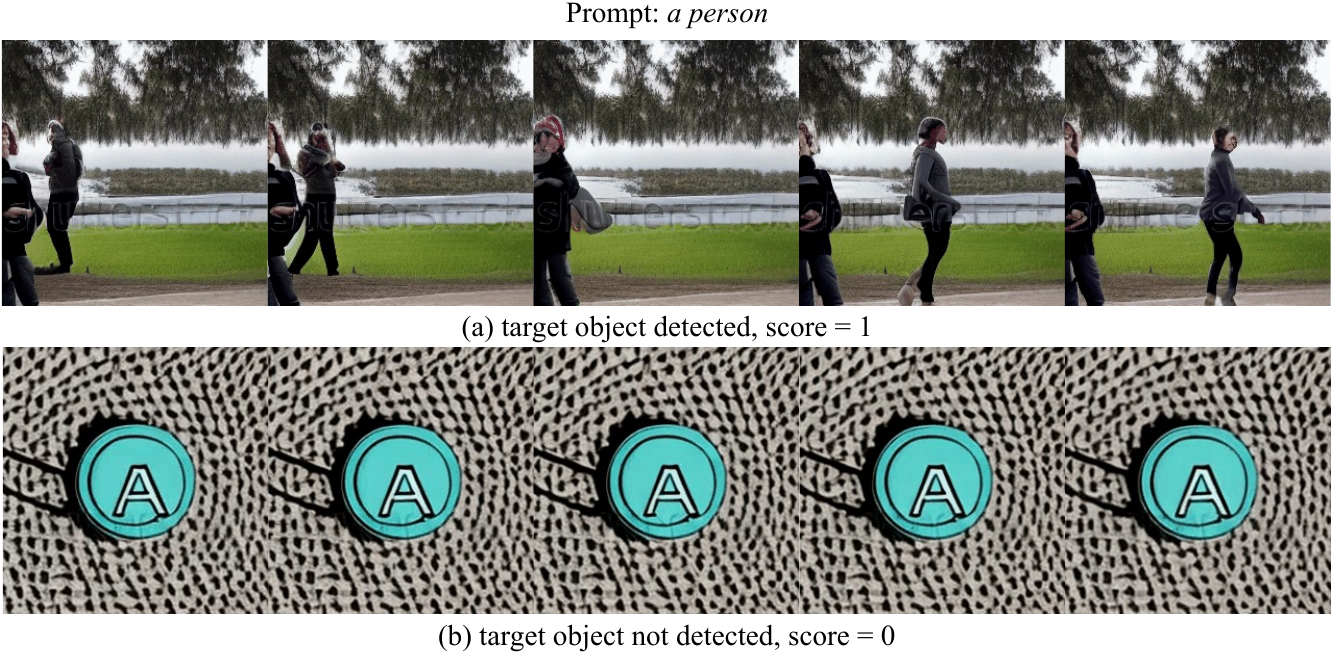}
\vspace{-5pt}
\caption{\textbf{Visualization of \textit{Object Class}}. We demonstrate generation examples for the target object at varying degrees, as reflected by the object score metrics (where 1 represents success, and 0 represents failure). \textbf{(a)} The target object ``person'' is successfully generated in the video. \textbf{(b)} The synthesized video does not contain the target object.}
\label{fig:object}
\end{figure}

\smallTitle{Object Class.} 
When a user specifies a certain type of object in the text prompt, we aim to evaluate whether the model can generate an object of the specified type. To this end, we use GRiT~\cite{wu2022grit} to detect objects in each frame of the generated video and check whether the specified object class is successfully detected in these frames. Subsequently, we report the proportion of frames in which the corresponding object class has been successfully detected. We employ GRiT for this dimension, as well as several other semantics dimensions such as \textit{Multiple Objects}, \textit{Color}, and \textit{Spatial Relationship} for two reasons: \textit{1)} GRiT is a versatile framework that can handle both detection and captioning tasks, predicting diverse object attributes, so that the VBench can use the same framework across different dimensions and save users from installing multiple frameworks or downloading multiple pre-trained models. \textit{2)} GRiT demonstrates reliable performance in evaluating our designated dimensions, with comparable performance with the state-of-the-art object detectors~\cite{wu2022grit}, and good alignment with human perception in terms of ``correct detection'' as validated by the human preference results in main paper Figure 5.

\begin{figure}[htbp]
\centering
\includegraphics[width=0.99\linewidth]{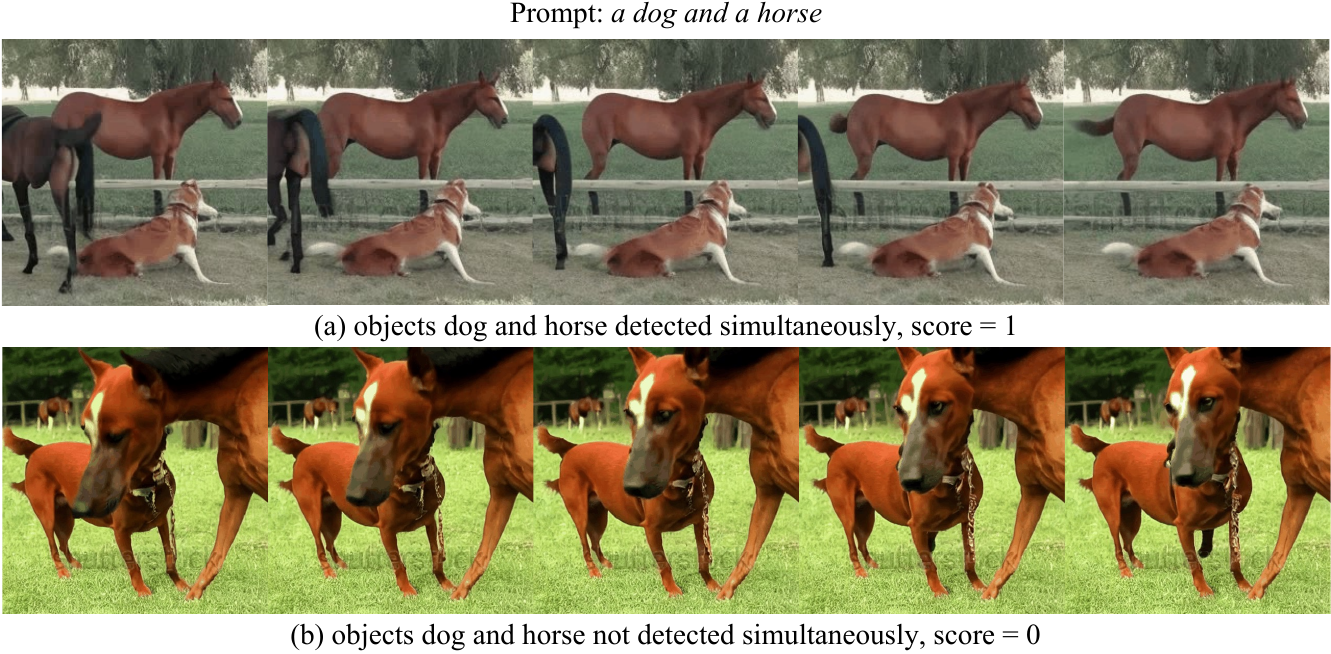}
\vspace{-5pt}
\caption{\textbf{Visualization of \textit{Multiple Objects}}. We showcase instances of generating multiple objects within a video simultaneously at different levels, as indicated by our multiple objects score metrics (where 1 signifies success, and 0 denotes failure). \textbf{(a)} The video effectively generates multiple required objects (\ie, dog and horse). \textbf{(b)} The video fails to produce the dog and horse at the same time.}
\label{fig:multiple_objects}
\end{figure}

\smallTitle{Multiple Objects.} 
Other than generating a single object, compositionality is also an essential aspect of video generation. Suppose the user requires generating multiple objects, we use GRiT for frame-wise object detection. For each frame, we check whether all the user-requested objects simultaneously appear in each frame. We then report the proportion of frames in which all the required objects have been successfully detected.

\begin{figure}[htbp]
\centering
\includegraphics[width=0.99\linewidth]{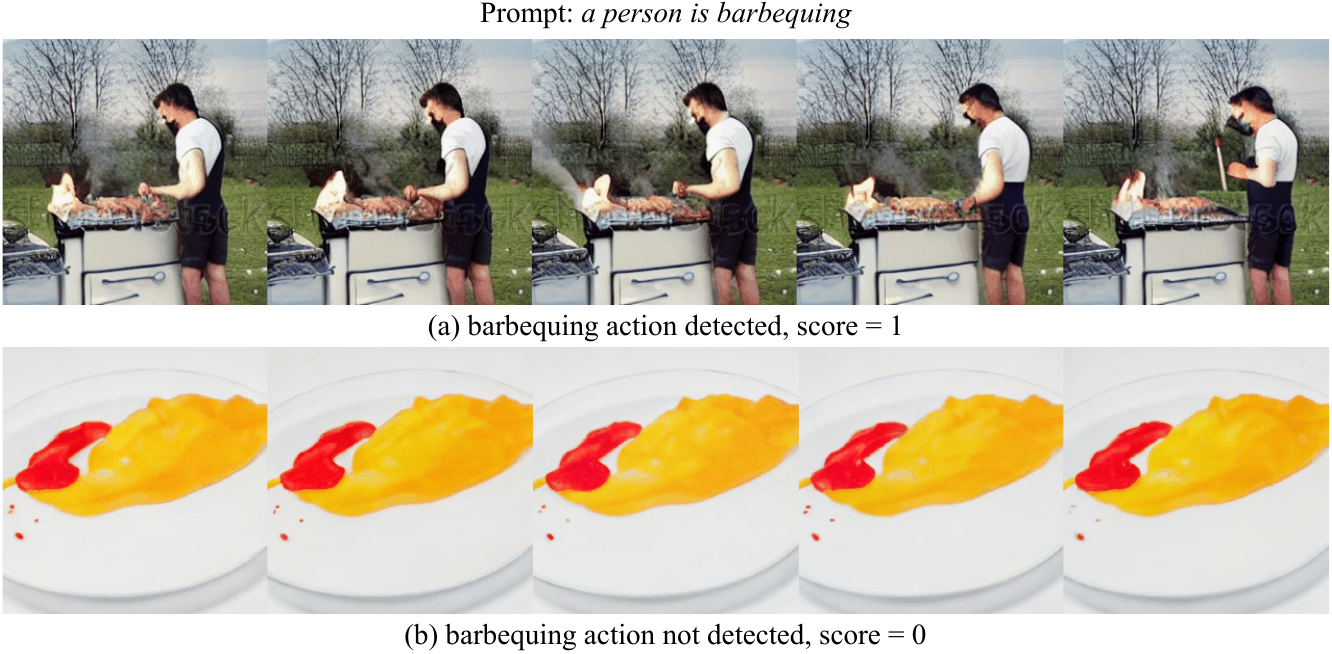}
\vspace{-5pt}
\caption{\textbf{Visualization of \textit{Human Action}}. We showcase examples of generating the target action at different levels, as indicated by our action score metrics (where 1 denotes success, and 0 denotes failure). \textbf{(a)} The video successfully generates the barbequing action. \textbf{(b)} The video does not generate the target action.}
\label{fig:action}
\end{figure}

\smallTitle{Human Action.} 
In the process of video synthesis from textual prompts, both the mentioned subjects in the prompt and the corresponding actions they engage in are important. Given the remarkable emergence of high-quality human-centric generated videos, we believe it is necessary to ensure that human subjects depicted in videos accurately execute the specific actions described by the textual prompts. To this end, we use the Kinetics-400 dataset~\cite{kay2017kinetics} as a reference due to its comprehensive characterization of diverse human actions.
To evaluate the accuracy of the generated videos, we uniformly sample 16 frames from each video and apply UMT~\cite{li2023unmasked}, which achieves the state-of-the-art classification performance on the Kinetics-400 dataset among open-sourced models to classify the action. The top 5 results with logits bigger than 0.85 are preserved as ground-truth candidates, and we check whether the actions mentioned in the text prompt appear in the ground-truth candidates. The average percentage of all classification results is reported to assess whether the generated videos have human actions aligned with the text prompts.

\begin{figure}[htbp]
\centering
\includegraphics[width=0.99\linewidth]{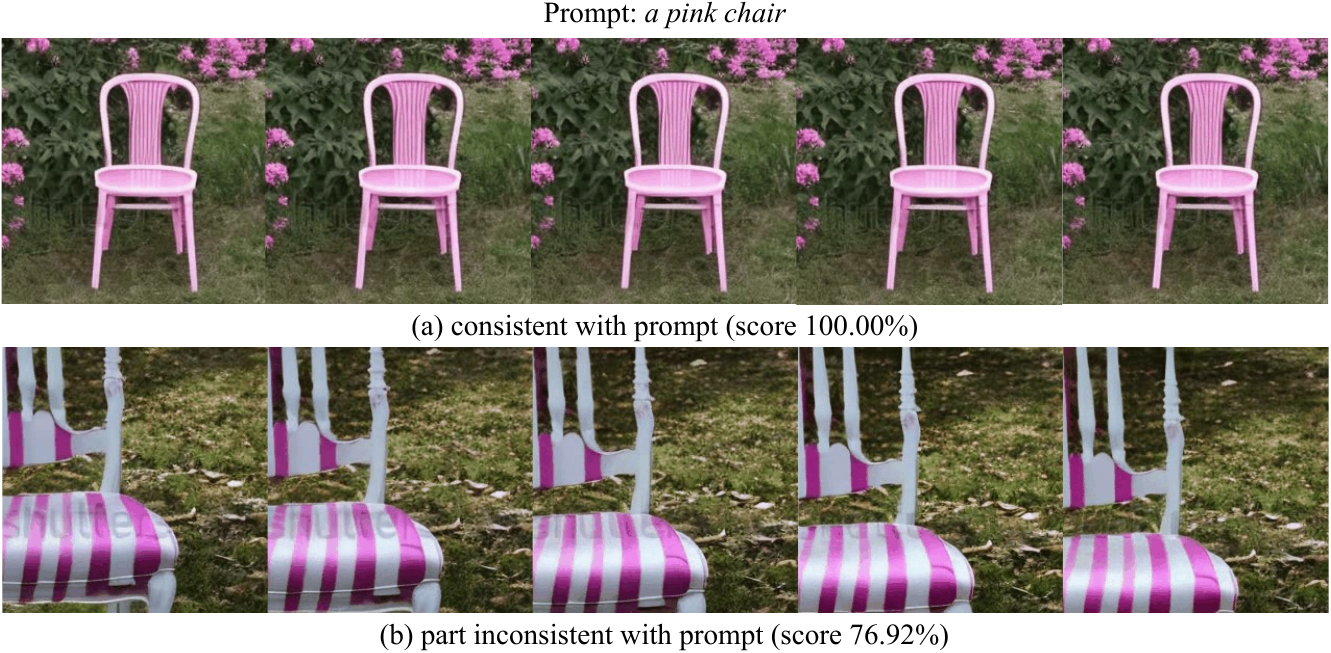}
\vspace{-5pt}
\caption{\textbf{Visualization of \textit{Color}}. We present examples of generating the target color within videos, depicting various levels of success through our color score metrics (larger denotes better). \textbf{(a)} The video accurately generates the target color. \textbf{(b)} The video only generated the target color in certain parts.}
\label{fig:color}
\end{figure}

\smallTitle{Color.}
To evaluate whether the color of an object is consistent with the specified condition, we use GRiT's captioning ability to describe colors, with slight modification to the GRiT pipeline. To remove the influence of the \textit{Object Class} dimension's ability, we only consider videos where the object has been successfully generated. Specifically, GRiT identifies the bounding boxes of objects, which are then fed to two text decoders: one for predicting categories and the other for generating dense captions on the synthesized video frame. 
We then verify if the corresponding object's color is successfully captioned in all frames. Among the frames where the corresponding object is generated and the caption contains color information, we compute the percentage of frames where the color required by the text prompt is successfully captioned.

\begin{figure}[htbp]
\centering
\includegraphics[width=0.99\linewidth]{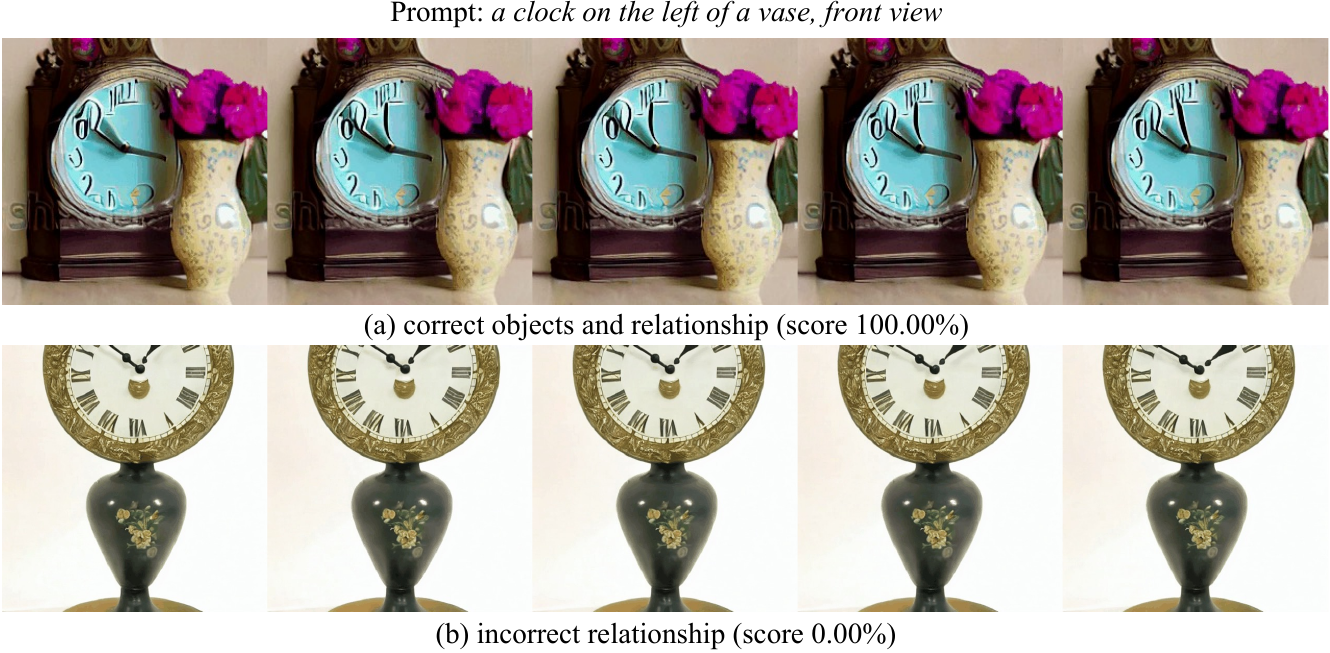}
\vspace{-5pt}
\caption{\textbf{Visualization of \textit{Spatial Relationship}}. We show examples of generating the spatial relationships mentioned in the prompt within videos. \textbf{(a)} The video successfully captures the spatial relationship and objects described in the prompt. \textbf{(b)} The generated video does not contain the intended relationship.}
\label{fig:relation}
\end{figure}

\smallTitle{Spatial Relationship.} 
We focus on \textit{left-right} and \textit{top-bottom} relationships and evaluate whether the video content adheres to the spatial relationship specified by the text prompts. Inspired by the T2I-CompBench~\cite{huang2023t2icompbench} evaluation, we compute the spatial relationship accuracy based on the horizontal and vertical positioning of object pairs. 
During evaluation, distances on the designated axis (e.g., left-right) are expected to be greater than those on the other orientation (e.g., top-bottom). 
Under this condition, we observe the intersection over the union metric (IoU) of two objects to obtain the final score, where IoU values that fall below a specified threshold result in a score of 100\%, and the values exceeding the threshold are multiplied by a coefficient based on the IoU to determine the final score.
We use GRiT to detect the objects and their locations within the generated video frames, and we also calculate the Intersection over Union (IoU) of the two objects' bounding boxes as the final spatial relationship score coefficient.

\begin{figure}[htbp]
\centering
\includegraphics[width=0.99\linewidth]{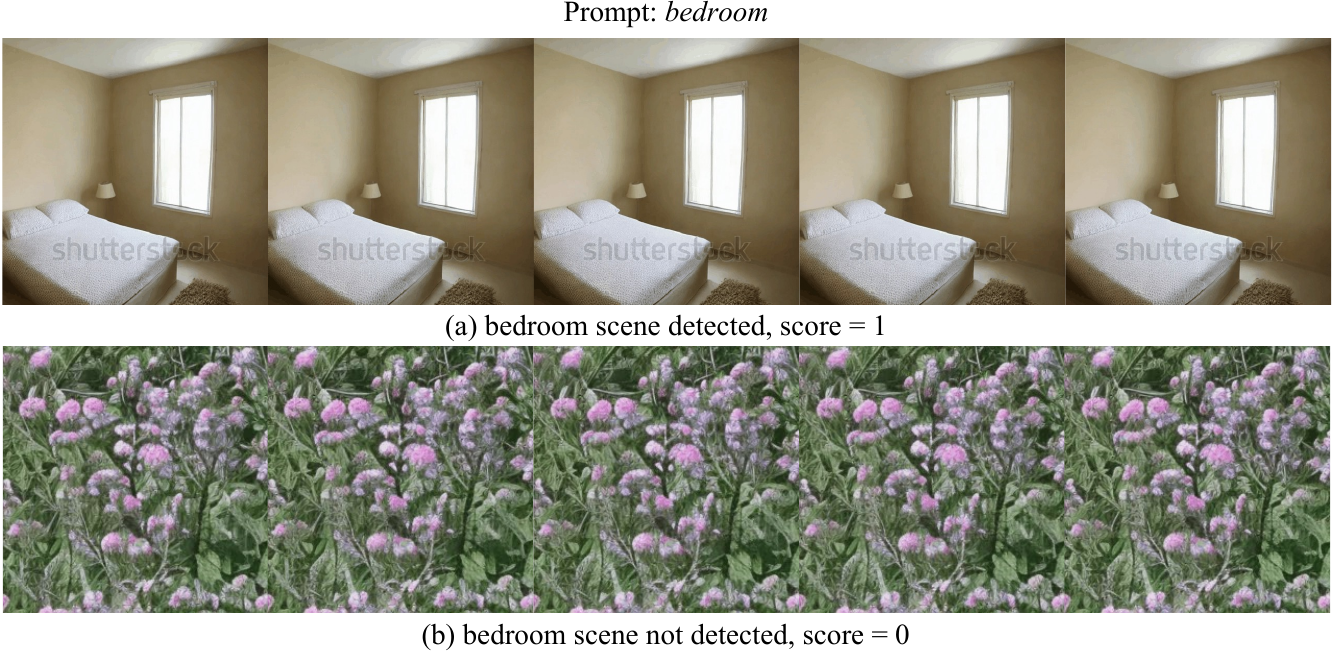}
\vspace{-5pt}
\caption{\textbf{Visualization of \textit{Scene}}. We present examples of generating the required scene (where 1 represents success, and 0 indicates failure). \textbf{(a)} The required scene is generated successfully. \textbf{(b)} The video does not show the scene as required.}
\label{fig:scene}
\end{figure}

\smallTitle{Scene.} 
For a scenario described by the text prompt, we need to evaluate whether the synthesized video is consistent with the intended scene. For example, when prompted to ``ocean'', the generated video should be ``ocean'' instead of ``river''.
We use Tag2Text~\cite{huang2023tag2text} to caption the generated scenes, and then check the correspondence with scene descriptions in the text prompt. Specifically, each word related to the scene in the text prompt needs to appear in the predicted caption, but the word order can be different. We then report the proportion of frames in which the corresponding scene has been successfully generated.

\begin{figure}[htbp]
\centering
\includegraphics[width=0.99\linewidth]{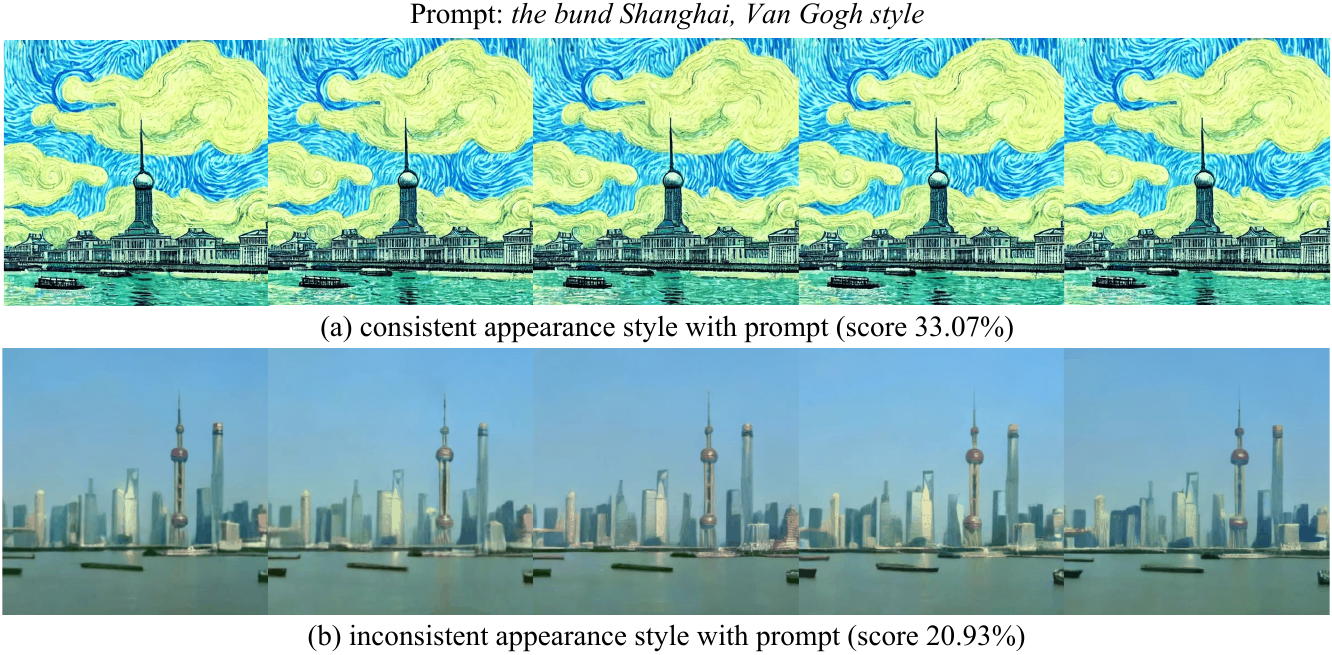}
\vspace{-5pt}
\caption{\textbf{Visualization of \textit{Appearance Style}}. We demonstrate examples of generating the required appearance style within videos, showcasing different levels of success as assessed by our appearance style score metrics. \textbf{(a)} The generated video follows the requested Van Gogh style. \textbf{(b)} The video does not show the desired appearance style.}
\label{fig:appearance_style}
\end{figure}

\smallTitle{Appearance Style.} 
For stylized video generation, we first extract the style description in the text prompt, then evaluate the video-text feature similarity to assess appearance style consistency. Specifically, We use CLIP~\cite{radford2021clip} to extract features from each frame and the text, and then compute the mean cosine similarity of the normalized features. CLIP demonstrates robust zero-shot performance in perceiving textual descriptions of styles, aiding our evaluation of style consistency.

\begin{figure}[htbp]
\centering
\includegraphics[width=0.99\linewidth]{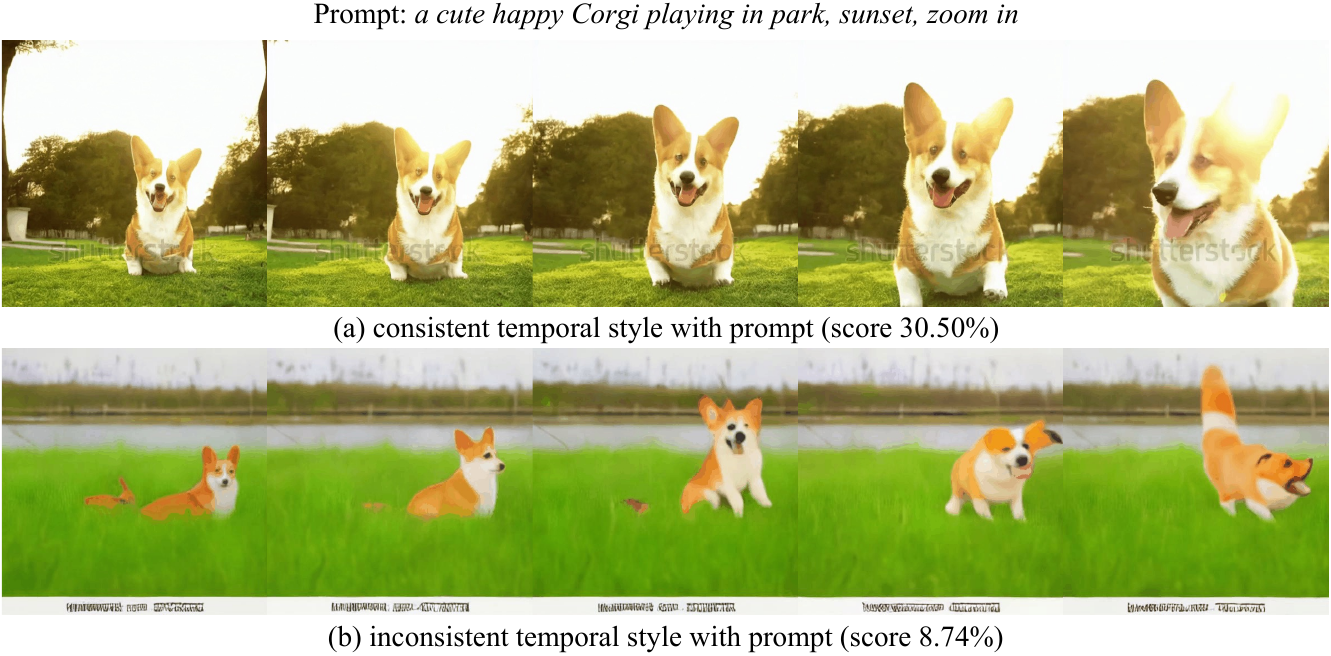}
\vspace{-5pt}
\caption{\textbf{Visualization of \textit{Temporal Style}}. We demonstrate two different generated videos to show the consistency of their temporal style with the prompt at various degrees, measured by our temporal style score. \textbf{(a)} The generated video follows the ``zoom in'' temporal style. \textbf{(b)} The video's temporal style does not align with the prompt.}
\label{fig:temporal_style}
\end{figure}

\smallTitle{Temporal Style.} 
In videos, style is not only spatially narrated in individual frames, but also temporally revealed in different types of object motions and camera motions. For example, we are interested in whether the text prompt specifies ``zoom in'' or ``zoom out'', ``pan left'' or ``pan right'', and whether the generated video can show such kind of camera motion. Additionally, there are different types of other temporal styles like ``super slow motion'', ``camera shaking'', and ``racking focus''. In terms of temporal awareness, ViCLIP~\cite{wang2023internvid} is pre-trained on a diverse 10M video-text dataset, which shows strong zero-shot learning capabilities in video-text retrieval tasks. When a video is generated based on a specified temporal style, we use ViCLIP to calculate the video-text feature similarity to reflect temporal style consistency.

\begin{figure}[htbp]
\centering
\includegraphics[width=0.99\linewidth]{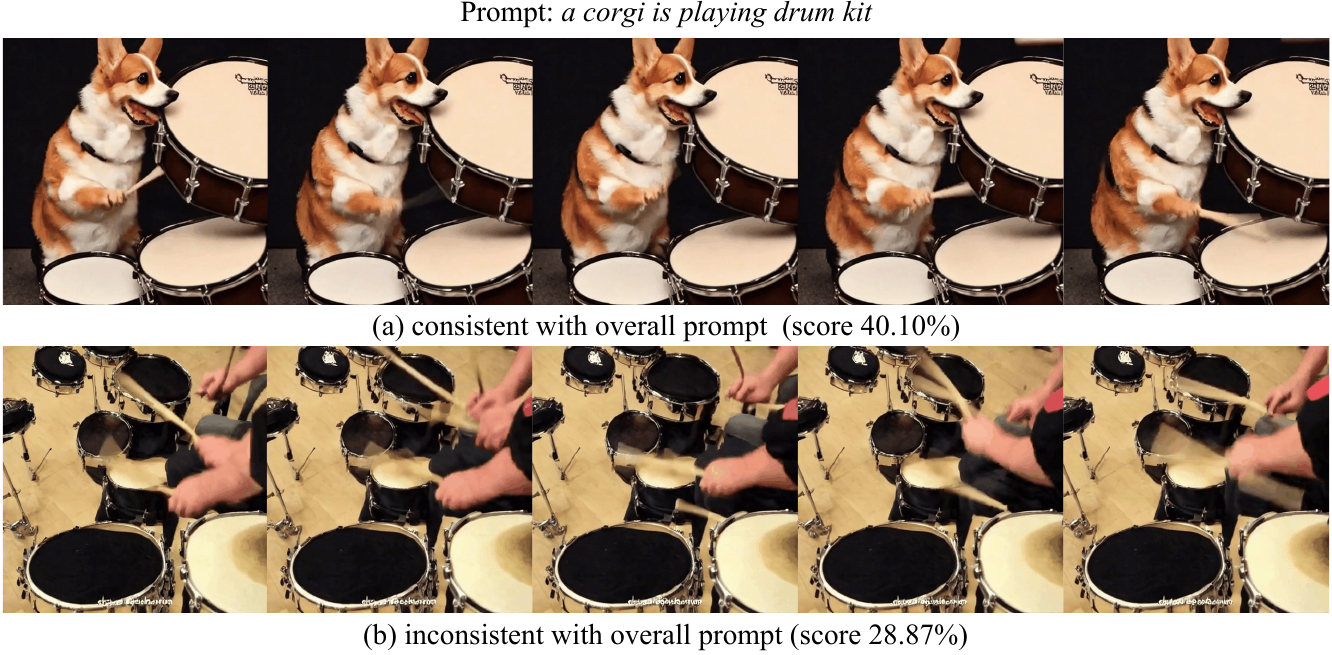}
\vspace{-5pt}
\caption{\textbf{Visualization of \textit{Overall Consistency}}. We demonstrate different examples that illustrate the extent to which they align with the prompt, as measured by our overall score metrics (larger values denote better consistency). \textbf{(a)} The video aligns closely with the prompt. \textbf{(b)} The video lacks alignment with the target concept.}
\label{fig:overall}
\end{figure}

\smallTitle{Overall Consistency.} 
We also use overall video-text consistency computed by ViCLIP as an aiding metric to reflect both semantics and style consistency, where the text prompts contain different semantics and styles.

\section{More Details on Prompt Suite}
\label{suppsec:prompt}

\subsection{Prompt Suite per Evaluation Dimension}  

\label{suppsubsec:prompt_per_dimension}

For each VBench dimension, we carefully designed around 100 prompts as the test cases. \textit{For semantics-related prompt suites, we provide clear semantics labels to each prompt in the prompt suites to facilitate efficient and accurate evaluation.} For example, we provide the object class labels for prompt suites of \textit{Object Class}, \textit{Multiple Objects}, and \textit{Spatial Relationship}. We also provide color labels for \textit{Color} prompts, relationship tags for \textit{Spatial Relationship} prompts, and style labels for \textit{Appearance Style}. We detail the prompt suite for each dimension as follows.

\smallTitle{Subject Consistency.} We choose 19 representative living or movable object categories from the COCO~\cite{lin2014mscoco} dataset's 80 object categories. These categories encompass animals and transportation-related items. Each object category is associated with a set of carefully crafted actions or movements, ensuring logical coherence between the actions and their respective objects. A list of distinct prompts used for evaluating subject consistency is therefore created.

\smallTitle{Background Consistency.} We carefully select a list of distinct and representative scenes from the Places365~\cite{zhou2014learning} dataset, aiming to include a diverse set of scenes within a limited number of prompts. The selected scenes contain indoor, modern, rural, and various other settings, thereby ensuring the representation of a wide range of environmental contexts. This prompt suite is applied to both the \textit{Background Consistency} dimension and the \textit{Scene} dimension.

\smallTitle{Temporal Flickering.} To more effectively evaluate temporal flickering, it is essential to eliminate interference from other temporal dimensions. According to observations in Section~\ref{suppsec:dimension}, whether the scene is static does not affect the temporal flickering ranking among models. Ultimately, we selected a set of prompts, covering various topics, scenarios, and prompt lengths. Each prompt is accompanied by a prefix instructing the model to generate a static scene. 

\smallTitle{Motion Smoothness.} Since \textit{Subject Consistency}'s prompt suite involves movements performed by different subjects, they serve as a good benchmark for \textit{Motion Smoothness} as well. To minimize the number of videos needed to be sampled for each model in evaluation, we share the same prompt suite for both dimensions.

\smallTitle{Dynamic Degree.} Considering the issue of the model tending to generate static videos even when prompted with descriptions of motion, we use the same prompt suite as \textit{Subject Consistency}'s, which includes a variety of motion descriptions.

\smallTitle{Object Class.} We use the COCO dataset~\cite{lin2014mscoco} and drop the object \texttt{mouse}, due to the potential confusion as it can be interpreted as both device and animal. We then append articles to the rest of the 79 objects and create a list of prompts related to different object classes.

\smallTitle{Multiple Objects.} We categorize COCO objects into various groups so that it will be reasonable for them to appear together. These categories include animals, indoor items, dining objects, bathroom items, and outdoor items. We then generate a list of prompts by composing objects within each category.

\smallTitle{Human Action.} From the Kinetics-400 dataset~\cite{kay2017kinetics}, we carefully extract a subset of 100 actions by considering both diversity and minimal overlaps in their meanings. Our approach involves selecting only the actions that are unique. For instance, within the category of actions related to playing musical instruments, we only keep those actions that are considered dissimilar in terms of human posture and actions. 
The resulting selection contains a wide spectrum of actions. Subsequently, we integrate each action in the form of ``a person is doing something'', and craft a list of human-centric action prompts.

\smallTitle{Color.} We select representative classes from COCO objects and establish the color scope of our prompt suite. On the selection of objects, we select objects that are unique in shape and similar objects. For example, ``skateboard'' and ``surfboard'' are excluded due to their similar shapes and potential wrong detection results by detection models. A similar criterion is applied to the color selection, we aim to select colors to include a broad spectrum while avoiding closely related colors. For example, ``gold'' and ``yellow'' are considered similar, therefore we only include ``yellow'' in our color scope. Our prompts are generated by combining each object with a few of their typical colors, and we only keep objects with more than three typical colors.

\smallTitle{Spatial Relationship.} 
We organize COCO objects into different groups so that it is natural for them to be composed in the same scene with each other. Some examples of the categories include personal items, animals, and sports-related items. Additionally, we define relationship categories to be ``left and right'' and ``top and bottom''. We then select relationships that are reasonable for the objects within each category, resulting in a list of prompts designed to describe spatial relationships between objects.

\smallTitle{Scene.} We use the same prompt suite as \textit{Background Consistency}, as both requires prompts describing different general scenes.

\smallTitle{Appearance Style.} 
We select a list of sentences covering a wide range of scenarios and themes and also define our list of appearance styles. 
The styles are carefully crafted to ensure diversity. For example, we include the representative ``Van Gogh style'' and traditional ``Ukiyo style'' for the clear contrast in their color schemes, brushwork techniques, and overall aesthetic expressions. Each scenario description is then composed with a list of appearance styles to form the prompts.

\smallTitle{Temporal Style.} 
We carefully curate a diverse list of representative temporal styles to represent a broad spectrum of camera movement and temporal effects commonly employed in video production. Our selected temporal styles include variations in motion speed, camera perspective, and dynamic effects, aiming to present a comprehensive range of cinematic techniques. Each sentence for a scenario is then composed with a list of temporal styles.

\smallTitle{Overall Consistency.} We create a range of prompts, covering different content categories and scenarios such as ``natural scenery'', ``fantasy and sci-fi'', ``character and fictional beings'' \etc, these prompts are of varied length, and we include both general and specific descriptions in our prompts.

\subsection{Prompt Suite per Category} 

In Section 3.2 of the main paper on \textit{Prompt Suite Per Category}, we employ LLM~\cite{zheng2023judging} as the first step to categorize the collection of human-curated prompts into eight content categories.
The input template for the language model is shown in Table~\ref{tab:llm_classification_prompt}. 
The accuracy of classification is around 95\%, and we manually go through each classified prompt to filter out 100 prompts for each content category.

\begin{table}[ht]\centering
\begin{minipage}{0.99\columnwidth}\vspace{0mm}    
    \centering
    \begin{tcolorbox} 
        \centering
        \hspace{-6mm}
        \begin{tabular}{p{0.99\columnwidth}}
        \hspace{1mm}
        \begin{minipage}{0.99\columnwidth}
        The assistant gives helpful, detailed, and polite answers to the user's questions. Please act as a language expert, able to choose one or more suitable categories from [\texttt{Animal, Architecture, Food, Human, Lifestyle, Plant, Scenery, Vehicles}] for the given text. Given the input text, you should return the answer without explanation. For example, if the input is [\texttt{A man eats hamburgers.}], the output tag format should be [\texttt{Food, Human}]. \\
        The given text is \texttt{Input text}. 
        \end{minipage}
        \end{tabular}
    \end{tcolorbox}
    \vspace{-2mm}
    \caption{\textbf{Category Classification.} We employ LLM to determine the content categories of collected text descriptions.}
    \label{tab:llm_classification_prompt}
    \vspace{-2mm}
\end{minipage}
\end{table}

\smallTitle{Animal.} 
These prompts focus on various animals and their behaviors in different environments, such as ``a frog eating an ant'',  ``a harbour seal swimming near the shore'', and ``a squirrel eating nuts''. This prompt suite captures diverse species from domestic pets to wild animals in various activities, such as feeding, playing, or simply existing in their natural or adapted environments.

\smallTitle{Architecture.} We keep prompts that include various types of architecture, including the different types of buildings and structures, such as ``the view of the Sydney opera house from the other side of the harbor'', ``illuminated tower in Berlin'', and ``a tree house in the woods''.

\smallTitle{Food.} 
These prompts are diverse and all revolve around food and beverages. They range from specific dishes and preparation methods to more conceptual food art and eating scenarios. Examples include ``Freshly baked finger-licking cookies", ``A person slicing a vegetable", and "Close-up video of Japanese food".

\smallTitle{Human.} 
These prompts describe a wide range of human activities, interactions, and scenes, each focusing on specific individuals or groups engaged in various actions. Here are some examples: ``A family wearing paper bag masks'', ``Boy sitting on grass petting a dog'', ``Group of people protesting'', and ``Father and son holding hands''. Each of these prompts paints a vivid picture of human life, capturing diverse moments from daily activities to special events, professional settings to personal interactions.

\smallTitle{Lifestyle.} 
These prompts describe various indoor scenes and activities, covering a wide range of settings and situations. For instance, ``Interior design of the bar section'' and ``Dog on floor in room'' are simple everyday indoor scenes. Each prompt captures a specific aspect of indoor life, ranging from personal moments and family interactions to professional and leisure activities, reflecting the diversity of experiences within indoor lifestyles.

\smallTitle{Plant.} 
These prompts mainly focus on plants and trees. Here are some examples: ``Video of an indoor green plant'', ``A coconut tree by the house'', and ``Variety of trees and plants in a botanical garden''.

\smallTitle{Scenery.} 
These prompts describe various natural and urban landscapes, each capturing a distinct aspect of the environment. Here are some examples: ``View of the sea from an abandoned building'', ``Aerial footage of a city at night'', and ``Scenery of desert landscape''. Each prompt can be of natural settings like beaches and mountains, the structured scenery of agricultural lands, or urban environments.

\smallTitle{Vehicles.} 
These prompts depict various forms of transportation and related scenes, including various vehicles like trains, cars, buses, motorcycles, and boats in diverse settings ranging from urban streets to natural landscapes. Here are some examples: ``A modern railway station in Malaysia used for public transportation'', ``Train arriving at a station'', ``Elderly couple checking engine of automobile'', and ``Helicopter landing on the street''.

\section{Human Preference Annotation }
\label{suppsec:human}

\subsection{Human Annotation Procedures}

\smallTitle{Labeling Instructions.} 
To systematically communicate with human annotators about labeling rules, we prepare a labeling instruction document for each of the 16 dimensions. 
Each labeling instruction document consists of several important elements. 
First, we introduce the labeling user interface (shown in Figure 4 of the main paper), including the two videos in comparison, the location of prompts and questions, the control for video playback and stop, and the three choices to make (\ie, ``A is better'', ``B is better'', or ``Same quality''). 
Second, we explain the dimension of interest. Since we want to verify the human alignment of VBench in each fine-grained dimension, we conduct the labeling of different dimensions separately. In each document, we elaborate on the definition of the current dimension, including aspects to consider or discard. For instance, for the \textit{Subject Consistency} dimension, annotators are asked to only focus on the look of the main subject, and not to consider the degree of temporal flickering, or the video alignment with the text prompt, and many other irrelevant dimensions. Each aspect to consider or discard is illustrated by both text descriptions and examples of synthesized videos. 
Third, we categorize various scenarios that annotators may encounter while annotating this dimension (\eg, what is considered as ``better'', and what is considered as ``same quality''). For each scenario, we provide explanatory examples.

\smallTitle{Quality Assurance in Preference Annotations.} 
To guarantee the accuracy of human preference annotations, we implement a systematic five-step approach:
\textbf{\textit{1)}} Labeling Instructions Preparation: For each evaluation dimension, we provide clear and well-organized labeling instructions with examples.
\textbf{\textit{2)}} Pre-Labeling Trial: Prior to the main annotation task, we conduct a pre-labeling trial, where annotators are assigned to annotate only 60 samples. We go through all 60 annotations and communicate with annotators about each wrong label, and clarify any misunderstanding or potential doubts in the labeling instructions.
\textbf{\textit{3)}} Labeling Instructions Update: We update the labeling instructions according to feedback from the human annotators, and supplement the wrongly labeled samples into the labeling instructions.
\textbf{\textit{4)}} Post-Labeling Checks by Annotators: Upon labeling all samples for a particular dimension, the samples are grouped as 60 samples per package. In each package of 60 samples, human annotators go through 20\% of randomly selected samples for quality checking. If for any package the error rate exceeds 10\%, the entire package is sent back for re-labeling conducted by a different annotator.
\textbf{\textit{5)}} Post-Labeling Checks by Authors: Upon labeling and possible re-labelings, we conduct the same post-labeling checks procedure similar to step 4. For any labeling errors spotted, we communicate with the human annotator for correction, and ask them to go through the entire package again. If any package reports an error rate higher than 10\%, the entire labeled samples (all packages) for this dimension are considered invalid. We communicate with human annotators regarding possible problems encountered during annotation, and go back to step 1 to conduct annotation for this dimension all over again.

\begin{figure*}[t]
\vspace{10pt}
\centering
\includegraphics[width=0.99\linewidth]{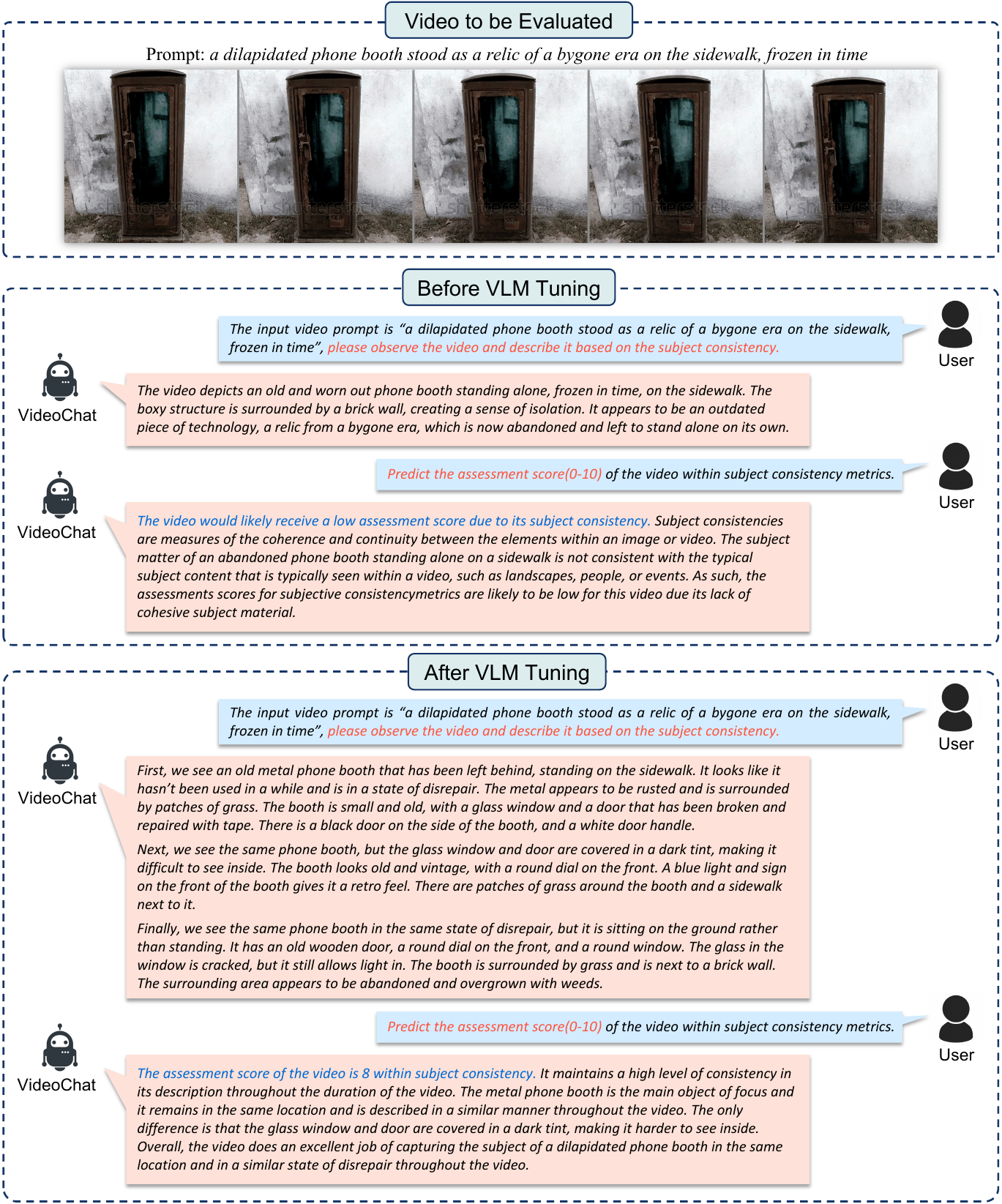}
\caption{\textbf{VLM Tuning Example.}
We use human preference annotation to fine-tune a VLM, and demonstrate that such fine-tuning can improve VLM's ability in evaluating video generation capabilities in specific dimensions.
}
\label{figsupp:vlm_tuning}
\end{figure*}

\subsection{VLM Tuning}

\begin{table}[ht]\centering
\begin{minipage}{0.99\columnwidth}\vspace{0mm}    
    \centering
    \begin{tcolorbox} 
        \centering
        \hspace{-6mm}
        \begin{tabular}{p{0.99\columnwidth}}
        \hspace{1mm}
        \begin{minipage}{0.99\columnwidth}
        \texttt{<SYSTEM>:} You are an assistant with fair and objective standards for video assessment. Given the input statement for the video generation model: \texttt{\{input text\}}, please select appropriate evaluation dimensions from \texttt{[`subject consistency', `background consistency', `temporal flickering', `motion smoothness', `dynamic degree', `aesthetic quality', `imaging quality', `object class', `multiple objects', `human action', `color', `spatial relationship', `scene', `temporal style', 'appearance style', `overall consistency']}. \\
        \rule[0.25\baselineskip]{\textwidth}{1pt}
        \texttt{<USER>:} a dilapidated phone booth stood as a relic of a bygone era on the sidewalk, frozen in time. \\
        \texttt{<ASSISTANT>:} Based on the input statement, the following evaluation dimensions may be relevant:
        \texttt{subject consistency, 
        background consistency, 
        temporal flickering, 
        motion smoothness, 
        dynamic degree, 
        aesthetic quality, 
        imaging quality, 
        scene, 
        temporal style, 
        overall consistency}
        \end{minipage}
        \end{tabular}
    \end{tcolorbox}
    \vspace{-2mm}
    \caption{\textbf{VLM Chain of Thought.} VLM suggests appropriate evaluation dimensions based on the user-input video prompt.}
    \label{tab:vlm_cot}
    \vspace{-2mm}
\end{minipage}
\end{table}

\smallTitle{VLM Tuning.} 
Our human preference annotations are mainly used to validate the alignment of VBench evaluations with human perception in each dimension. We show that these annotations have the potential to be used for broader applications, such as fine-tuning VLM models to align more closely with human preferences. 
VideoChat~\cite{2023videochat} uses a pre-trained video encoder and a large language model to align and train video VLM with instruction data.
We organize data from VBench for fine-tuning, enhancing the model's ability to assess video generation.
By utilizing instruction data of the VBench Dataset coupled with human preference tags,  VLM, through its cognitive chain, is with its Chain of Thought to choose suitable evaluation metrics and furnishing scores in accordance. Upon giving the prompt into the VLM and asking: ``\texttt{Give the appropriate evaluation metrics}", we obtain the dimensions to be evaluated from Chain of Thought, as shown in Table~\ref{tab:vlm_cot}. We feed the derived metrics along with the video into the VLM, and ask two additional questions: ``\texttt{Please observe the video and describe it based on the provided metrics.}" and ``\texttt{Predict the assessment score of the video within these metrics.}"
We show examples of before and after VLM Tuning in Figure~\ref{figsupp:vlm_tuning}

\smallTitle{Data Preparation.}
We organize human preference annotations and model scores across different dimensions into training datasets. For model scores, each video's ratings are mapped from 0-1 to a 0-10 scale and are coupled with questions used for human annotation to form our instruction data. Evaluations of different aspects of the same video are grouped to facilitate the VLM's ability to engage in multi-turn dialogues. As for human preference annotations, pairs of videos along with corresponding questions and preference options are organized into instruction data, thereby somewhat enhancing the model with the capability to make evaluative judgments.

\smallTitle{Implementation Details}
Our approach uses VideoChat-embed~\cite{2023videochat} as the baseline model. The model undergoes fine-tuning on a set of 30,000 instruction pairs. We fine-tune the model for 3 epochs using a learning rate of 2e-5 and an overall batch size of 64. The training for our model takes about one hour when performed on 8 A100-80GB GPUs.

\begin{figure*}[!htb]
    \centering
    \includegraphics[width=0.99\linewidth]{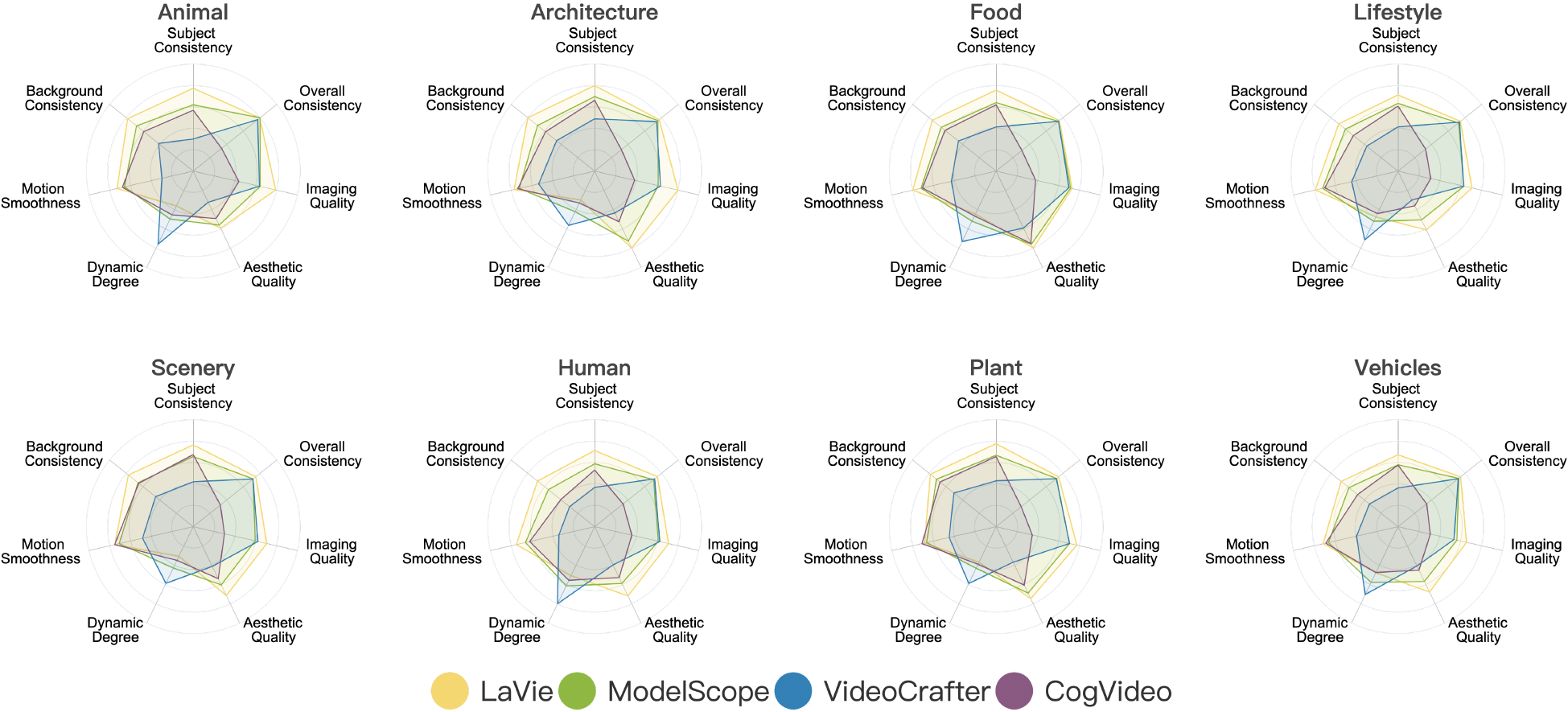}
    \caption{\textbf{VBench Results across Eight Content Categories (by Category per Chart)} (best viewed in color). For each chart, we plot the VBench evaluation results across different models on the same content category.}
    \label{figsupp:radar_per_category}
    \vspace{10pt}
\end{figure*}

\begin{table*}[!htb]
\vspace{5pt}
\centering
\setlength\tabcolsep{3pt}
\begin{center}
\small
\caption{
\textbf{Validate VBench’s Human Alignment.} We report \textit{VBench Win Ratios (left) / Human Win Ratios (right)} for each dimension and each model. Our experiments show that VBench evaluations across all dimensions closely match human perceptions. }
\vspace{-6pt}
\resizebox{1.03\linewidth}{!}{
\begin{tabular}{c|c|c|c|c|c|c|c|c}
\Xhline{1pt}
\textbf{Models}   & \textbf{\Centerstack{Subject\\Consistency}} & \textbf{\Centerstack{Background\\Consistency}} & 
\textbf{\Centerstack{Temporal\\Flickering}} & \textbf{\Centerstack{Motion\\Smoothness}} & \textbf{\Centerstack{Dynamic\\Degree}} & \textbf{\Centerstack{Aesthetic\\Quality}} &  \textbf{\Centerstack{Imaging\\Quality}} & \textbf{\Centerstack{Object\\Class}} \\ \Xhline{1pt}
LaVie~\cite{wang2023lavie}        & \textbf{67.87\%} / \textbf{69.95\%} & \textbf{85.27\%} / \textbf{65.04\%} & \textbf{73.42\%} / \textbf{87.96\%} & \textbf{69.54\%} / \textbf{65.65\%} & 41.81\% / 53.10\% & \textbf{77.56\%} / \textbf{83.41\%} & \textbf{77.20\%} / \textbf{79.46\%} & \textbf{57.55\%} / \textbf{79.20\%} \\ 
ModelScope~\cite{luo2023videofusion, wang2023modelscope}  & 49.07\% / 56.30\% & 49.96\% / 56.36\% & 65.42\% / 62.44\% & 58.61\% / 59.58\% & 52.92\% / 53.84\% & 67.74\% / 63.15\% & 60.00\% / 68.53\% & 49.37\% / 49.58\% \\ 
VideoCrafter~\cite{he2022lvdm} & 24.72\% / 20.42\% & 15.89\% / 27.21\% & 31.20\% / 43.64\% & 10.00\% / 13.80\% & \textbf{68.47\%} / \textbf{62.18\%} & 35.34\% / 32.33\% & 55.05\% / 37.85\% & 54.18\% / 41.77\% \\ 
CogVideo~\cite{hong2022cogvideo}  & 58.33\% / 53.33\% & 48.88\% / 51.40\% & 29.96\% / 5.96\% & 61.85\% / 60.97\% & 36.81\% / 30.88\% & 19.35\% / 21.11\% & 7.74\% / 14.16\% & 38.90\% / 29.45\% \\ \hline
Correlation  & 96.51\% & 94.80\% & 88.73\% & 99.80\% & 82.09\% & 98.65\% & 92.16\% & 80.37\% \\ \hline
\hline
\textbf{Models}   & \textbf{\Centerstack{Multiple\\Objects}} & \textbf{\Centerstack{Human\\Action}} & \textbf{Color} & \textbf{\Centerstack{Spatial\\Relationship}} & \textbf{Scene} & \textbf{\Centerstack{Appearance\\Style}} & \textbf{\Centerstack{Temporal\\Style}} & \textbf{\Centerstack{Overall\\Consistency}} \\
\Xhline{1pt}
LaVie~\cite{wang2023lavie}        & 53.37\% / 57.97\% & \textbf{54.43\%} / \textbf{58.13\%} & \textbf{52.31\%} / \textbf{51.37\%} & 52.30\% / 49.81\% & \textbf{59.69\%} / \textbf{77.52\%} & \textbf{61.85\%} / \textbf{58.22\%} & \textbf{69.07\%} / 55.73\% & \textbf{70.82\%} / \textbf{77.35\%} \\ 
ModelScope~\cite{luo2023videofusion, wang2023modelscope}  & \textbf{57.15\%} / \textbf{62.15\%} & 51.10\% / 53.07\% & 50.12\% / 49.73\% & 53.25\% / 53.15\% & 48.22\% / 50.00\% & 57.48\% / 54.93\% & 65.40\% / \textbf{57.50\%} & 66.31\% / 60.07\% \\ 
VideoCrafter~\cite{he2022lvdm} & 48.74\% / 49.63\% & 52.17\% / 47.87\% & 48.71\% / 47.92\% & \textbf{56.11\%} / \textbf{54.66\%} & 52.79\% / 46.05\% & 36.67\% / 40.07\% & 65.40\% / 51.90\% & 62.65\% / 48.10\% \\ 
CogVideo~\cite{hong2022cogvideo}  & 40.73\% / 30.24\% & 42.30\% / 40.93\% & 48.86\% / 50.98\% & 38.33\% / 42.38\% & 39.30\% / 26.43\% & 44.00\% / 46.78\% & 0.13\% / 34.87\% & 0.22\% / 14.48\% \\ \hline
Correlation  & 98.98\% & 89.15\% & 60.73\% & 97.59\% & 94.07\% & 99.65\% & 97.53\% & 93.27\% \\
\Xhline{1pt}
\end{tabular}
}
\label{tab:win_ratio}
\end{center}
\end{table*}

\begin{table*}[!htb]
\centering
\setlength\tabcolsep{3pt}
\begin{center}
\small
\vspace{15pt}
\caption{\textbf{VBench Evaluation Results on the WebVid-Avg Reference Baseline.} This table shows the VBench evaluation results on the \textit{WebVid-Avg} baseline. We provide results from other models and baselines as well for a comprehensive view.}
\vspace{-6pt}
\resizebox{1.0\linewidth}{!}{
\begin{tabular}{c|c|c|c|c|c|c|c|c|c}
\Xhline{1pt}
\textbf{Models}   & \textbf{\Centerstack{Subject\\Consistency}} & \textbf{\Centerstack{Background\\Consistency}} & 
\textbf{\Centerstack{Motion\\Smoothness}} & \textbf{\Centerstack{Dynamic\\Degree}} & \textbf{\Centerstack{Aesthetic\\Quality}} &  \textbf{\Centerstack{Imaging\\Quality}} & 
\textbf{\Centerstack{Appearance\\Style}} &
\textbf{\Centerstack{Temporal\\Style}} &
\textbf{\Centerstack{Overall\\Consistency}} \\ \Xhline{1pt}
LaVie~\cite{wang2023lavie}        & 91.41\% & \textbf{97.47\%} & 96.38\% & 49.72\% & \textbf{54.94\%} & \textbf{61.90\%} & \textbf{23.56\%} & \textbf{25.93\%} & \textbf{26.41\%} \\ 
ModelScope~\cite{luo2023videofusion, wang2023modelscope}  & 89.87\% & 95.29\% & 95.79\% & 66.39\% & 52.06\% & 58.57\% & 23.39\% & 25.37\% & 25.67\%\\ 
VideoCrafter~\cite{he2022lvdm} & 86.24\% & 92.88\% & 91.79\% & \textbf{89.72\%} & 44.41\% & 57.22\% & 21.57\% & 25.42\% & 25.21\%\\ 
CogVideo~\cite{hong2022cogvideo}  & \textbf{92.19\%} & 95.42\% & \textbf{96.47\%} & 42.22\% & 38.18\% & 41.03\% & 22.01\% & 7.80\% & 7.70\%\\ \hline
Empirical Min  & 14.62\% & 26.15\% & 70.60\% & 0.00\% & 0.00\% & 0.00\% & 0.09\% & 0.00\% & 0.00\% \\
WebVid Avg  & 96.17\% & 96.59\% & 98.17\% & 44.13\% & 42.37\% & 58.22\% & 22.15\% & 25.77\% & 34.14\% \\
Empirical Max  & 100.00\% & 100.00\% & 99.75\% & 100.00\% & 100.00\% & 100.00\% & 28.55\% & 36.40\% & 36.40\%\\ \hline
\Xhline{1pt}

\end{tabular}
}
\label{tabsupp:webvid-avg}
\end{center}
\end{table*}

\begin{table*}[!htb]
\vspace{-10pt}
\centering
\setlength\tabcolsep{3pt}
\begin{center}
\small 
\caption{\textbf{VBench Results across Eight Content Categories.} We show the VBench evaluation results on the four T2V models, across eight content categories, on various evaluation dimensions.}
\vspace{-6pt}
\begin{tabular}{c|c|C{1.7cm}|C{1.7cm}|C{1.7cm}|C{1.7cm}|C{1.7cm}|C{1.7cm}|C{1.7cm}}

\Xhline{1pt}
\textbf{Models}   &  
\textbf{Categories} &
\textbf{\Centerstack{Subject\\Consistency}} & \textbf{\Centerstack{Background\\Consistency}} & \textbf{\Centerstack{Motion\\Smoothness}} &  \textbf{\Centerstack{Dynamic\\Degree}} & \textbf{\Centerstack{Aesthetic\\Quality}} &
\textbf{\Centerstack{Imaging\\Quality}} & 
\textbf{\Centerstack{Overall\\Consistency}} \\ \Xhline{1pt}

\multirow{8}{*}{\Centerstack{\textbf{LaVie}\\\cite{wang2023lavie}}}
& \textbf{Animal}  & 97.49\% & 97.18\% & 97.29\% & 15.20\% & 48.26\% & 68.81\% & \textbf{26.43\%} \\ 
& \textbf{Architecture}   & \textbf{98.04\%} & \textbf{97.38\%} & 97.83\% & 5.20\% & \textbf{54.20\%} & \textbf{69.30\%} & 25.46\% \\ 
& \textbf{Food}  & 97.11\% & 96.90\% & \textbf{98.18\%} & 28.80\% & 54.15\% & 65.24\% & 24.88\% \\ 
& \textbf{Lifestyle}  & 96.10\% & 96.19\% & 98.08\% & 33.60\% & 48.76\% & 64.02\% & 24.43\% \\ 
& \textbf{Scenery}  & 97.27\% & 97.06\% & 97.58\% & 6.40\% & 51.76\% & 63.86\% & 24.56\% \\ 
& \textbf{Human}   & 96.11\% & 95.88\% & 97.57\% & \textbf{39.00\%} & 51.87\% & 64.07\% & 24.63\% \\ 
& \textbf{Plant}  & 97.52\% & 97.20\% & 96.73\% & 16.40\% & 52.68\% & 67.86\% & 24.50\% \\ 
& \textbf{Vehicles}  & 95.23\% & 95.82\% & 97.11\% & 34.00\% & 50.70\% & 61.02\% & 24.51\% \\ 
\Xhline{1pt}
\multirow{8}{*}
{\Centerstack{\textbf{ModelScope}\\\cite{luo2023videofusion, wang2023modelscope}}}
& \textbf{Animal}  & 94.08\% & 95.80\% & 96.40\% & 37.20\% & 47.32\% & 60.30\% & \textbf{26.58\%} \\ 
& \textbf{Architecture}   & \textbf{95.77\%} & 95.88\% & \textbf{97.20\%} & 24.80\% & 52.10\% & 58.38\% & 24.89\% \\ 
& \textbf{Food}  & 94.53\% & 95.53\% & 97.17\% & 40.80\% & \textbf{53.06\%} & \textbf{64.39\%} & 24.40\% \\ 
& \textbf{Lifestyle}  & 94.36\% & 95.17\% & 97.18\% & 41.00\% & 45.77\% & 59.62\% & 23.51\% \\ 
& \textbf{Scenery}  & 94.88\% & 95.57\% & 97.03\% & 26.00\% & 48.57\% & 57.49\% & 23.28\% \\ 
& \textbf{Human}   & 93.37\% & 94.21\% & 96.45\% & \textbf{56.00\%} & 48.14\% & 58.41\% & 22.84\% \\ 
& \textbf{Plant}  & 95.14\% & \textbf{96.26\%} & 96.48\% & 26.40\% & 51.03\% & 63.83\% & 23.55\% \\ 
& \textbf{Vehicles}  & 93.17\% & 94.61\% & 96.47\% & 50.20\% & 47.53\% & 55.75\% & 23.60\% \\ 
\Xhline{1pt}
\multirow{8}{*}{\Centerstack{\textbf{VideoCrafter}\\\cite{he2022lvdm}}}
& \textbf{Animal}  & 87.01\% & 92.40\% & 91.80\% & 79.60\% & 40.51\% & 59.79\% & \textbf{25.47\%} \\ 
& \textbf{Architecture}   & \textbf{91.18\%} & 92.93\% & \textbf{94.83\%} & 47.80\% & 43.71\% & 59.63\% & 24.27\% \\ 
& \textbf{Food}  & 89.50\% & 92.87\% & 93.44\% & 75.00\% & \textbf{48.19\%} & 63.47\% & 24.47\% \\ 
& \textbf{Lifestyle}  & 89.51\% & 91.87\% & 93.63\% & 72.20\% & 39.84\% & 59.44\% & 24.01\% \\ 
& \textbf{Scenery}  & 89.67\% & 92.86\% & 94.17\% & 51.80\% & 43.06\% & 58.98\% & 23.20\% \\ 
& \textbf{Human}   & 88.50\% & 90.92\% & 92.35\% & \textbf{86.20\%} & 42.62\% & 59.23\% & 23.31\% \\ 
& \textbf{Plant}  & 89.86\% & \textbf{93.57\%} & 93.72\% & 52.00\% & 41.81\% & \textbf{63.81\%} & 23.41\% \\ 
& \textbf{Vehicles}  & 88.38\% & 91.44\% & 93.04\% & 70.60\% & 42.95\% & 54.14\% & 23.39\% \\ 
\Xhline{1pt}
\multirow{8}{*}{\Centerstack{\textbf{CogVideo}\\\cite{hong2022cogvideo}}}
& \textbf{Animal}  & 92.95\% & 94.69\% & 96.65\% & 30.20\% & 45.37\% & \textbf{48.45\%} & 8.26\% \\ 
& \textbf{Architecture}   & 95.00\% & 94.65\% & 97.39\% & 10.20\% & 46.29\% & 45.33\% & 7.48\% \\ 
& \textbf{Food}  & 94.08\% & 94.94\% & 96.99\% & 32.00\% & \textbf{52.79\%} & 45.05\% & 7.01\% \\ 
& \textbf{Lifestyle}  & 93.80\% & 93.93\% & 96.93\% & 28.00\% & 41.57\% & 41.28\% & 7.85\% \\ 
& \textbf{Scenery}  & \textbf{95.27\%} & 95.46\% & \textbf{97.58\%} & 13.20\% & 46.72\% & 40.49\% & 7.66\% \\ 
& \textbf{Human}   & 92.08\% & 92.29\% & 95.93\% & \textbf{46.80\%} & 46.38\% & 43.81\% & 8.29\% \\ 
& \textbf{Plant}  & 94.86\% & \textbf{95.71\%} & 97.05\% & 19.60\% & 48.63\% & 43.22\% & 6.65\% \\ 
& \textbf{Vehicles}  & 93.11\% & 93.27\% & 96.80\% & 33.60\% & 44.18\% & 41.05\% & \textbf{8.34\%} \\ 
\Xhline{1pt}
\end{tabular}
\label{tab:per-category}
\end{center}
\end{table*}

\section{More Implementation Details}
\label{suppsec:implementation}

\subsection{Video Generation Models in Evaluation}
\label{suppsubsec:models_in_evaluation}

To evaluate our benchmark on recent advances, we adopt four models for comparison, and more will be added as they become open-sourced. Details of models and sampling strategy are listed as follows.  

\smallTitle{LaVie.}
LaVie~\cite{wang2023lavie} is a high-quality video generation model that incorporates cascaded latent diffusion models. Specifically, a set of temporal modules is attached to the vanilla Stable Diffusion~\cite{rombach2022ldm} model and the entire model is jointly trained on both images and videos to achieve video generation. For each prompt, we sample 16 continuous frames of size 512$\times$512 at 8 frames per second (FPS). We use the DDPM sampling of 250 steps. The initial random seed is set to 2 and the classifier-free guidance is set to 7.

\smallTitle{ModelScope.}
ModelScope~\cite{wang2023modelscope, luo2023videofusion} is a diffusion-based text-to-video generation model. We adopt its official inference code and sample 16 frames of size 256$\times$256 at 8 FPS.

\smallTitle{VideoCrafter.}
VideoCrafter~\cite{he2022lvdm} is a toolkit for text-to-video generation and editing. We adopt the VideoCrafter 0.9 version (\textit{a.k.a.}, LVDM) and utilize its base generic text-to-video generation model. We use the official inference code to sample 16 frames of size 256$\times$256 at 8 FPS. The initial random seed is set to 2 during sampling.

\smallTitle{CogVideo.}
CogVideo~\cite{hong2022cogvideo} is a transformer-based text-to-video generation model that inherits the pretrained text-to-image model CogView2~\cite{ding2022cogview2}. Since the official inference code requires simplified Chinese input, we translate all prompts into Chinese. We sample 33 frames of size 480$\times$480 at 10 FPS for each video, according to its default settings. During sampling, all stages are involved in the pipelines, including sequential generation, frame interpolation, and recursive interpolation. The initial random seed is also set to 2 for a fair comparison.

\subsection{Reference Baselines}

In the main paper, we devise the \textit{Empirical Max} and \textit{Empirical Min} baselines to approximate the maximum / minimum scores that videos might be able to achieve. We also devise the \textit{WebVid-Avg} baseline to reflect the average video quality of WebVid-10M dataset~\cite{bain2021frozen} as a reference. The numerical results are displayed in Table 1 in the main paper, and Table~\ref{tabsupp:webvid-avg} in this Supplementary File.  We provide additional details on approximating these values as follows.

\smallTitle{Empirical Max.} 
\textit{(1) WebVid-10M's Maximum.} For dimensions where the 100\% score is unlikely to be achieved by any video, we retrieve WebVid-10M's real videos and report the highest-scoring video's result. Examples of such dimensions include \textit{Motion Smoothness}, \textit{Scene}, \textit{Appearance Style}, \textit{Temporal Style}, and \textit{Overall Consistency}.
\textit{(2) Theoretical 100\%.} For dimensions where there exist videos that can achieve 100\%, we directly use 100\% as the empirical maximum value. For temporal consistency dimensions \textit{Subject Consistency}, \textit{Background Consistency}, and \textit{Temporal Flickering}, a completely static video corresponds to the 100\% score. For \textit{Dynamic Degree}, a set of highly dynamic videos can achieve the 100\% ratio of dynamic degree. For the frame-wise quality dimensions \textit{Aesthetic Quality} and \textit{Imaging Quality}, a video consisting of 100\%-scoring frames results in a final 100\% score. For video-text semantics dimensions \textit{Object Class}, \textit{Multiple Objects}, \textit{Human Actions}, \textit{Color}, and \textit{Spatial Relationship}, videos with the correct semantics specified in the text prompt can score 100\%.

\smallTitle{Empirical Min.}
\textit{(1) Gaussian Noise Videos.} 
For video-text feature similarity dimensions \textit{Appearance Style}, \textit{Temporal Style}, and \textit{Overall Consistency},
we use videos of i.i.d. Gaussian noise and the corresponding prompt suites to compute the corresponding score, and select the smallest value as the approximated empirical minimum (with some actually reaching 0\%).  For \textit{Temporal Flickering} and \textit{Motion Smoothness}, we directly compute the score of the Gaussian noise videos and take the minimum scoring video's result. For \textit{Human Action}, our method suite gives 0\% on the Gaussian noise videos.
\textit{(2) Composed Videos.} For temporal consistency dimensions \textit{Subject Consistency} and \textit{Background Consistency},
we randomly sample frames from different WebVid-10M~\cite{bain2021frozen} videos to form a video with dynamically shifting content. This procedure is repeated 1000 times, and the minimum score among all videos obtained serves as the empirical minimum reference. 
\textit{(3) Theoretical 0\%.} For dimensions where there exist videos that can achieve 0\%, we directly use 0\% as the empirical minimum value. For \textit{Dynamic Degree}, a set of static videos can achieve the 0\% ratio of dynamic degree. For the frame-wise dimensions \textit{Aesthetic Quality} and \textit{Imaging Quality}, a video consisting of 0\%-scoring frames results in a final 0\% score. For video-text semantics dimensions \textit{Object Class}, \textit{Multiple Objects}, \textit{Color}, \textit{Spatial Relationship}, and \textit{Scene}, videos with the incorrect semantics specified in the text prompt can score 100\%.

\smallTitle{WebVid-Avg.}
For dimensions where WebVid-10M videos can be retrieved with high confidence according to their captions, such as \textit{Subject Consistency}, \textit{Background Consistency}, \textit{Motion Smoothness}, \textit{Dynamic Degree}, \textit{Aesthetic Quality}, \textit{Imaging Quality}, \textit{Appearance Style}, \textit{Temporal Style}, and \textit{Overall Consistency}, we compute the average score for all retrieved videos in relation to the corresponding dimension. This average score serves as a reference value for the average of real videos. The results are visualized in the main paper Figure 6 (b), and detailed in Table~\ref{tabsupp:webvid-avg} in this Supplementary File.

\subsection{Normalization for Radar Chart Visualization}

In the radar charts, we perform normalization to clearly visualize the relative performance. We detail the normalization methods as follows:
\begin{itemize}[leftmargin=10pt]
\item \textit{Main Paper Figure 2. VBench Evaluation Results of Video Generative Models} - For each dimension, we map the maximum score achieved by one of the T2V models to 0.8, and the minimum score to 0.3, and linearly map the remaining models' scores to the radar chart axes. The radar chart axes have a range from 0.0 to 1.0.
\item \textit{Main Paper Figure 6 (a). T2V vs. T2I} - For each dimension, we map the maximum score achieved by one of the models (including T2I and T2V models) to 0.8, and the minimum score to 0.3, and linearly map the remaining models' scores to the radar chart axes. The radar chart axes have a range from 0.0 to 1.0.
\item \textit{Main Paper Figure 6 (b). T2V vs. WebV-Avg \& Max} - For each dimension, we map the maximum score achieved by one of the models (including the \textit{Empirical Max} and \textit{WebVid-Avg} baselines) to 0.8, and the minimum score to 0.3, and linearly map the remaining models' scores to the radar chart axes. The radar chart axes have a range from 0.0 to 1.0.
\item \textit{Main Paper Figure 7. VBench Results across Eight Content Categories (by Model)} - For each dimension, there are 32 numerical results corresponding to the four T2V models and eight content categories. We map the maximum score among the 32 results to 1.0, and the minimum score among the 32 results to 0.0, and linearly map the remaining 30 scores to respective radar charts' axes. The radar chart axes have a range from 0.0 to 1.0.
\item \textit{Supp File Figure~\ref{figsupp:radar_per_category}. VBench Results across Eight Content Categories (by Category)} - Unlike Figure 7 in the main paper which put different categories of the same model in one radar chart, in Figure~\ref{figsupp:radar_per_category} we use an alternative visualization method, that is, collecting different models' results of the same category in one radar chart. For each dimension, there are 32 numerical results corresponding to the four T2V models and eight content categories. We map the maximum score among the 32 results to 0.8, and the minimum score among the 32 results to 0.3, and linearly map the remaining 30 scores to respective radar charts' axes. The radar chart axes have a range from 0.0 to 1.0.
\end{itemize}

\begin{table*}[t]
\centering
\setlength\tabcolsep{3pt}
\begin{center}
\small
\caption{\textbf{VBench Evaluation Results of Video vs. Image Generation Models.} We compare the performance of four video generation models against three image generation models. For each evaluation dimension, a higher score represents relatively better performance. For \textit{Overall Consistency} we replaced the ViCLIP approach by CLIP to enable evaluating image generation models. 
}
\vspace{-6pt}
\begin{tabular}{c|c|c|c|c|c|c|c|c|c|c}

\Xhline{1pt}
\textbf{Models}   &  \textbf{\Centerstack{Aesthetic\\Quality}} &  \textbf{\Centerstack{Imaging\\Quality}} & \textbf{\Centerstack{Object\\Class}} &  \textbf{\Centerstack{Multiple\\Objects}} & \textbf{\Centerstack{Human\\Action}} &
\textbf{Color} & 
\textbf{\Centerstack{Spatial\\Relationship}} & 
\textbf{Scene} & 
\textbf{\Centerstack{Appearance\\Style}} & \textbf{\Centerstack{Overall\\Consistency}} \\ \Xhline{1pt}

LaVie~\cite{wang2023lavie}    & 54.94\% & 61.90\% & 91.82\% & 33.32\% & \textbf{96.80\%} & 86.39\% & 34.09\% & 52.69\% & 23.56\% & 32.96\% \\ 
ModelScope~\cite{luo2023videofusion, wang2023modelscope}   & 52.06\% & 58.57\% & 82.25\% & 38.98\% & 92.40\% & 81.72\% & 33.68\% & 39.26\% & 23.39\% & 31.99\% \\ 
VideoCrafter~\cite{he2022lvdm}  & 44.41\% & 57.22\% & 87.34\% & 25.93\% & 93.00\% & 78.84\% & 36.74\% & 43.36\% & 21.57\%  & 30.78\% \\ 
CogVideo~\cite{hong2022cogvideo}   & 38.18\% & 41.03\% & 73.40\% & 18.11\% & 78.20\% & 79.57\% & 18.24\% & 28.24\% & 22.01\%  & 27.80\% \\ 
\Xhline{1pt}
SD1.4~\cite{rombach2022ldm}  & 65.85\% & \textbf{69.86\%} & 91.14\% & 34.39\% & 91.80\% & 90.57\% & 61.89\% & 52.33\% & 25.35\% & 32.59\% \\ 
SD2.1~\cite{rombach2022ldm} & 66.50\% & 69.10\% & \textbf{93.42\%} & 51.22\% & 89.00\%  & \textbf{91.15\%} & 73.11\% & \textbf{58.14\%} & \textbf{25.48\%} & 33.08\% \\ 
SDXL~\cite{podell2023sdxl}  & \textbf{70.38\%} & 68.79\% & 91.39\% & \textbf{69.51\%} & 91.20\%  & 88.92\% & \textbf{86.17\%} & 54.65\% & 25.23\% & \textbf{33.77\%}  \\\Xhline{1pt}
\end{tabular}
\vspace{-0.3cm}
\label{tab:raw_metrics_sd}
\end{center}
\end{table*}

\section{Potential Negative Societal Impacts}
\label{suppsec:impact}

Video generation models could be maliciously applied to generate fake content involving human figures. Moreover, generative models can potentially inherit biases from the training datasets~\cite{esser2020note}.
Therefore, we recognize the importance of considering ethical and safety aspects when evaluating video generation models. We plan to include safety and equality dimensions in future iterations of VBench. We also urge users to apply video generation models with discretion.

\section{Limitations and Future Work}
\label{suppsec:limitation}

\noindent\textbf{Limited Amount of Open-Sourced T2V Models}: Currently, the number of open-sourced T2V models are still limited. We will open-source our VBench and encourage more T2V models to participate in the evaluation, including but not limited to \cite{pikalab, morphstudio, zeroscope, zhang2023show1, Gen2, blattmann2023videoldm}, so that we can provide more informed insights into the current state of T2V, and provide more annotated data on T2V generation results generated by different models.

\noindent\textbf{Evaluation of Other Video Generation Tasks}: Text-to-video (T2V) is a fundamental task in video generation, and there are other related video generation tasks such as video-driven (\ie, video editing)~\cite{liew2023magicedit, chai2023stablevideo, xing2023simda, ouyang2023codef, tokenflow2023, huang2023inve, couairon2023videdit, yang2023rerender, liu2023videop2p, zhang2023consistent, zhao2023controlvideo, vid2vid-zero, khachatryan2023text2videozero, ceylan2023pix2video, qi2023fatezero, molad2023dreamix, lee2023textvideoedit, zhao2023makeaprotagonist, wu2022tuneavideo}, image-driven (\ie, image-to-video)~\cite{yin2023dragnuwa, wang2023disco, wang2023leo, yu2023magvit, chen2023mcdiff, gu2023seer, ni2023conditional, fu2023tell, esser2023structure, wang2023learning, singer2022makeavideo, song2022textdriven, wu2021nuwa, chen2023seine, molad2023dreamix}, personalized video generation~\cite{dreampose_2023, zhao2023makeaprotagonist, he2023animate, guo2023animatediff}, and other types of multi-modal-controlled video synthesis~\cite{dreampose_2023, zhang2023magicavatar, hu2023videocontrolnet, lemoing2022waldo, 2023videocomposer, wang2023disco, xing2023make, chu2023video, chen2023controlavideo, zhang2023controlvideo, ma2023follow, khachatryan2023text2videozero, wu2022tuneavideo}. We build our VBench towards T2V as the initial step, and plan to extend our benchmark suite to accommodate other modalities' controls by adding towards the ``\textit{Video-Condition Consistency}'' dimensions. Our ``\textit{Video Quality}'' dimensions are readily available for evaluating these video generation tasks.
\section{Additional Experimental Results}
\label{suppsec:additional}

\begin{figure}[t]
\centering
\includegraphics[width=0.80\linewidth]{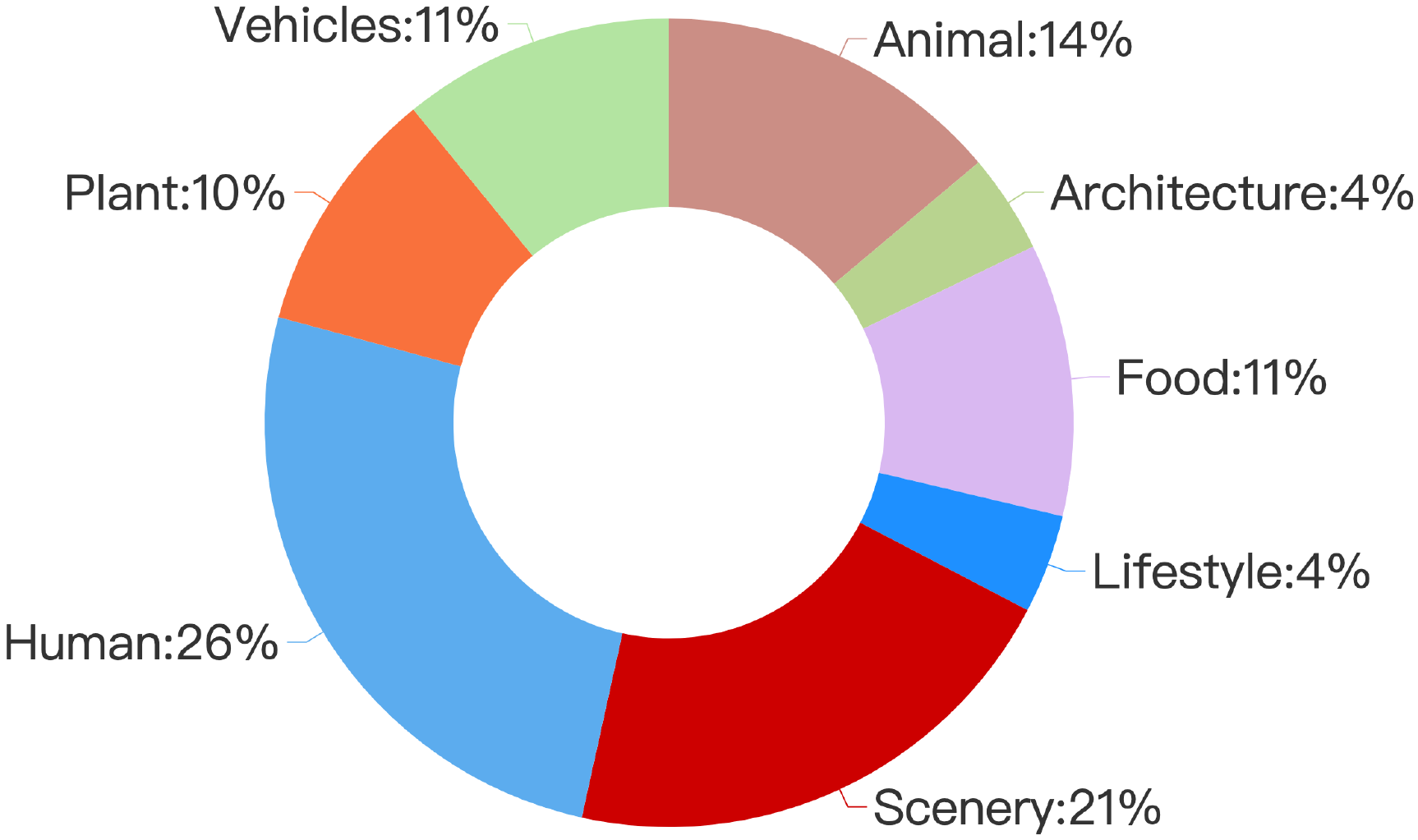}
\caption{\textbf{WebVid-10M Dataset Categorical Distribution.} We visualize the percentage of data amount of each of the eight content categories in the WebVid-10M dataset.}
\label{fig:pie_chart_of_webv}
\end{figure}

\begin{figure}[t]
\centering
\includegraphics[width=0.80\linewidth]{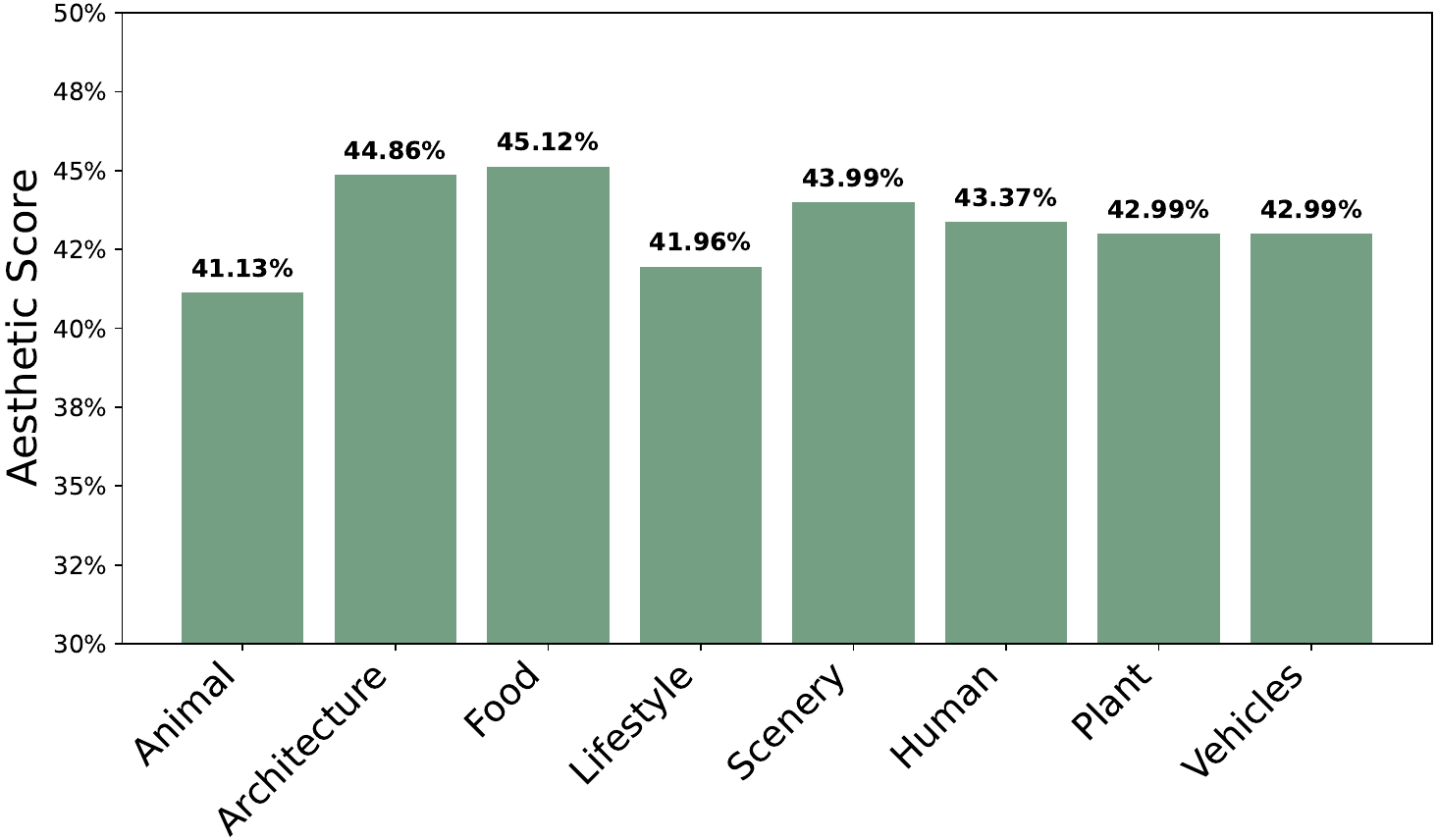}
\caption{\textbf{Aesthetic Quality for Eight Categories in WebVid-10M dataset.} We visualize the aesthetic score of each of the eight content categories in the WebVid-10M dataset.
}
\label{fig:bar_chart_of_aesthetic}
\end{figure}

In this section, we provide additional numerical results that correspond to the main paper visualizations. We list the resulting tables and figures as follows:
\begin{itemize}[leftmargin=10pt]
\item In Table~\ref{tab:raw_metrics_sd}, we show the VBench evaluation results of four video generation models and three image generation models, further illustrating through numerical results the significant differences that exist in certain dimensions between video generation models and image generation models (corresponding to \textit{main paper Figure 6 (a)}). For \textit{Overall Consistency} we replaced the ViCLIP approach by CLIP to enable evaluating image generation models. 
\item In Table~\ref{tab:win_ratio}, we show the win ratio on evaluation results predicted by VBench and Human across four models and all dimensions, along with the correlation ($\rho$) between Human and VBench results (corresponding to \textit{main paper Figure 5}).
\item In Table~\ref{tabsupp:webvid-avg}, we show the results of WebVid-Avg and compare them with the results of four models and other reference baselines (corresponding to \textit{main paper Figure 6 (b)}).
\item In Table~\ref{tab:per-category}, we show all the evaluation results of VBench across four models and eight different categories, providing numerical support for the relevant observations in the insights. (corresponding to \textit{main paper Figure 7}). Additionally, for the \textit{Dynamic Degree} dimension, intrinsic attributes of different categories naturally result in noticeable differences in the dynamic degrees among various categories. For instance, the \texttt{Human} category consistently exhibits the highest \textit{dynamic degree} across different models. Conversely, the \texttt{Architecture}, \texttt{Scenery}, and \texttt{Plant} categories consistently showcase the lowest \textit{dynamic degree} across various models, and the ascending order from lowest to highest remains consistent as \texttt{Architecture}, \texttt{Scenery}, and \texttt{Plant}. Due to this characteristic, the dynamic degree shows significant variability across different categories. Therefore, we isolate it as a supplementary dimension for additional analysis on top of other dimensions. 
\item In Figure~\ref{fig:pie_chart_of_webv}, we show the statistical distribution of data amount of each of the eight content categories in the WebVid-10M dataset (supporting observations and insights mentioned in the \textit{main paper Section 5}).
\item In Figure~\ref{fig:bar_chart_of_aesthetic}, we show the aesthetic scores of eight different categories within the WebVid-10M dataset (supporting observations and insights mentioned in the \textit{main paper Section 5}).
\end{itemize}

\clearpage

\end{document}